\documentclass[11pt]{ociamthesis}  % default square logo 

\usepackage[margin=3.5cm]{geometry}
\usepackage{amssymb}
\usepackage{amsmath}
\usepackage{amsthm}
\usepackage{apptools}
\usepackage{bbm}
\usepackage{bm}
\usepackage{breakcites}
\usepackage{centernot}
\usepackage{chngcntr}
\usepackage{dsfont}
\usepackage[inline, shortlabels]{enumitem}
\usepackage{mathtools}
\usepackage{setspace}
\usepackage[utf8]{inputenc}
\usepackage{parskip}
\usepackage{scalerel}
\usepackage{tikz-cd}
\usepackage{upquote}
\usepackage{xcolor}

\usepackage[backend=bibtex,bibstyle=numeric,citestyle=numeric,backref=true,maxcitenames=6]{biblatex}
\DefineBibliographyStrings{english}{%
  backrefpage = {page},
  backrefpages = {pages},
}

\usepackage{hyperref}
\hypersetup{
    colorlinks,
    linkcolor={red!50!black},
    citecolor={green!50!black},
    urlcolor={blue!80!black}
}
\usepackage[capitalize]{cleveref}

\DeclarePairedDelimiter{\set}{\{}{\}}
\DeclarePairedDelimiterX{\twobarparen}[2]{(}{)}{#1\;\delimsize\|\;#2}
\DeclarePairedDelimiterX{\inner}[2]{\langle}{\rangle}{#1, #2}
\DeclarePairedDelimiterX{\linner}[2]{\left\langle}{\right\rangle}{#1, #2}
\DeclarePairedDelimiterX{\binner}[2]{\lparen}{\rparen}{#1 \vert{} #2}

\newcommand{\abs}[1]{\vert{} #1 \vert{}}
\newcommand{\labs}[1]{\left\vert{} #1 \right\vert{}}
\newcommand{\norm}[1]{\lVert#1\rVert}
\newcommand{\lnorm}[1]{\left\lVert#1\right\rVert}
\newcommand{\fnorm}[1]{\norm{#1}_\text{F}}
\newcommand{\opnorm}[1]{\norm{#1}_\text{op}}

\newcommand{\dd}{\mathop{}\!\mathrm{d}}

\newcommand{\sign}{\operatorname{sign}}
\DeclareMathOperator{\tr}{Tr} 
 
\DeclareMathOperator{\id}{\operatorname{id}}

\newcommand{\argmin}{\operatorname*{argmin}}
 
\DeclareMathOperator{\Rad}{\widehat{\mathfrak{R}}}
\DeclareMathOperator{\ERad}{\mathfrak{R}}
\DeclareMathOperator{\pack}{\texttt{Pack}}
\DeclareMathOperator{\cover}{\texttt{Cov}}
\newcommand{\vc}[1]{\text{VC}(#1)}

\renewcommand{\emptyset}{\varnothing}
\newcommand{\sym}{\mathsf{S}}
\newcommand{\cyc}{\mathsf{C}}
\newcommand{\orth}{\mathsf{O}}
\newcommand{\SO}{\mathsf{SO}}
\newcommand{\GL}{\mathsf{GL}}

\newcommand{\N}{\mathbb{N}}

\newcommand{\R}{\mathbb{R}}

\DeclareMathOperator{\supp}{\text{supp}}

\let\P\relax
\DeclareMathOperator{\P}{\mathbb{P}}
\DeclareMathOperator{\E}{\mathbb{E}}

\newcommand{\1}[1]{\mathbbm{1}\left\{#1\right\}}
\newcommand{\indep}{\perp\!\!\!\perp}
\newcommand{\iid}{i.i.d.}
\DeclareMathOperator{\unif}{\operatorname{Unif}}
\DeclareMathOperator{\normal}{\mathcal{N}}

\newcommand{\ee}{\mathsf{e}}

\newcommand{\eqas}{\overset{\text{a.s.}}{=}}
\newcommand{\eqdist}{\overset{\text{d}}{=}}

\newcommand{\ifempty}[3]{%
  \if\relax\detokenize{#1}\relax
    #2
  \else
    #3
  \fi
}

\DeclareUnicodeCharacter{2212}{$-$}

\newcommand{\sperp}{{\scaleobj{0.75}{\perp}}}

\newcommand{\xx}{\bm{X}}
\newcommand{\yy}{\bm{Y}}
\newcommand{\zz}{\bm{Z}}
\newcommand{\xxi}{\bm{\xi}}

\newcommand{\qq}{\mathcal{Q}}
\newcommand{\A}{\mathcal{A}}
\newcommand{\D}{\mathcal{D}}
\newcommand{\G}{\mathcal{G}}
\newcommand{\K}{\mathcal{K}}
\renewcommand{\O}{\mathcal{O}}
\renewcommand{\S}{\mathcal{S}}
\newcommand{\X}{\mathcal{X}}
\newcommand{\Y}{\mathcal{Y}}
\newcommand{\Z}{\mathcal{Z}}

\newcommand{\Ohat}{\widehat{\O}}

\newcommand{\Lmu}{{L_2(\mu)}}
\newcommand{\Lmuy}{{L_2(\X, \Y, \mu)}}
\newcommand{\linf}[1]{\norm{#1}_{\infty}}

\newcommand{\Gr}{\mathbb{G}}

\newcommand{\F}{\mathcal{F}}

\newcommand{\vlin}{V_{\text{lin}}}
\newcommand{\wlin}{{W_\text{lin}}}

\newcommand{\proj}{\Phi}
\newcommand{\twine}{\Psi}

\newcommand{\kbar}[1]{\overline{k_{#1}}}
\renewcommand{\H}{\mathcal{H}}
\newcommand{\Hs}{\overline{\H}}
\newcommand{\Ha}{\H_\sperp}
\newcommand{\pJ}{J^\sperp}

\newcommand{\lmin}{\gamma_{\text{min}}}
\newcommand{\lmax}{\gamma_{\text{max}}} 
\newcommand{\Lmumu}{{L_2(\mu\otimes\mu)}}

\newcommand{\bias}{\Lambda_{n, \rho}} 
\newcommand{\popbias}{\Lambda_{\rho}}
\newcommand{\fopt}{f^\star}
\newcommand{\eval}{E_{\xx}}
\newcommand{\fhat}{\hat{f}}

\newcommand{\alg}{\texttt{alg}}
\newcommand{\fgequivalent}{$(\F, \G)$-equivalent}
\newcommand{\T}{\mathsf{T}}

\theoremstyle{definition}
\newtheorem{theorem}{Theorem}

\newtheorem{corollary}[theorem]{Corollary}
\newtheorem{definition}[theorem]{Definition}
\newtheorem{example}[theorem]{Example}
\newtheorem{lemma}[theorem]{Lemma}

\newtheorem{proposition}[theorem]{Proposition}
\newtheorem{remark}[theorem]{Remark}

\newtheorem{assumption}{Assumption}

\newtheorem*{theorem*}{Theorem}
\newtheorem*{assumption*}{Assumption}
\newtheorem*{claim*}{Claim}
\newtheorem*{definition*}{Definition}
\newtheorem*{lemma*}{Lemma}
\newtheorem*{question*}{Question}
\newtheorem*{proposition*}{Proposition}

\AtAppendix{\counterwithin{theorem}{section}}
\numberwithin{equation}{section}
\numberwithin{theorem}{chapter}

\newcommand{\figwidth}{0.75\textwidth}

\colorlet{notegreen}{green!50!black}

\title{Symmetry and Generalisation in Machine Learning}

\author{Hayder Bryn Elesedy}
\college{St Anne's College}

\degree{Doctor of Philosophy}     %the degree
\degreedate{Hilary 2023}         %the degree date

\begin{document}
\emergencystretch 3em

% which spacing to use?
%this baselineskip gives sufficient line spacing for an examiner to easily
%markup the thesis with comments
%\baselineskip=18pt plus1pt

\maketitle

\vspace*{40mm}
\begin{center}
    In memory of my dad, Captain Khalid El-Esedy,
    who would have enjoyed this being submitted at the end of Ramadan.
    Eid Mubarak, Baba.

    And in gratitude to my mum, Linda. Thank you, Mam, for your unconditional(!) love and support.
\end{center}
        % include a dedication.tex file

\begin{romanpages}          % roman page numbering
    {\hypersetup{linkcolor=black}
    \tableofcontents            
    }
    %\listoffigures              % generate and include a list of figures
    \doublespacing
    \section*{Abstract}
This work is about understanding the impact of invariance and equivariance
on generalisation in supervised learning.
We use the perspective afforded by an averaging operator to show 
that for any predictor that is not equivariant,
there is an equivariant predictor with strictly lower test risk on all 
regression problems where the equivariance is correctly specified.
This constitutes a rigorous proof that symmetry, in the form of invariance
or equivariance, is a useful inductive bias.

We apply these ideas to equivariance and invariance in
random design least squares and kernel ridge
regression respectively. This allows us to specify the 
reduction in expected test risk in more concrete settings
and express it in terms of properties of the group, the model
and the data.

Along the way, we give examples and additional results to 
demonstrate the utility of the
averaging operator approach in analysing equivariant predictors.
In addition, we adopt an alternative perspective and formalise the common
intuition that learning with invariant models reduces to a problem in terms
of orbit representatives. The formalism extends naturally to a similar intuition 
for equivariant models.
We conclude by connecting the two perspectives and giving some ideas for future work.
          % include the abstract
    \section*{Acknowledgements}

% Advisors: whom to mention first? 
I would like to thank my advisors Varun Kanade and Yee Whye Teh for their support
throughout this process.
Their guidance has been invaluable.
Their diverse expertise has broadened my perspective on the field
and their judicious nudges have made a world of difference.
As their student I had both of Berlin's concepts of liberty: % had liberties
freedom from any pressure to publish or produce results
as well as support to explore whatever ideas I found interesting.
Whatever topic I brought to our meetings, I would always get their honest 
and engaged feedback.

% OxCSML and ML Virgins
I have benefitted from the rich intellectual environment cultivated in
Yee Whye's group at OxCSML, which has exposed me to areas
outside of my research and provided inspiration for new questions.
Indeed, the work in this thesis is an attempt to answer questions
that arose during the regular reading groups.
A special mention goes to Sheheryar Zaidi, Bobby He and Michael Hutchinson,
who hold a humbling array of talents.
It has been a pleasure to bring down the average of our year group.
% DELVE?

There are many others to thank for my time at Oxford,
and probably more still that I can't bring to mind at this moment.
Martin Lesourd, John Mittermeier and Joel Hart for providing a counterbalance at home.
Panagiotis Tigas for a rubber duck code review that saved me from despair.
Tim Rudner and Adam Goli\'{n}ski for advice during the early stages of my PhD.
Wendy Poole, superhero of the Autonomous Intelligent Machines and Systems CDT,
for the million helpful things she has done.
Sheheryar Zaidi for his contributions to \parencite{elesedy2021provably} and enlightening conversations.
Shahine Bouabid and Jake Fawkes for notifying me of an error in \parencite{elesedy2021kernel}
and their subsequent attempts to fix it.
Benjamin Bloem-Reddy for helpful discussions on the topic of this thesis and related areas,
including one that lead to \cref{sec:connection}. 

Few people I'm connected to through Oxford know this, but the route to this point hasn't 
been straightforward.
In that respect, I intend the submission of this thesis to be the end of the beginning.
I stopped going to school as a teenager.
I left at around 16 with few qualifications and almost took a different trajectory. 
I ended up going back to education and since then I've just been putting one
foot in front of the other as best I can.
Encouragement from others has been a precious resource 
and I have countless people to thank.

As it happens, I can't remember the first guy's name, but he left me sufficiently 
humiliated after months of sitting unemployed at my mum's that I found a job and
later enrolled in a sixth form college.
When I got to college, despite what I thought I was capable of, I knew nothing.
Early on I was advised to drop A-Level maths, the reason being that if I worked really 
hard I might get a C grade, but it was a long shot.

I was thinking of doing so, but then Mike Knowles stepped in with some critical 
guidance: do I want to give up when I receive a challenge, or actually try?
This was the pivotal moment.
I kept going with the maths A-Level and it turned into my favourite subject.
I was well supported by my teacher Mr.~Jones, who left me to explore the subject at my own pace.
Around this time Juan Bercial and Peter Giblin took time to show me exciting ideas in mathematics,
while Rob Lewis and Glenn Skelhorn nurtured my interest in philosophy and provided useful feedback.

I ended up at Cambridge because I attended a summer school 
for under-represented students run by Geoff Parks.
I learned about the summer school after bumping into my economics teacher 
one evening in the college corridors (another name I can't remember, sorry!). 
She recommended that I apply, I shrugged and she submitted on my behalf
what would have had to have been quite a convincing application.
At the summer school I was encouraged to apply to Cambridge.
I did, and I was lucky enough to get in.

At Cambridge I was a student of Stephen Siklos, who invested a lot of time and energy in
me. He took most of my supervision sessions one on one and, for whatever reason,
was insistent that there was something to be made of me academically.
I suspect Stephen's recommendation letter was key to my acceptance to Oxford.
From what I can tell, he went out of his way to help many other students as well. 

Along the way, I have received lessons, help and encouragement from others including
Neil Smith,
Carola-Bibiane Sch\"{o}nlieb, 
Emanuel Malek,
Tom M\"{a}dler,
Jean-Gabriel Prince,
Filippo Altissimo,
Haydn Davies,
Alex Bakker,
Mike Osborne,
Neil Lawrence,
Federico Vaggi,
David Duvenaud,
Mihaela Rosca and 
Marcus Hutter.

%% Personal
There are countless people for whom I'm grateful in my personal life, but I won't go into details.
It doesn't feel like the forum.
I have great friends and during difficult times the fun has kept me going.
A loving and supporting family with a great sense of humour.
Welcoming (de facto) in-laws who feed me and treat me as one of their own.
And, saving the best until last,
a long-suffering other half, Issy, 
who has been carrying my heart for more than a decade.
She listens to me, supports me, and even laughs at my jokes.
I couldn't ask for more.

This DPhil was supported by funding from the UK EPSRC CDT in Autonomous Intelligent
Machines and Systems (grant reference EP/L015897/1).
  % include an acknowledgements.tex file
\end{romanpages}
\doublespacing

\chapter{Introduction}

\section{Motivation}

We will study how invariance and equivariance affect generalisation
in supervised learning.
Let $\G$ be a group acting on sets $\X$ and $\Y$,
then $f: \X \to \Y$ is invariant if
$f(gx) = f(x)$ and equivariant if 
$f(gx) = gf(x)$, each holding for all $g\in\G$ and $x\in\X$.
Invariance is the special case of equivariance where the action of $\G$ on 
$\Y$ is trivial.
In this work, the word \emph{symmetry} specifically refers
to some form of invariance or equivariance.

Supervised learning is the science of extrapolation from labelled data.
The basic task is as follows: given a sequence of
input-output pairs $(x_1, y_1), \dots, (x_n, y_n)$ 
generated by some unknown, possibly stochastic procedure,
predict unseen values $(x, y)$ generated by the same procedure.
Fundamentally, it is a problem of inductive inference.

More formally, let $X$ and $Y$
be random elements of $\X$ and $\Y$ respectively whose distributions are
unknown.
We call the sequence $(x_1, y_1), \dots, (x_n, y_n)$
the \emph{observations} and assume that they are, respectively, instances
of the random variables $(X_1, Y_1), \dots, (X_n, Y_n)$.
The tuple $((X_1, Y_1), \dots, (X_n, Y_n))$ is called the \emph{training sample}
and individual elements of the training sample are \emph{training examples} or
just \emph{examples}.
We assume the training examples are distributed independently and identically to $(X, Y)$.
The word $\emph{data}$ may refer to either fixed observations or the training sample
depending on the context.

We formulate supervised learning as the problem of finding, given the observations, 
a minimiser over $f: \X \to \Y$ of 
\[
    R[f] = \E[\ell(f(X), Y)]
\]
where $\ell: \Y \times \Y \to \R_+$ is the \emph{loss function}
which measures the quality of predictions.
We will refer to $R$ as the \emph{risk function} and to $R[f]$ as the
\emph{risk of $f$}; when the distinction isn't needed
we may refer to each of these as the \emph{risk}.
We call the candidate functions $f$ \emph{predictors} or
\emph{models}. 
A \emph{learning algorithm} is a map from training samples
to predictors.

Without access to the distribution of $(X, Y)$ it is not possible
to minimise the risk directly and, in any case, finding a minimiser of $R$
could be computational infeasible. In practice, an approximate solution is acceptable.
In the same vein, theoretical study is often specialised to certain relationships
between $X$ and $Y$ or the minimisation constrained to certain classes of predictors.

From a mathematical perspective, the problem of finding a predictor with 
small risk is critically underdetermined. 
Judicious choice of a predictor requires extrapolating from
the finite number of observations to the general relationship
between $X$ and $Y$.
The term \emph{inductive bias} refers loosely to a mode of extrapolation.
For instance, linear predictors share an inductive bias because they
extrapolate from the training sample in a conceptually similar fashion. 

Symmetry is also a form of inductive bias. Along an orbit of $\G$,
invariant models are constant while the values of equivariant models 
are related by the group action.
Symmetry has emerged as a popular method of incorporating domain knowledge
into models \parencite{cohen2016group,kondor2018clebsch,%
cohen2019general,weiler2019general,finzi2020generalizing}
and these models have applications in many areas. For instance, where
symmetry is known to be a fundamental property of the system
such as quantum chemistry \parencite{pfau2020ab}, or where
arbitrary experimenter choices or data representation 
give undue privilege to a specific coordinate system 
such as medical imaging or protein folding
\parencite{winkels2018d,jumper2021highly}.
The purpose of this thesis is to understand, insofar as it improves generalisation,
whether symmetry is a good idea.

We will mostly consider the relative performance of predictors or learning algorithms 
and the risk provides a comparator.
In particular, the statement 
$f$ \emph{generalises better than} $f'$ means that $R[f] \le R[f']$ and
generalising \emph{strictly} better means a strict inequality in the same
direction. 
We make the same comparison for learning algorithms, for which we define
the risk to be the risk of the returned predictor viewed 
as a function of the training sample.
In this case the risk is stochastic and the comparison will be in expectation.
The risk of a predictor or learning algorithm, the extent to which it
\emph{generalises}, is the only measure of quality we consider.
Although important, we do not consider other aspects such as 
interpretability or computational efficiency.

The theoretical study of generalisation
can be considered as a narrow
form of mathematical epistemology,
in that it offers a quantitative formulation and analysis of the
problem of induction.
More practically, it has the potential to offer performance guarantees 
that improve the reliability of machine intelligence.
In addition, theoretical analyses can provide conceptual schemes and intuitions that
lead to new methods.

Closely related to generalisation is learning. 
A learning algorithm with range $\F$
is said to \emph{learn} the class of functions $\F$ if 
$\exists m:(0, 1)^2 \to \N$ such that
$\forall \epsilon, \delta \in (0, 1)$
if $n \ge m(\epsilon, \delta)$ 
then for all distributions of $(X, Y)$,
with probability at least $1 - \delta$ over the training sample,
the output $\hat{f}$ of the learning algorithm satisfies
\[
    R[\hat{f}] \le \inf_{f\in\F} R[f] + \epsilon.
\]
The pointwise minimum over all $m$ that satisfy the above is called the
\emph{sample complexity} of the learning algorithm.

The above model of learning is based on
PAC learning, which was originally proposed by \textcite{valiant1984theory}.
The variation we give is due to \textcite{haussler1992decision}
and is known as agnostic PAC learning, because it is agnostic
as to the relationship between $X$ and $Y$ \parencite{kearns1994toward}.
See \parencite{shalev2014understanding,mohri2018foundations} for further
information and bibliographic remarks.

If the sample complexity of learning is established then one has non-asymptotic
control of the risk of the learning algorithm.
In the case of binary classification,
the sample complexity of learning is governed by a combinatorial measure of 
complexity of $\F$ called the VC dimension, originally from
\textcite{vapnik1971uniform}.
The same idea holds for learning in other settings such as regression,
just with different complexity measures, for instance the Rademacher complexity
\parencite{koltchinskii2000rademacher}
or covering numbers
\parencite{cucker2002mathematical}.
See \parencite{martin1999nnl} for more.

In the case of symmetry, it seems natural to try to derive sample complexity
upper bounds that reduce when $\F$ has the right symmetry. 
All prior works on the generalisation of invariant
and equivariant models take some form of this approach
(see \cref{sec:lit-review-theory} for a discussion)
and we also give some results of this flavour in 
\cref{prop:fa-rademacher,thm:sample-complexity}.\footnote{%
As far as we are aware, the only other work that doesn't take 
this approach is by \textcite{mei2021learning},
which appeared on arXiv only four days after the first work of this thesis.
The topic is similar to \cref{sec:kernel-generalisation}, but the approach
is quite different. A comparison is given in \cref{sec:other-work-comparison}.%
}
However, the main goal of this thesis is to provide theory that aides the
practitioner in deciding whether to incorporate symmetry (either engineered
or learned) into their model. For this purpose the aforementioned results
are suggestive, but insufficient.

We would like to understand the \emph{generalisation gap}
\[
    \Delta(f, f') = R[f] - R[f']
\]
between predictors $f$ and $f'$. 
Specifically, we are interested in the case where $f'$ is equivariant and $f$ is not.
If $\Delta(f, f') > 0$ then $f'$ is preferable to $f$. 
When $f$ and $f'$ are the outputs of learning algorithms the generalisation
gap is a function of the training sample so is stochastic.
In this setting, the complexity based results we mention above 
provide tail estimates for $R[f]$ and $R[f']$.
However, unless these estimates turn out to be exact
it is difficult even to tell the sign of $\E[\Delta(f, f')]$.
Ultimately, the techniques of statistical learning theory are too general
so we must take a different approach. We give an outline in the next section.
 
\section{Overview}
The technical setting and assumptions are described in \cref{sec:setup}.
Below we give an informal overview of our main results without reference
to these conditions. Throughout the work there are many examples and additional
results to illustrate the utility of the approach.
We make use of some standard facts which we give in \cref{chap:useful-results}.
We review related literature in \cref{chap:lit-review}.

\cref{chap:general-theory-i} contains some general observations.
Let $\G$ be a group and let 
$f: \X \to \Y$ be a function between two sets on which $\G$ acts.
Define the operator 
\[
    \qq f (x) = \int_\G g^{-1} f(gx) \dd g
\]
which takes any function and makes it equivariant.
By exploring the properties of $\qq$, we establish in \cref{lemma:l2-decomposition}
that any function can be written as 
\[
    f = \bar{f} + f^\sperp
\]
where $\bar{f}$ is equivariant, $\qq f^\sperp = 0$ and, crucially, the two terms are 
$L_2$-orthogonal as functions.

This turns out to be a useful observation. 
In particular, it means
that on any regression task with squared-error loss and an equivariant target
the generalisation gap $\Delta(f, \bar{f})$ is exactly the squared $L_2$-norm
of $f^\sperp$. If $f$ is not equivariant then this is strictly positive.
In other words, for any predictor $f$ that is not equivariant
there is an equivariant predictor $f'$ with strictly lower risk
on all regression problems with squared-error loss and an equivariant target.
Another view on this is that it gives a lower bound for not using an equivariant predictor.
All of this applies equally to invariance.

These insights are applied in \cref{chap:linear},
where the main result 
quantifies the expected generalisation benefit of equivariance in a linear model.
\Cref{thm:equivariant-regression} concerns the generalisation gap
$\Delta(f, f')$ in the case that $f: \R^d \to \R^k$ is the minimum-norm 
least squares predictor
and $f' = \qq f$ is its equivariant version. 

Let $\G$ act via orthogonal representations $\phi$ and $\psi$ on
inputs $X \sim \normal(0, I_d)$ and outputs $Y = h(X) + \xi \in \R^k$ respectively,
where $h: \R^d \to \R^k$ is an equivariant linear map, $\E[\xi]=0$ and
$\E[\xi^2]=\sigma^2<\infty$.
Let 
\[
    \binner{\chi_\psi}{\chi_\phi}= \int_\G  \tr(\psi(g))\tr(\phi(g))\dd g
\]
denote the scalar product of the characters of the representations (which are real). 
We show in \cref{thm:equivariant-regression} that
the generalisation benefit of enforcing equivariance in a linear model is given by
\[
    \E[\Delta(f, f')] 
    = \mathcal{E}(n, d) + \sigma^2 r(n, d)(dk - \binner{\chi_\psi}{\chi_\phi}) 
\]
where
\[
    r(n, d) = \begin{cases}
        {\frac{n}{d(d - n - 1)}} & n < d- 1\\
        (n - d - 1)^{-1} & n > d + 1\\
        \infty & \text{otherwise}
    \end{cases}
\]
and $\mathcal{E}(n, d) \ge 0 $ is the generalisation gap of the
corresponding noiseless problem, specified exactly in \cref{thm:equivariant-regression},
which vanishes when $n \ge d$.
The divergence at the interpolation threshold $n \in [d-1, d+1]$ 
is consistent with the literature on double descent \parencite{hastie2022surprises}. 

It's worth emphasising that if $f'$ is any other predictor such that
$\E[R[f']] \le \E[R[\qq f]]$ then the above result holds as a lower bound on
$\E[\Delta(f, f')]$. In particular, this means that the above
result applies to the case where $f'$ is generated by
transforming the features before training such that the least squares
estimate is automatically equivariant, provided that the resulting
predictor generalises at least as well as $\qq f$ in expectation over
the training sample.

The quantity $dk - \binner{\chi_\psi}{\chi_\phi}$ 
represents the significance of the group symmetry to
the task. The dimension of the space of linear maps $\R^d \to \R^k$ is $dk$,
while $\binner{\chi_\psi}{\chi_\phi}$ is the dimension of the space of equivariant linear maps. 
We will see later that the quantity $dk - \binner{\chi_\psi}{\chi_\phi}$ 
represents the dimension of the space of linear maps that vanish under $\qq$.
It is through the dimension of this space that the symmetry in the task
controls the generalisation gap.
Although invariance is a special case of
equivariance, we find it instructive to
discuss it separately. In \Cref{thm:invariant-regression} we provide
a result that is analogous to \Cref{thm:equivariant-regression} for invariant predictors,
along with a separate proof.

In \cref{chap:kernels} we adapt and extend these results to kernel methods. 
We define the operator
\[
    \O f (x) = \int_\G f(gx) \dd g
\]
which takes any function and makes it invariant.
Since $\O$ is a special case of $\qq$, we have the decomposition
$f = \bar{f} + f^\sperp$ with $\bar{f}$ invariant, $\O f^\sperp=0$
and the terms being $L_2$-orthogonal. In \cref{thm:kernel-invariance} we 
study the generalisation gap $\Delta(f, \O f)$ for kernel ridge 
regression on an invariant target. 
The comments made above about the use of $\qq f$ as a comparator 
apply here to $\O f$.

Let $X$ be invariant in distribution, so $X \eqdist gX$ for all $g\in\G$, and
set $Y = f^*(X) + \xi$ with $f^*$ invariant and $\E[\xi] =0$,
$\E[\xi^2] = \sigma^2 < \infty$.
Let $f$ be the solution to kernel ridge
regression with kernel $k$ and regularisation parameter $\rho > 0$ on $n$ \iid~training
examples $((X_i, Y_i):i=1, \dots, n)$ each distributed identically to and independently
from $(X, Y)$.
In \cref{thm:kernel-invariance}, we find that
\[
    \E[\Delta(f, \O f)]
    \ge
    \mathcal{E}_k(n, \rho)
    + \sigma^2 \frac{\norm{(\id - \O)k}_{\Lmumu}^2 }{(\sqrt{n}M_k + \rho/\sqrt{n})^2}
\]
where for any $j: \X\times\X\to\R$
\[
    \norm{j}^2_\Lmumu = \int_\X j(x, y)^2 \dd\mu(x)\dd\mu(y),
\]
$M_k = \sup_x k(x, x) <\infty$ and $\mathcal{E}_k(n, \rho) \ge 0$ 
is the generalisation gap for the corresponding noiseless problem.
By considering the linear kernel
we study how this result relates to the result for invariance in \cref{chap:linear}.

Further, we show that under mild additional conditions on $\X$, $\Y$ and $\rho$,
$\mathcal{E}_k(n, \rho) \to 0$ as $n\to\infty$ provided the kernel $k$
satisfies the identity
\[
    \int_\G k(gx, y)\dd g = \int_\G k(x, gy)\dd g
\]
for all $x, y\in\X$. Assuming this condition holds, we derive an independent
result about the structure of the RKHS $\H$. In particular,
\cref{thm:rkhs-decomposition} says that the above condition on the kernel
implies the orthogonal decomposition $\H = \O\H \oplus (\id - \O)\H$
where the orthogonality is now with respect to the inner product on
$\H$. 

In \cref{chap:general-theory-ii} we take a different approach, 
making use of the observation
that an invariant function can be specified by its values on one representative
from each orbit of $\X$ under $\G$.
We show rigorously how learning 
a class of invariant predictors is equivalent to
learning in a reduced
problem in terms of orbit representatives
and extend this framework to provide a new intuition for learning equivariant predictors.
In addition, we show how to use these equivalences to derive a sample complexity
bound for learning invariant/equivariant classes with
empirical risk minimisation.

We conclude in \cref{chap:outlook}
with a relation of the orbit averaging viewpoint on invariance
in \cref{chap:general-theory-i} to the orbit representative viewpoint 
in \cref{chap:general-theory-ii}, applications to neural networks,
connections to other works and, finally, 
some sugggestions for future work.

\section{Authorship}
The work in this thesis is based on 
\parencite{elesedy2021provably,elesedy2021kernel,elesedy2022group}.
The majority of content has been revised or generalised, sometimes
substantially, and some new results are given.
\cref{chap:general-theory-i,chap:linear,sec:neural-networks} are based on
\parencite{elesedy2021provably}, \cref{chap:kernels} is based on \parencite{elesedy2021kernel}
and \cref{chap:general-theory-ii} is based on \parencite{elesedy2022group}.

All original contributions of the thesis 
were discovered and proved 
by the author unless explicitly stated otherwise.
Where possible, Zaidi's contributions to \parencite{elesedy2021provably} 
are cited explicitly. Otherwise, his contributions
to \parencite{elesedy2021provably} were through discussions, proof reading
and presentation.
In addition, Zaidi suggested the connection to test time augmentation 
described in \cref{sec:tta}.

A list of other work completed during this DPhil is given in 
\cref{chap:other-phd-work}.

\section{Preliminaries}
We assume familiarity with the basic notions of group theory
including the definitions of a group action and a linear representation.
The reader may consult \parencite[Chapters 1-4]{wadsleyrep2012,serre1977linear}
for background.
We run through some results that we use often in \cref{sec:common-results}.
We provide some background definitions and technical material 
in \cref{sec:background-theory}.

\subsection{Setup, Assumptions and Technicalities}\label{sec:setup}
In this section we outline our setup and technical conditions which,
unless stated otherwise, are assumed throughout. 
Additional definitions and assumptions are given as needed.
Technical conditions are chosen so as to vary the least between results;
the reader interested in more general settings is encouraged to
inspect the proofs. That said, the conditions we impose are quite general.

\subsubsection{Spaces}
There will be a background probability space $(\Omega, \S_\Omega, \P)$ that is
assumed to be rich enough to support our analysis. 
The input and output spaces will be $\X$ and $\Y$ respectively,
where $\X$ is a non-empty Polish space and
$\Y$ is $\R^k$ with an inner product $\inner{\cdot}{\cdot} :\R^k \times \R^k \to \R$,
induced norm $\norm{\cdot}$ and corresponding topology. 
Sometimes $k=1$ or this inner product is the Euclidean one, but this will be specified.
All $\sigma$-algebras will be Borel.
In particular this makes $(\X, \S_\X)$ and $(\Y, \S_\Y)$
standard Borel spaces, but we won't make direct use of this level of detail.
Unless stated otherwise, $X$ and $Y$ will be random elements of $\X$ and $\Y$ respectively.

\subsubsection{Group and Action}
Let $\G$ be a measurable, second countable, Hausdorff and compact topological group.%
\footnote{The set of compact groups covers the majority of symmetries in machine learning,
including
all finite groups (such as permutations or reflections), many continuous
groups such as rotations or translations on a bounded domain (e.g., an image)
and combinations thereof.}
Let the Haar measure on $\G$ be $\lambda$, normalised so that $\lambda(\G)=1$.
This is the unique invariant probability measure on $\G$.
%Normalised so that
%$\lambda(\G) = 1$, which is possible because $\lambda$ is a Radon measure and
%$\G$ is compact.
%This measure 
%for any measurable subset $A \subset \G$ and for any $g\in \G$,
%$\lambda(gA) = \lambda(Ag) = \lambda(A)$ \parencite[Theorem 2.27]{kallenberg2006foundations}. 
We assume that $\G$ has a measurable action $\phi$ on $\X$ and
measurable representation $\psi$ on $\Y$.
By measurable we mean that $\phi: \G \times \X \to \X$ is a measurable map 
and the same for $\psi$.
We will assume that $\psi$ is unitary with respect to $\inner{\cdot}{\cdot}$, by
which we mean that 
$
    \inner{\psi(g) a}{\psi(g)b} = \inner{a}{b}
    $
$\forall a, b \in \R^k$ and $\forall g \in \G$.
Notice that this implies $\inner{\psi(g)a}{b} = \inner{a}{\psi(g^{-1})b}$.
If $\inner{\cdot}{\cdot}$ is the Euclidean inner product, then this is the usual
notion of an orthogonal representation (one for which the $\psi(g)$ is always
an orthogonal matrix).
Any inner product can be transformed to be such that $\psi$ is unitary using
the Weyl trick
$
    \inner{a}{b} \mapsto \int_\G \inner{\psi(g)a}{\psi(g)b} \dd\lambda(g)
    $.
The \emph{character} of a representation $\psi$ is defined
by $\chi_\psi(g) = \tr(\psi(g))$. The inner product of characters
is defined by 
\[
    \binner{\chi_1}{\chi_2} = \int_\G \chi_1(g)\chi_2(g) \dd\lambda(g).
\]
This definition typically appears with a complex conjugate, but it's not
needed because all the representations we encounter are real.
The inner product of the characters of two finite-dimensional
real representations of a compact group is equal to the dimension of the space
of linear maps that are equivariant with respect to these representations,
e.g., see \parencite[Theorem 3.34]{adams1982lectures}.
% This reference is for Lie groups, but I (lightly) checked that the results still follow for 
% arbitrary compact groups

\subsubsection{Invariance, Equivariance and Symmetry}
A function $f: \X \to \Y$ is \emph{$\G$-invariant} if
$f(\phi(g)x) = f(x)$ 
and is \emph{$\G$-equivariant} if
$f(\phi(g)x) = \psi(g)f(x)$,
each holding for all $x\in\X$ and for all $g\in\G$.
Invariance is the special case of equivariance where $\psi$ is the
\emph{trivial representation}, i.e., $\psi(g)$ is the identity for all $g\in\G$.
A measure $\mu$ on $\X$ is \emph{$\G$-invariant}
if for all $g \in \G$ and any $\mu$-measurable $B \subset \X$ 
the pushforward of $\mu$ by the action $\phi$ equals $\mu$, i.e., $(\phi(g)_* \mu)(B) = \mu(B)$.
This means that if $X \sim \mu$ then $\phi(g) X \sim \mu$ for all $g\in\G$.
We say that $X$ is \emph{$\G$-invariant in distribution} or just \emph{$\G$-invariant}
if $X \eqdist gX$ for all $g\in\G$.

We will often drop the $\G$- when the group is clear from the context and just say
invariant/equivariant.
We will often also drop $\psi$ and $\phi$, so
the quantity $g^{-1}f(gx)$ should be interpreted as $\psi(g^{-1})f(\phi(g)x)$, and so on.
We sometimes use the catch-all term \emph{$\G$-symmetric} to describe an object that is invariant
or equivariant.

\subsubsection{Function Space}
Throughout, $\mu$ will be a $\G$-invariant probability measure on $(\X, \S_\X)$.
We consider $\Lmuy$, which we define to be the Hilbert space of equivalence classes of
functions $f: \X \to \Y$ such
that $\norm{f}_\mu < \infty$ where the norm is induced by the following inner product
\[
    \inner{f}{h}_\mu = \int_\X \inner{f(x)}{h(x)} \dd \mu(x).
\]
Equality in $\Lmuy$ is defined $\mu$-almost-everywhere.
%When we're safe to do so, we will ignore the distinction between elements of
%$\Lmuy$ and the functions they contain.
This space is general enough to cover
pretty much any predictor used in machine learning.
In the case $\Y=\R$ we consider the usual $L_2$-space which we write $\Lmu$,
where the inner product on $\Y$ is just multiplication.

\subsubsection{Averaging Operator}
The averaging operator $\qq$ will be central to much of this work.
It has values
\[  
    \qq f (x) = \int_\G g^{-1}f (gx)\dd\lambda(g),
\]
where $\lambda$ is the normalised Haar measure on $\G$ with $\lambda(\G) = 1$.
The operator $\qq$ can be used to transform any function into an equivariant
function.%
\footnote{The operator $\qq$ bears similarity to the
\emph{twirl operator} in quantum computing,
see \parencite[Section 2.1.1]{meier2018randomized}
and \parencite[Section VII.B]{ragone2022representation}.}
Due to the compactness of $\G$, we can change
variables $g \mapsto g^{-1}$ above and view $\qq$ as averaging the action
of $\G$ on $\Lmuy$ defined by $f \mapsto g \circ f \circ g^{-1}$.
Our developments will apply equally to the operator
\[
    \O f(x) = \int_\G f(gx) \dd \lambda (g)
\]
which is the special case of $\qq$ in which $\psi$ is the trivial representation.
In this case the action of $\G$ on $\Lmuy$ defined above is akin to the 
regular representation.
The operator $\O$ enforces invariance, a special case of equivariance.
In other works both $\O$ and $\qq$ are often referred to as Reynolds operators.

\subsection{Additional Notation}
The sets $\R$, $\mathbb{Z}$, $\N$ and $\R_+$ denote the
reals, integers, naturals and non-negative reals respectively.
We use $\circ$ for function composition $(f \circ h)(x) = f(h(x))$
and we write $\ee = \exp(1)$.
For functions $f, h:\N\to\R_+$, $f = \omega(h)$ means
$\exists C>0$ $\exists m\in\N$ such that $\forall n> m$ $f(n)>Ch(n)$
and $f = O(h)$ means $f(n) \le Ch(n)$ under the same quantifiers.

We write $X \sim \mu$ to mean that $X$ has distribution $\mu$.
For probability spaces $(S, \S, \mu)$ and $(T, \mathcal{T}, \nu)$
the product measure is denoted by $\mu \otimes \nu$, see \cref{thm:bg-fubini}.
For random variables $A$ and $B$ independence is written $A\indep B$,
equality in distribution is written $A\eqdist B$ and almost sure
equality is written $A \eqas B$.
We write \iid~to stand for independent and identically distributed.

We will use the Einstein summation convention, in which repeated
indices are implicitly summed over.
The Kronecker tensor is written as
$\delta_{ij}$, which is $1$ when $i = j$ and $0$ otherwise.
We write $\Gr_n(\R^d)$ for the Grassmannian manifold of subspaces of dimension $n$ in $\R^d$.

We write $\id$ for the identity operator
and any linear operator $A$ we write its adjoint as $A^*$.
We use $\norm{\cdot}_2$ for the Euclidean norm of vectors,
$\norm{\cdot}_1$ for the sum of magnitudes of components
and $\norm{\cdot}_\infty$ for the component-wise max magnitude. 
On a function $f: \X \to \R^d$, $\linf{f} = \sup_{x\in\X} \norm{f(x)}_\infty$.

We write $I$ for the identity matrix, sometimes with a subscript to indicate
the dimension of the space on which it acts.
We write $\R^{m\times n}$ for the set of all real matrices
with $m$ rows and $n$ columns.
For any matrix $A \in \R^{n\times n}$
we define $\norm{A}_2 = \sup_{x \in \R^n} \frac{\norm{Ax}_2}{\norm{x}_2}$,
which is the operator norm induced by the Euclidean norm.
In general, the operator norm of any operator between normed spaces will be written
with $\opnorm{\cdot}$.
For any symmetric matrix $A$, we denote by $\lmax(A)$ and $\lmin(A)$ the largest and smallest
eigenvalues of $A$ respectively.
For any matrix $A$ we write 
$A^+$ for the Moore-Penrose pseudo-inverse
and $\fnorm{A} = \sqrt{\tr(A^\top A)}$ for the Frobenius norm.

Some notation for specific groups: $\cyc_m$ and $\sym_m$ are, respectively,
the cyclic and symmetric groups on $m$ elements,
while $\orth_m$ and $\SO_m$ are the $m$-dimensional orthogonal and special
orthogonal groups respectively. The group of $m\times m$ invertible real matrices
is written $\GL_m$.

\subsection{Commonly Used Results}\label{sec:common-results}
We will make use of the results in this section throughout the work.

\paragraph{The inclusion $\Lmu \subset L_1(\mu)$}
\begin{theorem}[{\parencite[Theorem 2]{villiani1985LpLq}}]%
    \label{thm:bg-LpLqinclusion}
    Let $(\X, \S_\X, \mu)$ be a measure space with $\mu \ge 0$.
    Let $A_\infty = \set{A \in \S_\X: \mu(A) < \infty}$.
    Then $\sup_{A\in A_\infty} \mu(A)< \infty$ is equivalent to
    $L_p(\mu) \subset L_q(\mu)$ for all $p > q$.
\end{theorem}
In our case $\mu$ is a probability measure, so $\Lmu \subset L_1(\mu)$.
We will use this fact frequently, in particular when applying Fubini's theorem.
  
\paragraph{Fubini's theorem}
\begin{theorem}[{Fubini's theorem \parencite[Theorem 1.27]{kallenberg2006foundations}}]%
    \label{thm:bg-fubini}
    For any $\sigma$-finite measure spaces $(S, \mathcal{S}, \nu_S)$ and
    $(T, \mathcal{T}, \nu_T)$ there exists a unique product measure 
    $\nu = \nu_S \otimes \nu_T$ on $(S \times T, \mathcal{S}\otimes\mathcal{T})$ such that 
    \[
        \nu (A \times B) = \nu_S(A)\nu_T(B) 
        \quad 
        \forall A \in \mathcal{S},\;
        B\in\mathcal{T}.
    \]
    Moreover, for any measurable $f: S\times T\to \R_+$ 
    \[
        \int_{S\times T} f(s, t) \dd \nu(s, t)
        = \int_S \dd\nu_S(s)\int_T f(s, t) \dd\nu_T(t)
        = \int_T \dd\nu_T(t)\int_S f(s, t) \dd\nu_S(s).
    \]
    The above remains valid for any $\nu$-integrable 
    $f: S\times T \to \R$.
\end{theorem}
Both $\mu$ and $\lambda$ are probability measures so are $\sigma$-finite.
When the integrand can be negative we must verify integrability.
In doing this, we will might apply Fubini's theorem to a non-negative 
function, but most of the time we will use \cref{thm:bg-LpLqinclusion}.

\paragraph{{Change of variables $g \mapsto g^{-1}$}}
% This is imperfect IMO, I don't understand it well enough.
\begin{theorem}[{\parencite[Corollary 2.28 and Proposition 2.31]{folland2016course}}]
    Let $\G$ be a measurable compact group with Haar measure $\lambda$, then 
    $\dd\lambda(g) = \dd\lambda(g^{-1})$.
\end{theorem}   

\subsection{Background Theory}\label{sec:background-theory}
We list some background definitions and results.
For more details see
\parencite{rudin1987realandcomplex,kallenberg2006foundations}.

\subsubsection{Topology}
A topological space is \emph{separable} if it contains a countable dense subset.

A metric space is \emph{complete} if every Cauchy sequence in the space converges
to a limit in the space.

A \emph{Polish space} is a separable topological space that admits a metric with
respect to which it is complete.

Let $(T, \tau)$ be a topological space, then $B \subset \tau$
is a \emph{base} for $\tau$ if any element of $\tau$ is the union
of elements of $B$.

A topological space is \emph{second countable} if its topology has a countable base.

A topological space is \emph{Hausdorff} if all pairs of distinct points
have disjoint open neighbourhoods.

A subset of a topological space is \emph{compact} if every open cover has a finite sub-cover
and
it is \emph{locally compact} if every point has a compact neighbourhood.
Clearly, any compact set is locally compact.

A \emph{topological group} is a group $\G$ equipped with topology such that
the group operations are continuous. For instance $g\mapsto g^{-1}$ is continuous
and $(g, h) \mapsto gh$ is continuous when $\G^2$ has the product topology.

\subsubsection{Measure Theory}
Let $f: A \times B \to C$ be a function. Then a \emph{$B$-section} of $f$ is a 
function $f_b: A \to C$ for some $b\in B$ with $f_b(a) = f(a, b)$.
An \emph{$A$-section} is defined similarly.

Let $(S, \S, \nu)$ be a measure space. We say $\nu$ is \emph{$\sigma$-finite}
if $S$ can be written as the disjoint union of a countable family of elements
of $\S$, each of which has finite measure.
All probability measures are $\sigma$-finite.

The \emph{Borel $\sigma$-algebra} on a topological space is the $\sigma$-algebra
generated by the topology.
A \emph{Borel measure} is a measure defined on a Borel $\sigma$-algebra.

A \emph{measurable group} is a measurable space $(\G, \S_\G)$ where
$\G$ is a topological group and $\S_\G$ is the Borel $\sigma$-algebra.
In this case the group operations are
measurable when $\G$ is locally compact, second countable and Hausdorff
\parencite[p.~39]{kallenberg2006foundations}.

A measure $\lambda$ on $(\G, \S_\G)$ is \emph{left-invariant} if
$\lambda(gA) = \lambda(A)$
for all $A \in \S_\G$ and for all $g\in \G$.
The definition of \emph{right-invariant} is the same but with $\lambda(Ag) = \lambda(A)$.
The measure $\lambda$ is \emph{invariant} if it is both left-invariant and right-invariant.

A \emph{Radon} measure is one that is finite on any compact measurable set.

\begin{theorem}[{Haar measure \parencite[Theorem 2.27]{kallenberg2006foundations}}]%
    \label{thm:bg-haar-measure}
    On any locally compact, second countable and Hausdorff measurable group $\G$
    there exists, uniquely up to normalisation, a left-invariant
    Radon measure $\lambda$ with $\lambda \ne 0$. If $\G$ is compact then $\lambda$ 
    is also right-invariant.
\end{theorem}   

In our setting $\G$ will be compact so we can normalise $\lambda$ to be the
unique invariant probability measure.

\chapter{Related Literature}\label{chap:lit-review}
In this work we are concerned with predictors that are 
equivariant or invariant functions. In particular we are
interested in a theoretical analysis of their generalisation.
We discuss works in this area first in \cref{sec:lit-review-theory},
then move on to an outline of various symmetric models and their
applications in \cref{sec:lit-review-models}.
We end this chapter with \cref{sec:lit-review-other-symmetry},
in which we point to some other notions of symmetry in machine learning.

We note that group symmetry plays a role in the nearby field of statistics.
We do not discuss this, but instead refer the reader to 
\parencite{eaton1989group,bloem2020probabilistic,bloem2023lecture}
and references therein.
Finally, in writing this literature review we found results that
are either related to or are special cases of the results in this thesis.
We give a comparison in \cref{sec:other-work-comparison}.

\section{Symmetry and Generalisation}\label{sec:lit-review-theory}
The first general theoretical justification for invariance
of which we are aware is from \textcite{fyfe1992invariance,abu1993hints},
which,
roughly, states that enforcing invariance cannot increase the VC dimension of a model.
Prior to that, the specific case of neural networks was addressed 
by John Shawe-Taylor, who calculated sample complexity bounds for 
neural networks whose weights can be partitioned into equivalence classes,
which is directly applicable to invariant/equivariant
networks \parencite{shawe1991threshold,shawetaylor1995sample}.

A heuristic argument for reduced sample complexity for invariant models
is made by \textcite{mroueh2015learning}
in the case that the input space has finite cardinality.
The sample complexity of linear classifiers with invariant representations
trained on a simplified image task is discussed briefly 
by \textcite{anselmi2014unsupervised},
the authors conjecture that a general result may be obtained using wavelet transforms.
A similar argument for sample complexity reduction due to translation invariance
is made by \textcite{anselmi2016invariance,anselmi2016unsupervised} where it is
argued that the sample complexity depends on the input space, which is, in
effect, significantly smaller 
from the perspective of a translation invariant model.
This idea is prevalent in the literature and is captured 
rigorously in \cref{sec:sample-complexity}.

Indeed, there are many sample complexity results 
that are similar in spirit to those in \cref{sec:sample-complexity},
where we consider coverings on the hypothesis class and input space that reduce
for invariant/equivariant hypotheses.
This result, along with the rest of \cref{chap:general-theory-ii}, 
is relatively recent work \parencite{elesedy2022group}.
Earlier, \textcite{sokolic2017generalization}
built on the work of \textcite{xu2012robustness}
by considering classifiers that are invariant to a finite
set of transformations. Their results apply to the case where
the metric on the input space is the Euclidean norm
contain an implicit margin constraint on the training 
data.
\textcite{sannai2021improved} prove a generalisation bound for 
invariant/equivariant neural networks in the case of finite groups
in terms of covering numbers of the quotient space $\X / \G$.
\textcite{zhu2021understanding}
use a pseudo-metric that exploits the group action
to express the covering number of the hypothesis
class on the training set in terms of a covering number of the hypothesis
class on the orbit of the training set under the group action.
The work of \textcite{shao2022a} was published concurrently
with what is presented in \cref{chap:general-theory-ii}
and, at least in its learning-theoretic perspective,
is most closely related.
Their main results include upper and lower sample complexity
bounds for learning with invariant hypotheses on an invariant 
distribution in terms of an adaptation of the 
VC dimension that incorporates the group action \parencite[Theorem 4]{shao2022a}.

At a high level, all but three works on the generalisation of
invariant/equivariant models of which we are aware
construct a generic generalisation bound that can
be shown to reduce under the assumption that the class of predictors
satisfies a symmetry.
\textcite{bietti2021on} study the restricted setting
of a discrete grid for the input space and subsets of
the symmetric group for symmetries, and
use spherical harmonics to 
derive a sample complexity upper bound
for kernel ridge regression that reduces by a factor of
the size of the group for invariant kernels.
Notably, their setup is a special case of our condition \cref{eq:kernel-switch} 
with a finite group and an inner product kernel \parencite[Eq. 8]{bietti2021on}.
\textcite{lyle2019analysis} study the effect of symmetry on
the marginal likelihood and use this for model selection.
\textcite{behboodi2022a} consider equivariant networks in the Fourier
domain to derive norm based PAC Bayes generalisation bounds.
\textcite{wang2023general} study lower bounds for the generalisation error
of equivariant predictors when the symmetry is misspecified
to varying degrees.
\textcite{tahmasebi2023exact} derive upper and lower sample complexity bounds 
for invariant kernel regression on manifolds. 

However, all of the aforementioned works are worst-case, in the sense that they
give sample complexity bounds that
apply to an entire class of predictors simultaneously.
Of course, this does not guarantee that the generalisation of 
any given predictor (at a fixed number of examples) would be improved
if it were somehow replaced with an invariant/equivariant one.
This issue was not resolved until 
\parencite{elesedy2021provably},
one of the works contributing to this thesis,
wherein it was shown in an abstract setting that for
any predictor that is not invariant/equivariant there is an invariant/equivariant
predictor with strictly lower risk on all regression tasks with
an invariant/equivariant target.
Explicit calculations of this generalisation
improvement were given for linear 
models \parencite[Theorem 7 and Theorem 13]{elesedy2021provably}
and subsequently for kernel ridge regression in a separate work
which also forms part of this thesis \parencite{elesedy2021kernel}.
Concurrently, \textcite{mei2021learning} analyse the
asymptotic generalisation benefit of invariance in random feature models and kernel
methods, providing the only other strict generalisation benefit for invariance
of which we are aware. A comparison of their results to the relevant results in
this thesis is given in \cref{sec:other-work-comparison}.

\section{Invariant and Equivariant Models}\label{sec:lit-review-models}
While there has been a recent surge in interest,
symmetry and invariance have a long history in machine learning,
for instance appearing in the classical work of \textcite{fukushima1980neocognitron}.
See \parencite{wood1996invariant} and references therein.
The vast majority of modern implementations concern multi-layer perceptrons
(MLPs) and variations of convolutional neural networks (CNNs)
and our focus will reflect this.
However, implementation of invariance/equivariance in other models has been explored too,
as we outline below.

\subsection{The Symmetric Zoo}
Invariance and equivariance have been implemented in various
models in machine learning and related areas.
To name a few: steerable filters for image processing \parencite{freeman1991design},
invariance in signal processing \parencite{holmes1987mathematical} and
scattering transforms \parencite{mallat2012group,bruna2013invariant},
kernels \parencite{haasdonk05invariancein,haasdonk2007invariant,zhang2013learning,raj2017local},
support vector machines 
\parencite{scholkopf1996incorporating,chapelle2001incorporating,%
decoste2002training,walder2007learning}
and feature spaces of invariant polynomials
\parencite{schulz1994constructing,schulz1992existence,haddadin2021invariant}.

In addition, recent works include
equivariant Gaussian processes and neural 
processes \parencite{holderrieth2021equivariant},
capsule networks \parencite{lenssen2018group},
self attention and transformers \parencite{%
hutchinson2021lietransformer,he2021gauge,fuchs2020se,fuchs2021iterative},
autoencoders \parencite{xu2021group,winter2022unsupervised},
normalising flows
\parencite{rezende2019equivariant,kohler2020equivariant,satorras2021en,boyda2021sampling}
and graph networks \parencite{maron2018invariant,satorras2021n,puny2022frame},
whose approximation capacity has been studied by
\textcite{azizian2021expressive}.

Methods and results also exist that can be applied to a range of models.
Of course, averaging any predictor over the group with $\O$
or $\qq$ produces invariance or equivariance respectively but is often
computationally infeasible.
\textcite{puny2022frame}
find a computationally tractable alternative by averaging over a selected subgroup.
Taking a viewpoint similar to \cref{chap:general-theory-ii},
\textcite{aslan2023group} propose to learn invariant/equivariant models by
projecting the data onto a space of orbit representatives. 
\textcite{villar2021scalars} show how 
functions equivariant
to physical symmetries such as rigid body transformations and
the Lorentz group can be expressed solely in terms of invariant
scalar functions.
In a similar vein, \textcite{blumsmith2022equivariant} show how to parameterise polynomial or
smooth equivariant functions in terms of invariant ones.
Finally, \textcite{anselmi2016invariance} discuss the foundations of learning
representations that are both invariant and \emph{selective}, meaning that the
representations for two inputs are equal only if the inputs belong to the same orbit.

\subsection{Equivariant MLPs and CNNs}\label{sec:lit-review-nn}
The design of invariant neural networks was first considered by
Shawe-Taylor and Wood,
in which ideas from representation theory are applied to find weight tying schemes 
in MLPs that result in group invariant 
architectures \parencite{shawe1989building,shawe1993symmetries,shawe1994introducing,%
wood1996representation}.
Similarly, \textcite{ravanbakhsh2017equivariance} observe that
the equivariance in a linear map $\R^d \to \R^k$ 
implies symmetries in the corresponding weight matrix 
$M \in \R^{k\times d}$ and use this to derive weight
tying schemes for equivariant neural networks.
\textcite{finzi2021practical} reduce layer-wise equivariance of the network
to a finite dimensional linear system that the weights must satisfy and
use this to engineer equivariant MLPs,
their method works even for infinite Lie groups (provided they have
finite dimension).
\textcite{simard1991tangent} take a different route,
regularising the directional derivatives of the learned model to
encourage invariance to local transformations.

For CNNs, weight tying for equivariance to $90^\circ$ rotations
was proposed by \textcite{dieleman2016exploiting}.
A weight-tying approach for arbitrary symmetry groups
was developed for CNNs by \textcite{gens2014deep}
using kernel interpolation.
\textcite{marcos2017rotation} apply convolutional filters at a range of 
orientations to produce rotation equivariant networks.
To achieve rotation invariance,
\textcite{kim2020cycnn} map images to polar co-ordinates before
performing the convolution.
In addition, an equivariant version of the 
subsampling/upsampling operations in CNNs
was proposed by \textcite{xu2021group}.

It is well known that CNNs are equivariant to (local) translations
\parencite{lenc2015understanding} and this is widely believed to be a
key factor in their generalisation performance.
The G-CNN was introduced by \textcite{cohen2016group}
who, by viewing the standard convolutional layer as the convolution operator on
feature maps over the translation group,
generalised the convolutional layer to be equivariant to other groups
that preserve the two dimensional lattice $\mathbb{Z}^2$.\footnote{%
The standard convolutional layer is actually interpreted as the 
group \emph{correlation} by \textcite{cohen2016group},
but we follow the authors in blurring this distinction.}
Following this were many works borrowing ideas from harmonic analysis,
representation theory and mathematical physics
to generalise the basic approach of the G-CNN to various groups and input spaces
\parencite{cohen2017steerable,cohen2018spherical,esteves2018learning,kondor2018clebsch,%
cohen2019general,weiler2019general,finzi2020generalizing,weiler2021coordinate}.
\textcite{esteves2020theoretical} gives an exposition of some of these
models and the underlying mathematical concepts.

Following the introduction of the G-CNN,
\textcite{kondor2018generalization} show that any equivariant
neural network layer on a homogeneous space 
(one for which $\forall x, y \in \X$ $\exists g\in\G$ such that $x=gy$)
can be expressed in terms of a group convolution.
This was elegantly generalised to non-homogeneous spaces
by \textcite[Section 4.3]{portilheiro2022tradeoff}.
\textcite{lang2021a} give a general parameterisation for the filters
in these equivariant convolutions that holds for any compact group.
Getting even more general,
\textcite{bloem2020probabilistic} characterise the structure
of invariant/equivariant probability distributions using an extension
of noise outsourcing 
(also known as transfer) \parencite[Theorem 6.10]{kallenberg2006foundations} 
to connect functional and probabilistic symmetries.

There are results that establish,
across a variety of settings,
that any continuous invariant/equivariant function
can be approximated to arbitrary accuracy with
an invariant/equivariant neural network \parencite{%
    yarotsky2022universal,%
    ravanbakhsh2020universal,%
    maron2019universality,%
}.
In addition, \textcite{lawrence2022barrons} study efficient approximation of 
smooth functions using neural network architectures that are
invariant to subgroups of the orthogonal group.
While \textcite{qin2023intrinsic} compare the computational complexity of two layer
ReLU networks with similar equivariant networks.

There are also works specialised to the theory of learning
equivariant neural networks.
\textcite{bietti2019group} study the invariance, stability
and sample complexity of convolutional neural networks
with smooth homogeneous activation functions by embedding them in a certain 
reproducing kernel Hilbert space.
\textcite{petersen2022vc} estimate the VC dimension of G-CNNs
and prove the existence of
two layer G-CNNs that are invariant to an infinite group,
yet have infinite VC dimension.
\textcite{lawrence2022implicit} analyse the implicit
bias of linear G-CNNs when trained by gradient descent,
while \textcite{chen2023implicit} study the implicit bias of linear equivariant
steerable CNNs trained by gradient flow.

A sub-field has emerged for learning functions on sets.
We discuss permutation symmetry in more detail below
but briefly mention a few related works now.
\textcite{maron2020learning} study the learning of functions on sets
whose elements themselves satisfy symmetries.
\textcite{wang2020equivariant} construct architectures for
hierarchical symmetries, such as permutation equivariant maps of sets
(that themselves are invariant to permutation of their elements)
and apply it to semantic segmentation of point cloud data.
A transformer for use on set valued data was developed by
\textcite{lee2019set}.

\subsubsection{Permutation Equivariance}\label{sec:permutations}
Particular attention has been paid to 
neural networks $f: \R^n \to \R^m$ 
that are equivariant to the action of the permutation group $\sym_n$
on their inputs. The representation corresponding to $\pi \in \sym_n$ on the inputs
is the natural action $f(x_1, \dots, x_n) \mapsto
f(x_{\pi^{-1}(1)}, \dots, x_{\pi^{-1}(n)})$.
The most common representations on the outputs are
the trivial representation, the natural representation 
when $m=n$, or $\sign(\pi)$ where here $\sign$
denotes the $\sign$ of the permutation $\pi$, 
which is the parity of the product of the parities of the transpositions
in its decomposition thereof.
These three possibilities fall under the names
of \emph{permutation invariant}, \emph{permutation equivariant} and 
\emph{fermionic} networks respectively.
Fermionic networks are also known as anti-symmetric networks,
which is a special case of the usage of the term $\G$-anti-symmetric in this work.
In particular, it is straightforward to see that if $f$ is a fermionic network,
then one must have $\O f = 0$ if $n> 1$.\footnote{%
    We have $\abs{\sym_n} = n!$ which is even when $n > 1$.
    The transposition that switches the first and second 
    elements of a list is a bijection, so must map half of the 
    elements of $\sym_n$ to the other half. Split the sum
    in $\O f$ into these two halves and the 
    two terms cancel.}

\textcite{zaheer2017deep} propose the use of permutation invariant
neural networks for tasks with inputs that are sets
and
show that any continuous permutation
invariant function $f: [0,1]^n \to \R$ must have the form
\begin{equation}\label{eq:permutation-invariant-decomposition}
    f(x) = f_1\left(\sum_{i=1}^n f_2(x_i)\right)
\end{equation}
for some
continuous functions $f_1$ and $f_2$ with $f_2$
independent of $f$.
The importance of continuity and the dimension of any
intermediate latent spaces used in the computation of $f$
(of particular relevance to neural networks) is considered
by \textcite{wagstaff2019limitations,wagstaff2022universal}.
In a sense, these results say that the only
permutation invariant function is summation.

There has been considerable work on the
theory of permutation equivariant networks.
\textcite{abrahamsen2023anti,hutter2020representing}
consider approximating and representing fermionic 
functions in terms of sums of 
determinants respectively. 
\textcite{han2022universal} study
approximating permutation invariant and fermionic 
functions. While universal approximation of permutation equivariant 
functions was proved by
\textcite{sannai2019universal,segol2020universal}.
Finally, \textcite{nachum2020symmetry} study the ability of a generic
two layer neural network to learn a certain class of permutation
invariant functions.

One application of permutation invariant networks is
point cloud modelling.
Point cloud networks are permutation invariant neural networks
that are designed to preserve the shape of point
clouds \parencite{qi2017pointnet}.
Point cloud networks may also be 
invariant or equivariant to rigid body transformations
(translation and rotation) and
have applications in computer vision and 
robotics \parencite{guo2020deep}. Their approximation properties
have been studied by \textcite{dym2021on}.

\subsection{A Remark on the Representation of Invariant Functions}
The work from \textcite{bloem2020probabilistic}
which we referred to earlier
offers a probabilistic generalisation of 
\cref{eq:permutation-invariant-decomposition}
in an analogous expression for random variables.
Permutations are generalised to any compact group
and the sum is replaced with a \emph{maximal invariant},
which is a function $M$ on $\X$ that takes a distinct constant value 
on each orbit of $\X$ under $\G$.
That is, $M$ is invariant and $M(x) = M(y)$ implies $gx = y$ for some $g\in\G$.
As it happens, using the below theorem,
if there exists a function $\varphi$ such that $\O \varphi$
is a maximal invariant then a deterministic generalisation 
of 
\cref{eq:permutation-invariant-decomposition}
to any compact group is immediate.

\begin{theorem}[{\parencite[Theorem 6.2.1]{lehmann2005testing}}]\label{thm:maximal-invariant}
    Let $M$ be a maximal invariant on $\X$ with respect to $\G$.
    For all functions $f$ on $\X$, $f$ is $\G$-invariant if and only if it
    can be written $f(x) = h(M(x))$ for some function $h$.
\end{theorem}
\begin{proof}
    Clearly $h(M(x))$ is invariant and
    if $M(x) = M(y)$ then maximality implies $x = gy$ for some $g\in\G$
    so $f(x) = f(y)$. 
\end{proof}

\subsection{Applications}
There are many applications of symmetric models.
Most often these are in domains where the problem is known to satisfy a symmetry 
a priori, where experimenter choices or data representation are arbitrary
and not of intrinsic significance (e.g.,~choice of reference frame
for measurement
in classical dynamics or an ordered representation of a set), or simply where 
symmetry is believed to be a useful
inductive bias. Below we give a few examples.

Neural networks invariant to Galilean transformations 
have been used for predicting the Reynolds stress anisotropy tensor in 
fluid dynamics
\parencite{ling2016reynolds}.
Rotation invariant neural networks have been applied to
galaxy morphology prediction by \textcite{dieleman2015rotation}
and for learning energy potentials of
molecular systems by \textcite{anderson2019cormorant}.

In particle physics, 
\textcite{bogatskiy2020lorentz}
apply Lorentz group equivariant neural networks to
tagging top quark decays and proton-proton collisions.
For quantum mechanical systems, invariant neural networks have been used to 
model ground state wave functions and dynamics 
\parencite{luo2021gauge,luo2023gauge} and
fermionic networks have been applied to quantum chemistry
\parencite{pfau2020ab}.

In the life sciences, 
equivariant attention was applied to protein structure prediction by
\textcite{jumper2021highly}
and permutation equivariant networks applied to protein-drug binding
by \textcite{hartford2018deep}.
While, in medicine, rotation invariant networks have been applied to
image analysis \parencite{bekkers2018roto} and G-CNNs to pulmonary nodule
detection \parencite{winkels2018d}.

\textcite{rahme2021permutation,qin2022benefits}
propose and study the application of permutation equivariant networks to 
auction design.
Finally, \textcite{duan2023equivariant} study the benefit of permutation
equivariance when predicting game theoretic equilibria.

\subsection{Learning Symmetries from Data}
For all of the works we mentioned previously, the symmetry of the
task must be known in advance of designing the model.
In some circumstances this may not be possible, in which case
learning the symmetry from data is a natural approach.
In addition, by using this approach 
the model could uncover symmetries in the system unknown to the
user or even refute hypothesised symmetries which it would otherwise
be constrained to represent if hard-coded.
Eliminating a stage of the design process using data is an exciting prospect.
We mention some examples of work in this area below.

\textcite{anselmi2019symmetry}
outline principles for
learning symmetries of data and learning equivariant representations
without explicit knowledge of the symmetry group.
In particular, they propose to learn symmetries from data using a regularisation scheme.
In addition, \textcite{desai2022symmetry} use a generative adversarial network to discover
symmetries in data and
\textcite{christie2022testing} develop statistical tests for equivariance.

\textcite{benton2020learning} use data augmentation and learn a
parameterised group of transformations in conjunction with the model
parameters to learn partial invariances from data (e.g.,~invariance to 
a connected subset of $\SO_2$).
\textcite{van2018learning} learn parameterised invariances in 
Gaussian processes through their effect on the marginal likelihood.
In an adjacent area, \textcite{cubuk2018autoaugment} propose the automatic learning
of data augmentation policies.

\textcite{dehmamy2021automatic}
learn Lie group symmetries in a G-CNN by parametrising the filters in
terms of the basis of the Lie algebra.
Learning partial equivariances in G-CNNs is studied by \textcite{romero2022learning}.
In addition, \textcite{zhou2021metalearning} propose a meta-learning approach to learn
weight sharing patterns that produce equivariant neural networks.
Finally, addressing the theoretical limitations of this general approach,
\textcite{portilheiro2022tradeoff} shows, by considering aliasing
between symmetries, that under certain conditions
it is impossible to simultaneously learn group symmetries and functions 
equivariant with respect to them using an equivariant ansatz.

\section{Other Forms of Symmetry in Machine Learning}\label{sec:lit-review-other-symmetry}
General ideas of symmetry appear in many places in machine learning,
but not necessarily in the form of invariance/equivariance
of the predictor.
These areas are outside of the topic of this thesis, but we 
mention a few examples.

\paragraph{Geometric Deep Learning}
The study of invariant/equivariant models can be placed inside in
the broader sub-field of \emph{geometric deep learning}, that uses ideas from
geometry, algebra and graph theory to enforce inductive 
biases \parencite{bronstein2017geometric,bronstein2021geometric,gerken2021geometric}.

\paragraph{Conservation Laws}
Neural networks have been used to parameterise Hamiltonians and Lagrangians
for models of physical systems, automatically enforcing conservation laws
\parencite{greydanus2019hamiltonian,cranmer2019lagrangian}.
For each conserved quantity there is a symmetry and vice versa by
Noether's theorem.

\paragraph{Passive Symmetries}
\textcite{villar2023passive} highlight the potential importance of
recognising
symmetries that arise in tasks due to arbitrary experimenter choices,
for instance when collecting or processing data.
This is particularly important 
when the fundamental properties of a problem are independent of the 
co-ordinate description used in computation.
In a similar vein, \textcite{villar2022dimensionless} study the importance
of \emph{units equivariance}, meaning equivariance of
the model to the choice of units used to represent the data.

\paragraph{Algorithmic Symmetry}
\textcite{abbe2022nonuniversality} exploit \emph{algorithmic symmetries},
symmetries of the training algorithm when viewed as a mapping from 
the training sample to a predictor, to prove limitations on what can
be learned by neural networks trained by noisy gradient descent.
This is a generalisation of the orthogonal equivariance of
gradient descent which was exploited in the famous work 
of \textcite{ng2004feature} 
to argue for the LASSO over the ridge for regularisation in logistic regression.
The orthogonal equivariance property was also
applied in \textcite{li2020convolutional} to demonstrate a 
separation in
sample complexity 
between convolutional and fully connected networks.

\paragraph{Weight Space Symmetries in Neural Networks}
It is well known that the description in terms of the parameters
of the function computed by standard neural network architectures is
degenerate. That is, the map from ordered lists of weights to functions is not
injective.
As far as we are aware, this fact is yet to make a significant impact on the mainstream
theory of neural networks. In any case, list a few works in this area.
\textcite{chen1993geometry} analyse the group theoretical properties of a 
class of output-preserving weight transformations.
\textcite{ruger1996clustering} derive a metric on the weight space of MLPs that
removes the permutation symmetries.
\textcite{simsek2021geometry} study the effect of weight-space permutation symmetries
on the loss landscape.
\textcite{kunin2021neural} examine the effect of weight space symmetries on optimisation.

\paragraph{Data Augmentation}
Data augmentation is a procedure in which the covariates 
are replaced by their image under a random transformation.
If the transformations are sampled uniformly from a compact group
then in expectation (over the transformations only)
data augmentation replaces the training loss function $\ell$ with $\O \ell$,
where for the purposes of the averaging it is viewed as a function of the covariates.
Data augmentation has been shown to help learn transformation invariances
from data \parencite{fawzi2015manitest}.
Moreover, \textcite{kashyap2021robustness} show that robustness to 
input transformations, a form of approximate invariance (although not
always with respect to a group), is a strong predictor
of generalisation performance in deep learning.
From a theoretical perspective,
\textcite{lyle2020benefits} derive a PAC Bayes bound 
and use it to study the relative benefits of feature
averaging and data augmentation.
\textcite{chen2020group} study the statistical properties of estimators
using data augmentation when the transformations form a group.
\textcite{shao2022a} use the framework of PAC learning
and some adaptations of the VC dimension to quantify the effect
of data augmentation on the sample complexity of learning with
empirical risk minimisation.

\chapter{General Theory I: Averaging Operators}\label{chap:general-theory-i}

\section*{Summary}
We explore symmetry from a functional perspective,
finding in \cref{lemma:l2-decomposition} that any function $f$ can be written
\[
    f = \bar{f} + f^\sperp
\]
where $\bar{f}$ is equivariant, $f^\sperp$ represents the $\G$-anti-symmetric 
(non-equivariant) part of $f$ and, most importantly, the two
terms are orthogonal as functions. As a warm up, we use this result to
give some new perspectives on feature averaging. We then apply it to generalisation 
in \cref{lemma:generalisation}, deriving a strict generalisation benefit
for equivariant predictors. 

\section{Averaging and the Structure of $\Lmuy$}
The following result shows that
any function in $\Lmuy$ can be decomposed
orthogonally into two terms that we call its
$\G$-symmetric and $\G$-anti-symmetric parts.
Recall that since $\O$ is just a special case of $\qq$, \Cref{lemma:l2-decomposition} 
applies to both operators.

\newcommand{\LemmaDecomposition}{%
    Let $U$ be any subspace of $\Lmuy$ that is closed under $\qq$,
    meaning $\qq U \subset U$.
    Define the subspaces $S$ and $A$ of $U$ consisting of
    the $\G$-symmetric and $\G$-anti-symmetric functions respectively,
    $S = \{f \in U: \text{$f$ is $\G$-equivariant}\}$
    and 
    $A = \{f \in U: \qq f = 0 \}$.
    Then $U$ admits the orthogonal decomposition $U = S \oplus A$.
}
\begin{lemma}\label{lemma:l2-decomposition}
    \LemmaDecomposition
\end{lemma}
Although discovered independently, \cref{lemma:l2-decomposition} is not a fundamentally
new idea.
The same theme appears in \textcite{reisert2007learning} and likely many other works.
A proof is given in \cref{sec:proof-l2-decomposition} which 
establishes that $\qq$ is a self-adjoint orthogonal projection on
$\Lmuy$, from which the conclusion follows.

\Cref{lemma:l2-decomposition} says that any function $u \in U$ can be written $u = s + a$,
where $s$ is equivariant, $\qq a  = 0$ and $\inner{s}{a}_\mu = 0$.
We refer to $s$ and $a$ as the $\G$-symmetric and $\G$-anti-symmetric parts of $u$
respectively.
In general this does not imply that $a$ is an odd function, that it 
outputs an anti-symmetric matrix or that its values are negated by
swapping arguments. 
These are, however, special cases.
If $\G = \cyc_2$ acts by $x\mapsto -x$ then odd functions $f: \R
\to \R$ will be $\cyc_2$-anti-symmetric.
If $\G=\cyc_2$ acts on matrices by $M \mapsto
M^\top$ then $f: M \mapsto \frac12 (M - M^\top)$ is also $\cyc_2$-anti-symmetric, but with
respect to a different action. Finally, if
$\G = \sym_n$ and $f: \R^n \to \R$ with $f(x_1, \dots, x_i,\dots, x_{j}, \dots, x_n) =
-f(x_1, \dots, x_{j},\dots, x_{i}, \dots, x_n)$  $\forall i,j \in \{1, \dots, n\}$
then $f$ is $\sym_n$-anti-symmetric.

\begin{figure}[h]
    \centering
    \includegraphics[width=\figwidth]{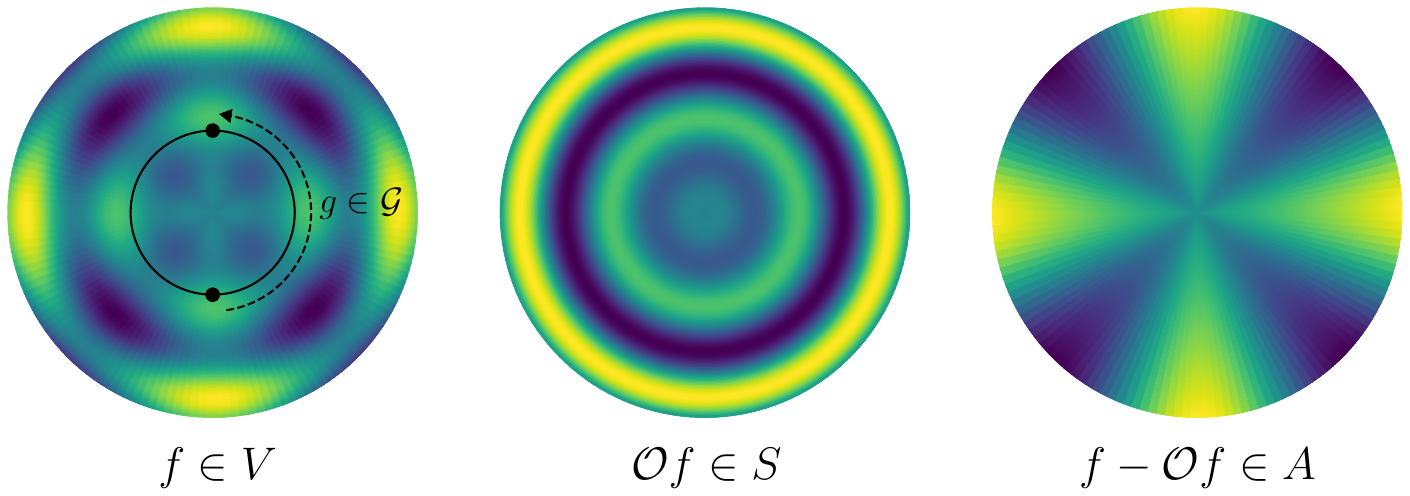}
    \vspace{-10pt}
    \caption{Example of a function decomposition.
    The figure shows $f(r, \theta) = r \cos{(r - 2\theta)} \cos{(r + 2\theta)}$
    decomposed into its $\G$-symmetric and $\G$-anti-symmetric parts in $V = S \oplus A$
    under the natural action of $\G = \SO_2$ on $\R^2$.
    See \Cref{example:rotations}. 
    Image credit: Sheheryar Zaidi. Best viewed in colour.}\label{fig:orbit}
\end{figure}

\begin{example}\label{example:rotations}
    Let $\X=\R^2$, $\Y=\R$, $\mu = \normal(0, I_2)$ and set $V= \Lmu$.
    Let $\G = \SO_2$ act by rotation about the origin, with respect to which
    the normal distribution is invariant.
    Using \Cref{lemma:l2-decomposition} we may write $V = S \oplus A$.
    Alternatively, consider polar coordinates $(r, \theta)$,
    then for all $f \in V$
    we have $\O f (r, \theta) = \frac{1}{2\pi} \int_0^{2\pi} f(r, \theta')\dd \theta' $.%
    \footnote{%
    This is to be interpreted informally as, strictly, evaluation is not defined.
    The existence of $\O f$ is guaranteed by \cref{prop:q-well-defined} 
    so the integral is finite for almost all $r$ and the rest are harmless.%
    }
    So, naturally, any invariant $f$ depends only on the radial coordinate.
    Similarly, any $h$ for which $\O h = 0$ must have 
    $\O h (r, \theta)  
    = \frac{1}{2\pi} \int_0^{2\pi}h(r, \theta') \dd \theta' 
    = 0$ for all $r$, and $A$ consists entirely of such functions.
    For example, $r^{3} \cos \theta \in A$.
    We then recover 
    $\inner{s}{h}_\mu
      = \frac{1}{2\pi}\int_\X s(r) h(r, \theta)\ee^{-r^2/2}r \dd r \dd \theta  
      = 0$ for all $s \in S$
    by integrating $h$ over $\theta$.
    Intuitively, one can think of the functions in $S$ as only varying perpendicular to the flow
    of $\G$ on $\X = \R^2$ and so are preserved by it, while the functions 
    in $A$ average to $0$ along this flow,
    see \cref{fig:orbit}.
\end{example}

\begin{remark}
\textcite{villar2022dimensionless} discuss
generalising \cref{lemma:l2-decomposition}
to the non-compact group of scalings $x\mapsto c x$ 
and demonstrate the impossibility of doing so for
real (as opposed to complex) $c$.
\end{remark}

\subsection{Proof of \cref{lemma:l2-decomposition}}\label{sec:proof-l2-decomposition}
In this section we derive the following result.
\begin{lemma*}[\cref{lemma:l2-decomposition}]
    \LemmaDecomposition
\end{lemma*}

We first check that $\qq$ is well-defined.
\begin{proposition}\label{prop:q-well-defined}
    Let $f \in \Lmuy$, then
    \begin{enumerate}
        \item $\qq f$ is $\mu$-measurable,
            and
        \item $\qq f \in \Lmuy$ with $\norm{\qq f}_\mu \le \norm{f}_\mu$.
    \end{enumerate}
\end{proposition}
\begin{proof}
        The mapping
        $m: (g, x) \mapsto g^{-1}f(gx) \in \Y$
        is $(\lambda \otimes \mu )$-measurable by considering the following
        composition and applying \cref{lemma:measurable-collection}
        \[
            (g, x)
            \mapsto (g, gx)
            \mapsto (g, f(gx))
            \mapsto g^{-1}f(gx).
        \]
        The lemma requires that each co-ordinate of each tuple is
        generated by a measurable mapping of the previous tuple, which is true
        either by assumption or trivially for $(g, x) \mapsto g$.

        \cref{lemma:measurable-collection} also allows us to
        verify the $\mu$-measurability of $\qq f$ component-wise.
        By \cref{lemma:integrate-section}, it's sufficient to 
        verify that each component of $m$ is $(\lambda \otimes \mu)$-integrable. 
        Consider, using Fubini's theorem, the unitarity of $\psi$ and invariance of $\mu$
        \begin{align*}
            \int_\G \int_\X \norm{m(g, x)}^2 \dd\mu(x)\dd \lambda (g) 
            &= \int_\G \int_\X \norm{g^{-1}f(gx)}^2 \dd\mu(x)\dd \lambda (g) \\
            &= \int_\G \int_\X \norm{f(gx)}^2 \dd\mu(x)\dd \lambda (g) \\
            &= \int_\X \norm{f(x)}^2 \dd\mu(x) \\
            &= \norm{f}_\mu^2 < \infty
        \end{align*}
        so that each component of $m$ is in $L_2(\lambda \otimes \mu)$.
        \cref{thm:bg-LpLqinclusion} gives
        $L_2(\lambda \otimes \mu) \subset L_1(\lambda \otimes \mu)$
        because $\lambda \otimes \mu$ is bounded.

        So far we have established the first assertion in
        the statement,
        we now verify that $\qq f \in \Lmuy$. 
        In \cref{eq:q-well-defined-a} we use Fubini's theorem which
        we will justify at the end of the proof,
        in \cref{eq:q-well-defined-b} we use unitarity and
        in \cref{eq:q-well-defined-c} we use the invariance of $\mu$ and $\lambda$
        \begin{align*}
            \norm{\qq f}^2_\mu
            &= \int_\X \left\langle 
            \int_\G g_1^{-1} f(g_1x) \dd\lambda(g_1),
            \int_\G g_2^{-1} f(g_2x) \dd\lambda(g_2)
            \right\rangle \dd\mu(x) 
            \\ 
            &=  \int_\X \int_\G \int_\G \left\langle 
                g_1^{-1} f(g_1x),
                g_2^{-1} f(g_2x) 
            \right\rangle \dd\lambda(g_1)\dd\lambda(g_2) \dd\mu(x) 
            \\ 
            &=  \int_\G \int_\G \int_\X
            \left\langle 
                g_1^{-1} f(g_1x),
                g_2^{-1} f(g_2x) 
            \right\rangle 
            \dd\mu(x) 
            \dd\lambda(g_1)\dd\lambda(g_2) 
            \tag{a}\label{eq:q-well-defined-a}\\
            &=  \int_\G \int_\G \int_\X
            \left\langle 
                 f(g_1x),
                (g_2g_1^{-1})^{-1} f(g_2x) 
            \right\rangle 
            \dd\mu(x) 
            \dd\lambda(g_1)\dd\lambda(g_2) 
            \tag{b}\label{eq:q-well-defined-b}\\
            &=  \int_\G \int_\X
            \left\langle 
                 f(x),
                g^{-1} f(gx) 
            \right\rangle 
            \dd\mu(x) 
            \dd\lambda(g)
            \tag{c}\label{eq:q-well-defined-c}\\
            &= \int_\G \inner{f}{g^{-1}\circ f \circ g}_\mu \dd\lambda(g) 
              \\
            &\le \norm{f}_\mu \int_\G\norm{g^{-1}\circ f \circ g}_\mu \dd\lambda(g).
        \end{align*}
        It remains to show that $\norm{g^{-1}\circ f \circ g}_\mu = \norm{f}_\mu$.
        For all $g\in\G$,
        \begin{align*}
            \norm{g^{-1}\circ f \circ g}_\mu^2
            &= \int_\X \inner{g^{-1}f(gx)}{g^{-1}f(gx)}\dd\mu(x)\\
            &= \int_\X \inner{f(gx)}{f(gx)}\dd\mu(x)\\
            &= \int_\X \inner{f(x)}{f(x)}\dd\mu(x)\\
            &= \norm{f}_\mu^2.
        \end{align*}
        We now justify \cref{eq:q-well-defined-a}. To apply Fubini's theorem
        we need each $\G$-section of the integrand to be 
        $(\lambda \otimes \mu)$-integrable, for which \cref{prop:inner-product-integrable}
        is sufficient.
\end{proof}

\begin{proposition}\label{prop:inner-product-integrable}
    For all $h, f \in \Lmuy$ the function $\inner{h(x)}{g^{-1}f(gx)}$ 
    is $(\lambda\otimes\mu)$-integrable.
\end{proposition}
\begin{proof}
    The function is measurable by expanding the inner product in a basis.
    Then by Cauchy-Schwarz and $a^2 + b^2 \ge ab$
    \begin{align*}
        &\int_\G \int_\X \abs{\inner{h(x)}{g^{-1}f(gx)}} \dd\mu(x)\dd\lambda(g)\\
        &\le
        \int_\G \int_\X \norm{h(x)}\norm{g^{-1}f(gx)} \dd\mu(x)\dd\lambda(g)\\
        &\le
        \int_\G \int_\X \norm{h(x)}^2 \dd\mu(x)\dd\lambda(g)
        +
        \int_\G \int_\X \norm{g^{-1}f(gx)}^2  \dd\mu(x)\dd\lambda(g)\\
        &= \norm{h}_\mu^2 + \norm{f}_\mu^2 \\
        &< \infty
    \end{align*}
    where in the last line we use unitarity and the invariance of $\mu$.
\end{proof}

Now for the important observation that $\qq$ identifies equivariance, as well as 
enforcing it. If there are any equivariant functions in $\Lmuy$ then in combination 
with \cref{prop:q-well-defined} this shows that $\qq$ has unit operator norm.
\begin{proposition}\label{prop:sym-cpt}
    $f$ is equivariant if and only if $\qq f = f$.
\end{proposition}

\begin{proof}
    Recall that $f$ is equivariant if $f(gx) = gf(x)$ for all $g\in \G$ and all $x \in \X$.
    Suppose $f$ is equivariant then for all $x \in \X$
    \[
        \qq f(x) 
        = \int_\G g^{-1}f(g x)\dd\lambda(g)
        = \int_\G g^{-1}gf(x)\dd\lambda(g)
        = f(x).
    \]
    Now assume that $\qq f = f$, so 
    $
    f(x) = \int_\G g^{-1}f(gx)\dd\lambda(g)
        $
    for all $x \in \X$.
    Take any $\tilde{g} \in \G$, then 
    \begin{align*}
        f(\tilde{g}x) 
        &= \int_\G g^{-1}f(g\tilde{g}x)\dd\lambda(g) 
        = \tilde{g}\int_\G (g\tilde{g})^{-1}f(g\tilde{g}x)\dd\lambda(g) 
        = \tilde{g}\int_\G g^{-1}f(gx)\dd\lambda(g)\\
        &= \tilde{g}f(x)
    \end{align*}
   where in the third equality we used the invariance of the Haar measure. 
\end{proof}

By \Cref{prop:sym-cpt}, $\qq^2 f = \qq f$ so $\qq$ is a projection.
The operator $\qq$ is also linear.
Let $U$ be a subspace of $\Lmuy$ such that $\qq U \subset U$.
Set $S = \qq U$ and $A = (\id - \qq)U$.
$S\subset U$, $A\subset U$ and $S \cap A$ is trivial, so
$S \oplus A \subset U$.
Any $f \in U$ can be written uniquely as $f = \bar{f} + f^\sperp$ where
$\bar{f} = \qq f$ and $f^\sperp = f - \qq f$,
so $U \subset S \oplus A$.
Hence $U = S \oplus A$.
\Cref{prop:sym-cpt} gives
$S = \{f \in U: \text{$f$ is $\G$-equivariant}\}$ and $A =  \{f\in U: \qq f = 0\}$
is easily established: $\qq A = \{0\}$ by linearity and idempotence, while
any $f$ with $\qq f=0$ has $f = (\id - \qq)f$.
Next we show that $\qq$ is self-adjoint with respect to $\inner{\cdot}{\cdot}_\mu$ which shows
that $\qq$ is an orthogonal projection.
The orthogonality in \cref{lemma:l2-decomposition} follows immediately,
since if $f \in S$ and $h \in A$ then
\[
    \inner{f}{h}_\mu = \inner{\qq f}{h}_\mu = \inner{f}{\qq h}_\mu = \inner{f}{0}_\mu = 0.
\]

\begin{proposition}\label{prop:self-adjoint}
    $\qq$ is self-adjoint with respect to $\inner{\cdot}{\cdot}_\mu$.
\end{proposition}

\begin{proof}
    At various points we
    apply Fubini's theorem, the unitarity of $\psi$, the invariance of $\mu$
    and use the change of variables $g \mapsto g^{-1}$.
    The application of Fubini's theorem is valid by \cref{prop:inner-product-integrable}.
    The inner product on $\Y$ commutes with integration (e.g., by expanding in a basis).
    Let $f, h \in \Lmuy$, then
    \begin{align*}
        \inner{\qq f}{h}_\mu
        &= \int_{\X} \left\langle \int_\G g^{-1}f(gx)\dd\lambda(g), h(x) \right\rangle 
        \dd \mu(x) \\
        &= \int_{\X}\int_\G  \inner{g^{-1}f(gx)}{h(x)} \dd\lambda(g)\dd\mu(x) \\
        &= \int_\G \int_{\X} \inner{g^{-1}f(gx)}{h(x)} \dd\mu(x) \dd\lambda(g)\\
        &= \int_\G \int_{\X} \inner{f(x)}{gh(g^{-1}x)} \dd\mu(x) \dd\lambda(g)\\
        &= \int_{\X} \left\langle f(x), \int_\G gh(g^{-1}x) \dd\lambda(g)\right\rangle \dd\mu(x) \\
        &= \int_{\X} \left\langle f(x), \int_\G g^{-1}h(gx) \dd\lambda(g)\right\rangle \dd\mu(x) \\
        &= \inner{f}{\qq h}_\mu.
    \end{align*}
\end{proof}

\subsection{The invariance of $\mu$}
The invariance of $\mu$ is used throughout the proof of \cref{lemma:l2-decomposition}.
There are many tasks for which invariance of the input distribution is a 
natural assumption, for instance in medical imaging \parencite{winkels2018d}, but
the reader may wonder whether it is necessary for our results.
In the following example from Sheheryar Zaidi,
invariance is equivalent to the orthogonality in the decomposition 
$\Lmuy = S \oplus A$. 
This means invariance is necessary for \cref{lemma:l2-decomposition} to hold in general.
An alternative phrasing is that the projection $\qq$ is only orthogonal
in the below setting if $\mu$ is invariant.
Note that in the example below $\psi$ is the trivial representation. 

\begin{example}[Sheheryar Zaidi \parencite{elesedy2021provably}]
    Let $\cyc_2$ act on $\X = \{-1, 1\} \times \{1\}$ by
    multiplication on the first coordinate. Let $V$ be the vector space of
    functions $f_{(t_1, t_2)} : \X \to \R$ with 
    $f_{(t_1, t_2)}(x_1, x_2)= t_1x_1+ t_2x_2$ 
    for all $t_1, t_2 \in \R$. We note that any
    distribution $\mu$ on $\X$ can be described by its probability mass
    function $p$, which is defined by $p(-1, 1)$ and $p(1, 1)$. Moreover, $\mu$
    is $\cyc_2$-invariant if and only if $p(-1, 1) = p(1, 1)$. Next, observe that
    the $\G$-symmetric functions $S$ and $\G$-anti-symmetric functions $A$ are precisely
    those for which $t_1 = 0$ and $t_2 = 0$ respectively. The inner product
    induced by $\mu$ is given by $\inner{f_{(a_1, a_2)}}{f_{(b_1, b_2)}}_\mu =
    (a_1 + a_2)(b_1 + b_2)p(1, 1) + (a_2 - a_1)(b_2 - b_1)p(-1, 1)$. With this,
    we see that the inner product $\inner{f_{(0, t_2)}}{f_{(t_1, 0)}}_\mu = t_1
    t_2 p(1, 1) - t_1 t_2 p(-1, 1)$ is zero for all $f_{(0, t_2)} \in S$ and
    $f_{(t_1, 0)} \in A$ if and only if $p(-1, 1) = p(1, 1)$. That is, if and
    only if $\mu$ is invariant.  
\end{example}

\section{Warm Up}\label{sec:warm-up}
In this section we consider some basic consequences 
of \cref{lemma:l2-decomposition} and of our setup more generally.
Recall the special case of $\qq$ where $\psi$ is the trivial representation,
corresponding to invariance rather than equivariance,
\[
    \O f (x) = \int_\G f (gx) \dd\lambda(g).
\]

\subsection{Feature Averaging as a Least Squares Problem}
When $f$ is a feature extractor, e.g., the final layer representations of a neural network,
$\O$ can be thought of as performing feature averaging.
Using \Cref{lemma:l2-decomposition}, feature averaging can be viewed as 
solving a least squares problem in $\Lmuy$.
That is, feature averaging sends $f$ to $\bar{f}$ where $\bar{f}$ is the 
closest invariant feature extractor to $f$.
This just a well known fact about orthogonal projections written in different terms,
but we give a proof for completeness. The same result holds for $\qq$.

\begin{proposition}\label{prop:fa-least-squares}
    Define $S$ and $A$ as in \Cref{lemma:l2-decomposition}.
    For all $f \in \Lmuy$, feature averaging with $\O$ maps $f \mapsto \bar{f}$
    where $\bar{f}$ is the unique solution to the least squares
    problem
    $
        \bar{f} = \argmin_{s \in S} \norm{f - s}^2_\mu
    $.
\end{proposition}
\begin{proof}
    By \cref{lemma:l2-decomposition} we can write $f = \bar{f} + f^\sperp$
    where $\bar{f} = \O f \in S$, $f^\sperp \in A$ and the two terms are orthogonal.
    Take any $h \in S$, then using the orthogonality
        \[
        \norm{f - h}_\mu^2 
        = \norm{(\bar{f} - h) + f^\sperp}_\mu^2
        = \norm{\bar{f} - h}_\mu^2 + \norm{f^\sperp}_\mu^2
        \ge \norm{f^\sperp}_\mu^2
        = \norm{f - \bar{f}}_\mu^2.
        \]
    Set $s = \bar{f}$ and suppose $\exists s' \in S$ with 
    $s' \ne s$ and $\norm{f -s '}_\mu = \norm{f - s}_\mu$, then 
    since $S$ is a vector space we have $s_{\frac12} = \frac12(s + s') \in S$.
    It follows, using Cauchy-Schwarz, that
    \begin{align*}
        \norm{f - s_{\frac12}}_\mu^2 
        &= \norm{ (f - s)/2  + (f - s')/2}_\mu^2\\
        &= \frac14 \norm{f - s}_\mu^2 + \frac14 \norm{f - s'}_\mu^2 + 
        \frac14 \inner{f - s}{f - s'}_\mu \\
        &\le \frac14 \norm{f - s}_\mu^2 + \frac14 \norm{f - s}_\mu^2 + \frac14 \norm{f -s}_\mu^2 \\
        & = \frac34 \norm{f - s}_\mu^2,
    \end{align*}
    which a contradiction unless $\norm{f - s}_\mu^2 = 0$, in which case
    $f = s$ in $\Lmuy$.
\end{proof}

\begin{example}
    Consider again the setting of \cref{example:rotations}.
    For simplicity, let $f(r, \theta)= f_{\text{rad}}(r)f_{\text{ang}}(\theta)$
    be separable in polar coordinates.  Notice that $\O f = c_f f_{\text{rad}}$
    where $c_f  = \frac{1}{2\pi} \int_0^{2\pi} f_\text{ang}( \theta)\dd \theta$.
    Then for any $s \in S$ we can calculate:
    \begin{align*}
        \norm{f - s}_\mu^2 
        &= \frac{1}{2\pi}\int_\X (f(r, \theta) - s(r))^2 \ee^{-r^2/2}r \dd r \dd \theta  \\
        &= \frac{1}{2\pi}\int_\X (f(r, \theta)- c_ff_{\text{rad}}(r))^2
        \ee^{-r^2/2}r \dd r \dd \theta \\ 
        &\phantom{=}+ \frac{1}{2\pi}\int_\X(c_ff_{\text{rad}}(r) -s(r))^2 
        \ee^{-r^2/2}r \dd r \dd \theta 
    \end{align*}
    which is minimised by $s = c_f f_\text{rad}$.
\end{example}

\subsection{Averaging and the Rademacher Complexity}
Let $T = (x_1, \dots, x_n)$ be a collection of points from $\X$.
The \emph{empirical Rademacher complexity} of a set $\F$ of functions $f: \X \to \R$
evaluated on $T$ is defined by
\[
    \Rad_T(\F)
    = \E\left[ \sup_{f \in \F}\labs{\frac1n  \sum_{i=1}^n \varsigma_i f(x_i)}\right]
\]
where the expectation is over the random variables 
$\varsigma_i \sim \unif \{-1, 1\}$ which follow the 
Rademacher distribution.
If the $T$ is random then
the empirical Rademacher complexity 
$\Rad_T(\F)$ is a random quantity and
the \emph{Rademacher complexity} $\ERad_n(\F)$ is defined by
$
    \ERad_n(\F) = \E[\Rad_T(\F)]
$.
The Rademacher complexity appears in the study of generalisation in statistical learning,
for instance see \parencite[Theorem 4.10 and Proposition 4.12]{wainwright2019high}
for upper and lower bounds respectively. A reduction in the Rademacher complexity
means that fewer examples are required to achieve a specified worst-case risk.

Let $\F \subset \Lmu$ and consider its $\G$-symmetric and
$\G$-anti-symmetric projections
$ \overline{\F} = \O \F $
and $ \F_\sperp =  (\id - \O)\F $
respectively.
We assume that both versions of the
Rademacher complexity are well-defined on $\F$, $\overline{\F}$ 
and $\F_\sperp$.

\begin{proposition}\label{prop:fa-rademacher}
    The Rademacher complexity of the feature averaged class satisfies
   \[
        0 \le  \ERad_n (\F)  -  \ERad_n (\overline{\F})  \le  \ERad_n (\F_\sperp)
   \]
   whenever the terms are finite and
   where the data are distributed $T\sim\otimes^n\mu$.
\end{proposition}

\begin{proof} 
    We will show that
    $
        \ERad_n (\overline{\F}) \le 
        \ERad_n (\F)  
        \le \ERad_n (\overline{\F})  + \ERad_n (\F_\sperp )
        $,
    from which the proposition follows immediately.
    We start by establishing
    $
        \ERad_n (\overline{\F}) \le \ERad_n ( \F )
        $.
    The action of $\G$ on $\X$ induces
    an action on $\X^n$ by $g(x_1, \dots, x_n) = (g x_1, \dots, g x_n)$ 
    under which $\otimes^n \mu$ is invariant.
    Let $X_1, \dots, X_n \sim \mu$, then note that
    \begin{align*}
        \ERad_n (\overline{\F} )
        &= \frac1n \E \left[\sup_{\bar{f}\in \overline{\F}} 
        \labs{\sum_{i=1}^n \varsigma_i \bar{f}(X_i)} \right]
        \\
        &= \frac1n \E 
        \left[
            \sup_{f\in \F} \labs{\sum_{i=1}^n \varsigma_i 
        \int_\G f(g X_i) \dd\lambda(g)}
        \right] \\
        &\le \frac1n \E\left[
        \sup_{f\in \F}\int_\G 
        \labs{\sum_{i=1}^n \varsigma_i f(g X_i)}\dd\lambda(g)
        \right].
    \end{align*}
    We obtain
    $
        \ERad_n (\overline{\F}) \le \ERad_n (\F)
    $
    by the fact that the last line above
    is dominated by \cref{eq:rad-b} below
    \begin{align*}
        \ERad_n (\F)
        &= \frac1n 
        \E  
        \left[
        \sup_{f\in \F}\labs{\sum_{i=1}^n \varsigma_i f(X_i)}
        \right]\\
        &= \frac1n 
        \int_\G 
        \E  
        \left[
        \sup_{f\in \F}\labs{\sum_{i=1}^n \varsigma_i f(g X_i)}
        \right]
        \dd\lambda(g)
        \tag{a}\label{eq:rad-a} \\
        &= \frac1n \E \left[
        \int_\G 
        \sup_{f\in \F}
        \labs{\sum_{i=1}^n \varsigma_i f(g X_i)}\dd\lambda(g)
        \right].
        \tag{b}\label{eq:rad-b} 
    \end{align*}
    In \cref{eq:rad-a} we used the invariance of $\mu$. In \cref{eq:rad-b}
    we use Fubini's theorem which also guarantees that the expression is well-defined
    (although possibly infinite).

    We now prove that
    $
        \ERad_n (\F)  
        \le \ERad_n (\overline{\F})  + \ERad_n (\F_\sperp)
        $.
    For any $f \in \F$ we can write $f = \bar{f} + f^\sperp$ where $\bar{f} \in \overline{\F}$ 
    and $f^\sperp \in \F_\sperp$ by \cref{lemma:l2-decomposition}.
    The result follows from
    taking expectations over $T \sim \otimes^n \mu$ of the below.
    For any $T = (x_1, \dots, x_n)$ 
    \begin{align*}
        \Rad_T (\F)  
        &= \frac1n \E\left[ \sup_{f\in \F} \labs{\sum_{i=1}^n \varsigma_i f(x_i)} \right]\\
        &= \frac1n \E\left[
            \sup_{f\in \F} \labs{\sum_{i=1}^n (\varsigma_i \bar{f}(x_i) 
            + \varsigma_if^\sperp(x_i)) }
            \right]\\
        &\le \frac1n 
        \E \left[
            \sup_{\bar{f}\in \overline{\F}} 
            \labs{\sum_{i=1}^n \varsigma_i \bar{f}(x_i)}
            \right]
        + \frac1n \E 
        \left[
            \sup_{f^\sperp \in \F_\sperp}\labs{ \sum_{i=1}^n \varsigma_i  f^\sperp(x_i) }
            \right]
        \\
        &= \Rad_T (\overline{\F})  + \Rad_T (\F_\sperp )
    \end{align*}
\end{proof}

\Cref{prop:fa-rademacher} says that the Rademacher complexity is reduced by
orbit averaging, but not by more than the complexity of the $\G$-anti-symmetric
component of the class.
This quantifies the improvement in worst-case generalisation 
from enforcing invariance by averaging in
terms of the extent to which the inductive bias is already present in the function class.
We provide stronger results in later sections and chapters, studying 
individual predictors and the average case for learning algorithms.

\section{Implications for Generalisation}
\subsection{The Generalisation Gap in Regression Problems}
In this section we apply \Cref{lemma:l2-decomposition} to derive a strict (i.e., non-zero) 
generalisation gap between predictors that have and have not been specified to have the 
symmetry that is present in the task.

Given some pair of random elements $(X, Y)$ with values on $\X \times \Y$
defining a supervised learning task (with inputs in $\X$ and outputs in $\Y$),
we define the \emph{risk} of a predictor $f$ as
\[
    R[f] = \E[\norm{f(X) - Y}^2].
\]
This is the same definition of the risk, just specialised to the loss function
$\ell(y, y') = \norm{y -y'}^2$, where $\norm{\cdot}$ is the inner product 
norm on $\Y$.
A consequence of the following result is that, on a task with equivariant structure,
the difference in risk between a predictor $f$ and any equivariant predictor $f'$ such that 
$R[f'] \le R[\qq f]$ is at least the norm of the $\G$-anti-symmetric component of of $f$.
This shows a barrier to generalisation if $f$ is not equivariant. It also shows
that the lower bound on the risk can be (approximately) eliminated by making $f$ 
(approximately) equivariant.
In short: for any non-equivariant predictor, there is an equivariant predictor that performs
better.
The projection $\qq f$ is the archetypal equivariant predictor to which $f$ should be compared.
From \cref{lemma:l2-decomposition}, $\qq f$ can be thought of the equivariant part of $f$
and in addition $\qq f$ is the closest equivariant predictor to $f$.

\begin{lemma}\label{lemma:generalisation}
    Let $X \sim \mu$ and let $Y$ have finite second moment.
    Assume either of the following models for $Y$
    \begin{enumerate}
        \item $Y \eqdist \fopt(X) + \xi$ where $\fopt \in \Lmuy$ is equivariant,
            $\E[\xi] = 0$ and $\E[\norm{\xi}^2] < \infty$, or
        \item $Y \eqdist \tilde{f}(X, \eta)$ where 
            $\tilde{f}: \X \times [0,1] \to \R$ 
            is equivariant in its first argument,
            $\eta \sim \unif[0, 1]$ and $\eta \indep X$.
    \end{enumerate}
    Let $f \in \Lmuy$ and let $f' \in \Lmuy$ be any predictor such that $R[f'] \le R[\bar{f}]$
    where $\bar{f}=\qq f$.
    Let $f^\sperp = f - \bar{f}$ be the $\G$-anti-symmetric component of $f$, then
    \[
        R[f] - R[f'] \ge \norm{f^\sperp}_\mu^2
    \]
    with equality when $R[f'] = R[\bar{f}]$.
\end{lemma}
\begin{proof}
    Assume the statement in the case of equality.
    Consider $f' \in \Lmuy$ such that $R[f'] \le R[\bar{f}]$, then
    $
        R[f] - R[f'] = R[f] - R[\bar{f}] + R[\bar{f}] - R[f'] \ge \norm{f^\sperp}_\mu^2
        $. We now address the case of equality for each model.
    Suppose $Y \eqdist \fopt(X) + \xi$. Using \cref{lemma:l2-decomposition} we write
    $f = \bar{f} + f^\sperp$ where the terms are orthogonal and the first is equivariant.
    Then using $\E[\xi]=0$ and the orthogonality 
    \begin{align*}
        R[f] 
        &= \E[\norm{f(X) - Y}^2] \\
        &= \E[\norm{\bar{f}(X) - \fopt(X) + f^\sperp(X)}^2] + \E[\norm{\xi}^2]\\
        &= \norm{\bar{f} - \fopt + f^\sperp}^2_\mu + \E[\norm{\xi}^2]\\
        &= \norm{\bar{f} - \fopt}_\mu^2 + \norm{f^\sperp}^2_\mu + \E[\norm{\xi}^2]\\
        &= R[\bar{f}] + \norm{f^\sperp}^2_\mu.
    \end{align*}
    Now consider the second model for $Y$.
    Let us write $\fopt_t(x) = \tilde{f}(x, t)$.
    Since $Y$ has finite second moment
    \[
        \infty > \E[\norm{Y}^2] 
        = \int_{[0, 1]} \norm{\fopt_t}_\mu^2 \dd t
    \]
    by Fubini's theorem so $\fopt_t \in \Lmuy$ for almost all $t \in [0, 1]$.
    As before, we can use \cref{lemma:l2-decomposition} to get the orthogonal decomposition
    $f = \bar{f} + f^\sperp$. Additionally, the equivariance of $\fopt_t$
    means $\inner{\fopt_t}{f^\sperp}_\mu = 0$. Hence
    \[
        R[f]
        = \E[\norm{f(X) - Y}^2] 
        = \E[\norm{f - \fopt_\eta}_\mu^2]
        = \E[\norm{\bar{f} - \fopt_\eta}_\mu^2] + \norm{f^\sperp}_\mu^2
        = R[\bar{f}] + \norm{f^\sperp}_\mu^2.
    \]
\end{proof}

The first model for $Y$ in \cref{lemma:generalisation} covers
the standard regression setup with an equivariant target function.
The second model, inspired by \parencite{bloem2020probabilistic},
gives conditional equivariance in distribution: 
$Y \vert{} gX \eqdist gY \vert{} X$ for all $g \in \G$ 
and is aimed at stochastic equivariant functions.
The first is a special case of the second when $\xi$ is invariant in distribution.
Additionally, \textcite{bloem2020probabilistic} show that
conditional invariance in distribution, i.e., $Y \vert{} g X \eqdist Y \vert{} X$ 
for all $g\in\G$, is equivalent to $Y \eqas \tilde{f}(X, \eta)$ for some $\tilde{f}$ 
that's invariant in its first argument. 
It's straightforward to derive a version of \cref{lemma:generalisation} for the
case that $\fopt$ or $\tilde{f}$ are approximately equivariant.

In later chapters we will use \Cref{lemma:generalisation}
to calculate explicitly the generalisation benefit of invariance/equivariance in 
random design least squares regression and kernel ridge regression.
We will see that $\norm{f^\sperp}_\mu^2$ displays a natural relationship
between the number of training examples and the dimension of the space of
$\G$-anti-symmetric predictors $A$, which is a property of the group action.
Intuitively, the learning algorithm needs enough examples to learn to be orthogonal to $A$.

\begin{remark}
    The same idea used in \cref{lemma:generalisation} can be used to give a lower
    bound on the excess risk in the misspecified case
    where the predictor $f$ is equivariant but the target $\fopt$ is not.
    For instance, under the first model in \cref{lemma:generalisation}
    \[
        R[f] - R[\fopt]
        = \norm{f - \qq \fopt}_\mu^2 + \norm{(\id - \qq)\fopt}_\mu^2
        \ge \norm{(\id - \qq)\fopt}_\mu^2.
    \]
\end{remark}

\subsection{Detour: Test-Time Augmentation}\label{sec:tta}
\emph{Test-time augmentation} consists of
averaging the output of a learned function $f$ over random transformations of
its input and can be used to increase test accuracy \parencite{Simonyan15, Szegedy15, He16}.
When the transformations belong to a group $\G$ and are sampled from its Haar measure,
test-time augmentation can be written as
\[
    \Ohat_n f(x) = \frac1n \sum_{i=1}^n f(G_i x) 
\]
where $G_1, \dots, G_n \sim \lambda$ are independent and identically distributed.
We can view $\Ohat_n f$ as an unbiased Monte-Carlo estimate of $\O f$.
For any bounded function $f$ and any $x$, $\Ohat f(x) \to \O f(x)$ 
$\lambda$-almost-surely by the strong law of large 
numbers \parencite[Theorem 4.23]{kallenberg2006foundations}.
By considering this limit, \cref{lemma:generalisation} hints at an explanation
for the generalisation improvement from test-time augmentation.

However, more work is needed to connect \cref{lemma:generalisation} to 
\parencite{Simonyan15, Szegedy15, He16}. 
In particular, \textcite{Simonyan15, Szegedy15, He16} average softmax outputs
rather than predictions and use classification accuracy to calculate the test error.

\chapter{The Linear Model}\label{chap:linear}
\section*{Summary}
In this chapter we apply \cref{lemma:l2-decomposition,lemma:generalisation} to 
linear models in a random design setting.
Although invariance is a special case of equivariance, 
we find it instructive to discuss separately the fully general result 
of least squares regression with an equivariant target 
and the special case of an invariant, scalar valued target.
These results are given in \cref{thm:equivariant-regression,thm:invariant-regression} 
respectively.
The work in this chapter provides the first proofs of a strict generalisation benefit of 
invariance/equivariance in each case.

\section*{Notation}
In this chapter we will write the actions $\phi$ and $\psi$ explicitly.
In particular we write
\[
    \qq f(x) = \int_\G \psi(g^{-1})f(\phi(g)x)\dd\lambda(g),
\]
and analogously for $\O$.

\section{Regression with an Invariant Target}\label{sec:invariant-regression}
Let $\X = \R^d$ with the Euclidean inner product and $\Y = \R$ with
multiplication.
Consider linear regression with the squared-error loss $\ell(y, y') = (y - y')^2$.
Let $\G$ act on $\X$ via an orthogonal representation $\phi: \G \to \orth_d$
and let $X \sim \mu$ be such that $\Sigma \coloneqq \E[X X^\top]$ is 
finite and positive definite.%
\footnote{If $\Sigma$ is only positive semi-definite then the developments are similar.}
We consider linear predictors $h_w: \X \to \Y$ with $h_w(x) = w^\top x$ where $w \in \X$.
Define the space of all linear predictors $\vlin = \{h_w : w \in \X\}$ which is a subspace of 
$\Lmu$. Notice that $\vlin$ is closed under $\O$: for all $x \in \X$
\begin{align*}
    \O h_w (x) 
        &= \int_\G  h_w(gx) \dd\lambda(g) \\
        &= \int_\G w^\top \phi(g) x \dd\lambda(g)\\
        &= \left(\int_\G \phi(g^{-1}) w\dd\lambda(g)\right)^\top x \\
        &= h_{\proj(w)}(x)
\end{align*}
where we substituted $g \mapsto g^{-1}$ and defined the 
linear map $\proj: \R^d \to \R^d$ by $\proj(w) = \int_\G  \phi(g) w\dd \lambda(g)$.
We also have
\[
    \inner{h_a}{h_b}_\mu = \int_\X a^\top xx^\top b\dd\mu(x)  = a^\top \Sigma b.
\]
We denote the induced inner product on $\X$ by $\inner{a}{b}_\Sigma \coloneqq a^\top \Sigma b$
and the corresponding norm by $\norm{\cdot}_\Sigma$.
Since $\vlin$ is closed under $\O$ we can apply \Cref{lemma:l2-decomposition}
to decompose $\vlin = S \oplus A$ with the orthogonality with respect to
$\inner{\cdot}{\cdot}_\Sigma$.
It follows that we can write any $h_w \in \vlin$ as
\[
    h_w = \overline{h_w} + h_w^\sperp
\]
where we have shown that there must exist $\bar{w}, w^\sperp \in \X$ with
$\inner{\bar{w}}{w^\sperp}_\Sigma = 0$ such that $\overline{h_w} = h_{\bar{w}}$
and $h_w^\sperp = h_{w^\sperp}$.
There is an isomorphism $\X \to \vlin$ where $w \mapsto h_w$.
Using this identification, we abuse notation slightly and write $\X = S \oplus A$ 
to represent the induced structure on $\X$.

Recall the definition of the \emph{risk} of a predictor $f$
\[
    R[f] = \E[\norm{f(X) -Y}^2].
\]
The risk is implicitly conditional on any randomness in $f$, e.g., occurring from its dependence
on training data.
We will refer to the difference in risk between two predictors as the \emph{generalisation gap}.
So the generalisation gap between $f$ and $f'$ is $R[f] - R[f']$. If this quantity is positive,
then we have strictly better test performance from $f'$. This gives a method of
comparing predictors.

Suppose examples are labelled by a target function $h_\theta \in \vlin$ that is $\G$-invariant.
Let $X \sim \mu$ and $Y = \theta^\top X + \xi$ where $\xi$ is independent of $X$,
has mean 0 and finite variance.
We calculate the difference in risk between $h_w$ and its invariant version
$h_{\bar{w}}$. \cref{lemma:generalisation} gives
\begin{equation}\label{eq:linear-invariant-gen-gap}
    R[h_w] - R[h_{\bar{w}}] = \norm{h_{w^\sperp}}_\mu^2 = \norm{w^\sperp}_\Sigma^2
\end{equation}
In \Cref{thm:invariant-regression} we calculate this quantity exactly.
where $w$ is the minimum-norm least squares estimator and $\bar{w} =\Phi_\G(w)$.
To the best of our knowledge, this is the first result to specify 
the generalisation benefit of invariant models.

\begin{theorem}\label{thm:invariant-regression}
    Let $\X = \R^d$, $\Y = \R$ and let $\G$ be a compact group with an orthogonal representation
    $\phi$ on $\X$. Let $X \sim \normal(0, \sigma_X^2 I)$ 
    and $Y = h_\theta(X) + \xi$ where $h_\theta(x) = \theta^\top x$ 
    is $\G$-invariant
    with $\theta\in\R^d$
    and where $\xi$ has mean $0$, variance $\sigma_\xi^2 <
    \infty$ and is independent of $X$. Let $w$ be the least squares estimate
    of $\theta$ from \iid~training examples $((X_i, Y_i): i=1, \dots, n)$ 
    distributed independently of and identically to $(X, Y)$
    and let
    $A$ be the orthogonal complement of the subspace of $\G$-invariant 
    linear predictors (as in \Cref{lemma:l2-decomposition}).
    \begin{itemize}
        \item If $n > d + 1$ then the generalisation gap satisfies
            \[
                \E\left[R[h_w] - R[h_{\bar{w}}]\right] = \sigma_\xi^2\frac{\dim A}{n - d - 1}.
            \]
        \item At the interpolation threshold $n \in [d-1, d+1]$, if $h_w$ is
            not $\G$-invariant then the generalisation gap diverges to
            $\infty$.
        \item If $n < d- 1$ the generalisation gap is 
            \[
                \E\left[R[h_w] - R[h_{\bar{w}}]\right] = 
                \dim A \left( \sigma_X^2 \norm{\theta}_2^2\frac{n(d-n)}{d(d-1)(d+2)}
                + \sigma_\xi^2\frac{n}{d(d - n - 1)} \right) .
            \]
    \end{itemize}
    The expectations are over the training sample.
    All of the above hold with $\ge$ when $h_{\bar{w}}$ is replaced by any predictor
    $h'$ such that $R[h'] \le R[h_{\bar{w}}]$ (see \cref{lemma:generalisation}).
\end{theorem}

\begin{proof}
    Note that $X$ is $\G$-invariant for all $\G$ since the representation $\phi$ is orthogonal.
    We have seen above that the space of linear maps $\vlin = \{h_w : w \in \R^d\}$ is closed under $\O$,
    so by \Cref{lemma:l2-decomposition} we can write $\vlin = S \oplus A$.
    Let $\proj^{A} = I - \proj$, which is the orthogonal projection onto the subspace $A$.
    By isotropy of $X$ and \cref{eq:linear-invariant-gen-gap} we have
    \[
        R[h_w] - R[h_{\bar{w}}] = \sigma_X^2 \norm{w^\sperp}_2^2
    \]
    for all $w \in \R^d$, where $w^\sperp = \proj^{A}(w)$.
    The proof consists of calculating this quantity in the case that $w$ is the
    least squares estimator.
        
    Let $\xx \in \R^{n \times d}$ and $\yy \in \R^n$ correspond to row-stacked training examples
    drawn \iid~as in the statement, so $\xx_{ij} = (X_i)_j$ and $\yy_i = Y_i$.
    Similarly, set $\bm{\xi} = \xx\theta - \yy$.
    The least squares estimate is the minimum norm solution 
    of $ \argmin_{u \in \R^d} \norm{\yy - \xx u}_2^2$,
    i.e.,
    \begin{equation}
        w = (\xx^\top\xx)^+ \xx^\top \xx \theta + 
        (\xx^\top \xx)^+ \xx^\top \bm{\xi} \label{eq:soln}
    \end{equation}
    where $(\cdot)^+$ denotes the Moore-Penrose pseudo-inverse.
    Define $P_{E} = (\xx^\top \xx)^+ \xx^\top \xx$,
    which is an orthogonal projection onto $E$, the
    rank of $\xx^\top \xx$ (this can be seen by diagonalising).

    We first calculate $\E[\norm{w^\sperp}_2^2 \mid{} \xx]$ 
    where $w^\sperp = \proj^{A}(w)$.
    The contribution from the first term of \cref{eq:soln} is
    \[
        \norm{\proj^{A}(P_E \theta)}_2^2,
    \]
    the cross term vanishes using $\xi \indep X$ and $\E[\xi] = 0$,
    and the contribution from the second term of \cref{eq:soln} is
    \[
        \E[\norm{\proj^{A}((\xx^\top \xx)^+ \xx^\top \bm{\xi}) }_2^2 \mid{} \xx] .
    \]
    $\proj^{A}$ is an orthogonal projection, so as a matrix is symmetric and
    idempotent. Hence, briefly writing $\proj^{A}$ without the parenthesis to emphasise 
    the matrix interpretation,
    \begin{align*}
        \E[\norm{\proj^{A}((\xx^\top \xx)^+ \xx^\top \bm{\xi}) }_2^2 \mid{} \xx] 
        &=  \E[  \tr( 
        \bm{\xi}^\top \xx  (\xx^\top \xx)^+\proj^{A}(\xx^\top \xx)^+ \xx^\top \bm{\xi}  
        )\mid{} \xx] \\
        &=  \tr( 
        \xx  (\xx^\top \xx)^+\proj^{A}(\xx^\top \xx)^+ \xx^\top \E[\bm{\xi}  \bm{\xi}^\top] 
        ) \\
        &=  \sigma_\xi^2\tr( \proj^{A}(\xx^\top \xx)^+).
    \end{align*}
    We have obtained
    \[
        \E[\norm{w^\sperp}_2^2\mid{} \xx]
        = \norm{\proj^{A} (P_E \theta )}_2^2 
        +  \sigma_\xi^2\tr( \proj^{A}((\xx^\top \xx)^+))
    \]
    and conclude by taking expectations, treating each term separately.
    \paragraph{First Term}
    If $n \ge d$ then $\dim E = d$ with probability $1$, so the first term 
    vanishes almost surely because $\proj^{A}(\theta)=0$.
    We treat the $n < d$ case using Einstein notation, 
    in which repeated indices are implicitly summed over.
    In components, recalling that $\proj^{A}$ is a matrix,
    \[
        \E[ \norm{\proj^{A} (P_E \theta )}_2^2 ] = \proj^{A}_{fa}\proj^{A}_{fc}\E[P_E \otimes P_E]_{abce} \theta_b \theta_e
    \]
    and applying \cref{lemma:proj-variance} we get
    \begin{align*}
        \E[ \norm{\proj^{A} (P_E \theta )}_2^2 ] 
        &= 
        \frac{n(d-n)}{d(d-1)(d+2)}
        \left(
            \proj^{A}_{fa}\proj^{A}_{fa} \theta_b \theta_b
            + \proj^{A}_{fa}\proj^{A}_{fb} \theta_b \theta_a
        \right)\\
        &\phantom{=}+ \frac{n(d-n) + n(n-1)(d+2)}{d(d-1)(d+2)} \proj^{A}_{fa}\proj^{A}_{fc}\theta_a \theta_c\\
        &= \norm{\theta}_2^2\dim A\frac{n(d-n)}{d(d-1)(d+2)}
    \end{align*}
    where we have used that $\proj^{A}(\theta) = 0$ and $\fnorm{\proj^{A}}^2 = \dim A$.

    \paragraph{Second Term}
    By linearity,
    \[
        \E[\tr( \proj^{A}((\xx^\top \xx)^+))] = \tr( \proj^{A}(\E[(\xx^\top \xx)^+])).
    \]
    Then \Cref{lemma:expected-inv-wishart-singular,lemma:expected-inv-wishart}
    give $\E[(\xx^\top\xx)^+] = \sigma_X^{-2} r(n,d)I_d$ where 
    \[
        r(n, d) = \begin{cases}
            {\frac{n}{d(d - n - 1)}} & n < d- 1\\
            (n - d - 1)^{-1} & n > d + 1 \\ 
            \infty & \text{otherwise}
        \end{cases}.
    \]
    When $n \in [d-1, d+1]$ it is well known that the expectation diverges, 
    see \cref{sec:wishart}. Hence
    \[
        \E[\tr( \proj^{A}((\xx^\top \xx)^+))] = \sigma_X^{-2} r(n, d)\dim A.
    \]
\end{proof}

In each case in \cref{thm:invariant-regression},
the generalisation gap has a term of the form $\sigma_\xi^2 r(n,
d) \dim A$ that arises due to the noise in the target distribution. 
In the overparameterised setting $d > n+1$ there is an additional term (the first)
that represents the generalisation gap in the noiseless setting $\xi \eqas 0$.
This term is the error in the least squares estimate of $\theta$ in the noiseless problem,
which of course vanishes in the fully determined case $n > d + 1$.
In addition, the divergence at the so called interpolation threshold $n\approx d$ 
is consistent with the literature on double
descent \parencite{hastie2022surprises}.

Notice the central role of $\dim A$ in \Cref{thm:invariant-regression}.
This quantity is a property of the group action as it describes the codimension
of the space of invariant models.
The generalisation gap is then dictated by how significant the symmetry is to
the problem.
We give two examples representing (non-trivial) extremal cases.

\begin{example}[Permutations, $\dim A = d - 1$]
    The matrix $\proj$ is invariant under the action of any $g' \in \G$
    \[
        \phi(g')\proj 
        = \phi(g')\int_\G \phi(g) \dd\lambda(g) 
        = \int_\G \phi(g'g) \dd\lambda(g)
        = \int_\G \phi(g) \dd\lambda(g)
        =\proj.
    \]
    In the case of $\sym_d$ with its representation as permutations matrices
    this implies $\Phi_{\sym_d} = c\bm{1} \bm{1}^\top$ for some
    $c$ that may depend on other quantities. This implies $\dim S = 1$ so $\dim A = d-1$ and
    the invariance $\Phi_{\sym_d} \bm{1} = \bm{1}$ gives $c=1/d$.
\end{example}

\begin{example}[Reflection, $\dim A = 1$]
    Let $\cyc_2$ be the cyclic group of order 2 and let it act on $\X = \R^d$ by reflection
    in the first coordinate.
    $A$ is then the subspace consisting of $w$ such that for all $j=1, \dots, d$
    \[
        \Phi_{\cyc_2}( w)_j = \frac{1}{\abs{\cyc_2}}\sum_{g\in \cyc_2} (\phi(g)w)_j = 0.
    \]
    Since the action fixes all coordinates apart from the first,
    $A = \{t(1, 0, \dots, 0)^\top: t \in \R\}$.
\end{example}

\section{Regression with an Equivariant Target}\label{sec:equivariant-regression}
One can apply the same construction to equivariant models.
Assume the same setup, but now let $\Y = \R^k$ with the Euclidean inner product and 
let the space of predictors be
$\wlin = \{f_W: \R^d \to \R^k, \,\, f_W(x) = W^\top x : W \in \R^{d \times k}\}$.
We consider linear regression with the squared-error loss $\norm{y - y'}_2^2$.
Let $X \sim \mu$ be such that $\Sigma \coloneqq \E[X X^\top]$ is 
finite and positive definite.
Let $\psi$ be an orthogonal representation of $\G$ on $\Y$.
We define the linear map, which we call the \emph{intertwining average},
$\twine :\R^{d \times k} \to \R^{d \times k} $ by%
    \footnote{The reader may have noticed that we define $\twine$ backwards, in the sense that its
        image contains maps that are equivariant in the direction $\psi \to \phi$.
        This is because of the transpose in the linear model,
        which is there for consistency with the $k=1$ invariance case.
    This choice is arbitrary and gives no loss in generality.}
\begin{equation}\label{eq:intertwine-linear}
    \twine(W) = \int_\G  \phi(g) W \psi(g^{-1})\dd \lambda(g).
\end{equation}
Similarly, define the \emph{intertwining complement} as $\twine^A :\R^{d \times k} \to \R^{d \times k} $ by
$
    \twine^A(W) = W - \twine(W)
$.
We establish the following results, which are generalisations of the invariant case.
In the proofs we will leverage the expression of $\twine$ as a $4$-tensor with components
\[
    \twine_{abce} = \int_\G  \phi(g)_{ac}\psi(g)_{be}\dd \lambda(g)
\]
where $a, c = 1, \dots d$ and $b, e = 1, \dots, k$.
This expression follows from the orthogonality of $\psi$ by
\[
    \twine(W)_{ab}
    = \int_\G \phi(g)_{ac} W_{ce}\psi(g^{-1})_{eb}\dd \lambda(g) 
    = \int_\G \phi(g)_{ac} W_{ce}\psi(g)_{be} \dd \lambda(g) 
    = \twine_{abce} W_{ce}.
\]

\begin{proposition}\label{prop:closure}
    $\forall f_W \in \wlin$: $\qq f_W = f_{\twine( W)}$ so $\wlin$ is closed under $\qq$.
\end{proposition}
\begin{proof}
    Let $f_W(x) = W^\top x$ with $W \in \R^{d \times k}$.
    By orthogonality and substituting $g\mapsto g^{-1}$ 
    \[
        \qq f_W (x) 
        = \int_\G  \psi(g^{-1}) W^\top \phi(g) x \dd \lambda(g)
        = \left(\int_\G  \phi(g) W \psi(g^{-1}) \dd \lambda(g)\right)^\top x
        = \twine(W)^\top x .
    \]
\end{proof}

\begin{proposition}\label{prop:inner-product}
    For all $f_{W_1}, f_{W_2} \in \wlin$,
    $\inner{f_{W_1}}{f_{W_2}}_\mu = \tr( W_1^\top \Sigma W_2)$.
\end{proposition}
\begin{proof}
    \begin{align*}
        \inner{f_{W_1}}{f_{W_2}}_\mu
        &= \int_\X  \left( W_1^\top x\right)^\top W_2^\top x \dd \mu (x)\\
        &= \int_\X  x^\top W_1 W_2^\top x \dd \mu (x)\\
        &= \int_\X  \tr(x^\top W_1 W_2^\top x) \dd \mu (x)\\
        &= \tr( W_1^\top \Sigma W_2)
    \end{align*}
\end{proof}

\Cref{prop:closure} allows us to apply \Cref{lemma:l2-decomposition} to 
write $\wlin = S \oplus A$,
so for all $f_W \in \wlin$
there exists $\overline{f_{W}} \in S$ and 
$f_W^\sperp \in A$ with $\inner{\overline{f_W}}{f_W^\sperp}_\mu = 0$.
The corresponding parameters $\overline{W} = \twine(W)$ and 
$W^\sperp = \twine^A(W)$ must therefore satisfy 
$\tr(\overline{W}^\top \Sigma W^\sperp) = 0$.
Repeating our abuse of notation, we identify 
$\R^{d \times k} = S \oplus A$ with $S = \twine(\R^{d \times k})$ and $A$ its orthogonal
complement with respect to the induced inner product.

\begin{proposition}\label{prop:generalisation-gap}
    Let $X \sim \mu$ and let $\xi$ be a random vector in $\R^k$ that is independent of
    $X$ with $\E[\xi] = 0$ and finite variance.
    Set $Y = h_\Theta (X) + \xi$ where $h_\Theta$ is $\G$-equivariant.
    For all $f_W \in \wlin$, the generalisation gap satisfies
    \[
        R[f_W] - R[f_{\overline{W}}]
        \coloneqq \E[\norm{Y - f_W(X)}_2^2] - \E[\norm{Y - f_{\overline{W}}(X)}_2^2] 
        = \fnorm{\Sigma^{1/2} W^\sperp}^2
    \]
    where $\overline{W} =\twine(W)$, $W^\sperp = \twine^A(W)$
    and $\Sigma = \E[XX^\top]$.
\end{proposition}
\begin{proof}
    Recall that $W = \overline{W} + W^\sperp$ and that these 
    satisfy $\tr(\overline{W}\Sigma W^\sperp) =0$ from the above.
    Then, using \Cref{lemma:generalisation} and \cref{prop:inner-product},
    \[
        R[f_W] - R[f_{\overline{W}}]
        = \norm{f_{W^\sperp}}_\mu^2 
        = \tr((W^\sperp)^\top \Sigma W^\sperp)
        = \fnorm{\Sigma^{1/2} W^\sperp}^2.
    \]
\end{proof}

Having followed the same path as the previous section,
we provide a characterisation of the generalisation benefit of equivariance.
In the same fashion, we compare the least squares estimate $W$ with its equivariant
version $\overline{W} = \twine(W)$. The choice of $\overline{W} = \twine(W)$ as
a comparator is natural. Indeed, following directly from
\cref{lemma:generalisation} it costs us nothing in terms of the strength of the
following result.

\begin{theorem}\label{thm:equivariant-regression}
    Let $\X = \R^d$, $\Y = \R^k$ and let $\G$ be a compact group with orthogonal representations
    $\phi$ on $\X$ and $\psi$ on $\Y$. Let $X \sim \normal(0, \sigma_X^2 I_d)$ 
    and $Y = h_\Theta(X) + \xi$ where $h_\Theta(x) = \Theta^\top x$ is $\G$-equivariant 
    and $\Theta \in \R^{d \times k}$.
    Assume $\xi$ is a random element of $\R^k$, independent of $X$,
    with mean $0$ and $\E[\xi \xi^\top ] = \sigma_\xi^2 I_k < \infty$.
    Let $W$ be the least squares estimate of $\Theta$ from $n$ \iid~training examples
    $((X_i, Y_i): i=1, \dots, n)$ 
    distributed independently of and identically to 
    $(X, Y)$
    and let 
    $\binner{\chi_\psi}{\chi_\phi} = \int_{\G} \chi_\psi(g) \chi_\phi(g)\dd \lambda(g) $
    denote the scalar product of the characters of the representations of $\G$.
    \begin{itemize}
        \item     If $n > d + 1$ the generalisation gap is 
    \[
        \E\left[R[f_W] - R[ f_{\overline{W}}]\right] =
        \sigma_\xi^2 \frac{ dk - \binner{\chi_\psi}{\chi_\phi}}{n - d - 1}. 
    \]
    \item At the interpolation threshold $n \in [d-1, d+1]$, if $f_W$ is not $\G$-equivariant then
    the generalisation gap diverges to $\infty$.
    \item If $n < d- 1$ then the generalisation gap is 
        \begin{align*}
            \E\left[R[f_W] - R[ f_{\overline{W}}]\right] 
            &=
            \sigma_X^2 \frac{n(d-n)}{d(d-1)(d+2)} \left( (d+1) \fnorm{\Theta}^2
                - \tr( J_\G \Theta^\top \Theta) \right) \\
            &\phantom{=}+ 
            \sigma_\xi^2 \frac{n( dk - \binner{\chi_\psi}{\chi_\phi})}{d (d - n - 1)}
        \end{align*}
    where each term is non-negative and 
    $J_\G \in \R^{k \times k}$ is given by
    \[
        J_\G = \int_\G (\chi_\phi(g) \psi(g) + \psi(g^2)) \dd \lambda(g).
    \]
    \end{itemize}
    All of the above hold with $\ge$ when $f_{\overline{W}}$ is replaced by any predictor
    $f'$ such that $R[f'] \le R[f_{\overline{W}}]$ (see \cref{lemma:generalisation}).
\end{theorem}

\begin{proof}
    We use Einstein notation, in which repeated indices are summed over.
    Since the representation $\phi$ is orthogonal, $X$ is $\G$-invariant for all $\G$.
    We have seen from \Cref{prop:generalisation-gap} that 
    \[
        \E\left[ R[f_W] - R[f_{\overline{W}}] \right]
        = \sigma_X^2\E[ \fnorm{W^\sperp}^2]
    \]
    and we want to calculate this quantity for the least squares estimate
    \[
        W = (\xx^\top \xx)^+ \xx^\top \yy =  (\xx^\top \xx)^+ \xx^\top \xx \Theta 
        +  (\xx^\top \xx)^+ \xx^\top \xxi
    \]
    where $\xx \in\R^{n \times d}$, $\yy \in \R^{n \times k}$ are the row-stacked 
    training examples
    with $(\xx)_{ij} = (X_i)_j$, $(\yy_i)_j = (Y_i)_j$ and $\xxi = \yy - \xx \Theta$.
    We have
    \begin{align*}
        &\E\left[ R[f_W] - R[f_{\overline{W}}]\right]\\
        &= \sigma_X^2\E[ \fnorm{\twine^A(W)}^2]\\
        &= \sigma_X^2\E[ \fnorm{\twine^A((\xx^\top \xx)^+ \xx^\top \xx \Theta
        +  (\xx^\top \xx)^+ \xx^\top \xxi)}^2]\\
        &= \sigma_X^2\E[ \fnorm{\twine^A((\xx^\top \xx)^+ \xx^\top \xx \Theta)}^2] 
        + \sigma_X^2\E[\fnorm{\twine^A ((\xx^\top \xx)^+ \xx^\top \xxi)}^2]
    \end{align*}
    using linearity and $\E[\xxi] = 0$. We treat the two terms separately,
    starting with the second.

    \paragraph{Second Term}
    Setting $\zz = (\xx^\top \xx)^+ \xx^\top$ we have 
    \[
        \E[\fnorm{\twine^A (\zz \xxi)}^2]
        = \E[\tr(\twine^A (\zz \xxi)^\top \twine^A (\zz\xxi))].
    \]
    One gets
    \begin{align}
        \E[\tr(\twine^A (\zz \xxi)^\top \twine^A (\zz\xxi))]
        &= 
        \E[\twine^A_{abcj} \zz_{ce}\xxi_{ej} \twine^A_{abfg} \zz_{fh}\xxi_{hg}]
        \nonumber \\ 
        &= \sigma_\xi^2 
        \E[\twine^A_{abcj} \zz_{ce} \twine^A_{abfg} \zz_{fh} \delta_{eh}\delta_{jg}] 
            \nonumber\\ 
        &= \sigma_\xi^2 \twine^A_{abcj} \twine^A_{abfj} \E[\zz_{ce}  \zz_{fe} ]
        \nonumber\\
        &= \sigma_\xi^2 \twine^A_{abcj} \twine^A_{abfj} \E[(\zz \zz^\top)_{cf}]
        \label{eq:dagger} 
    \end{align}
    and then (WLOG relabelling $f \mapsto e$)
    \begin{align*}
        \twine^A_{abcj} \twine^A_{abej}
        &= \left(\delta_{ac}\delta_{bj} - \int_\G\phi(g)_{ac}\psi(g)_{bj}\dd \lambda(g) \right) 
           \left(\delta_{ae}\delta_{bj} - \int_\G\phi(g)_{ae}\psi(g)_{bj}\dd \lambda(g) \right)  
           \\ 
        &= \delta_{ac}\delta_{bj}\delta_{ae}\delta_{bj}                                     
         -  \delta_{ae}\delta_{bj}\int_\G\phi(g)_{ac}\psi(g)_{bj}\dd \lambda(g) \\
        &\phantom{=} - \delta_{ac}\delta_{bj}\int_\G  \phi(g)_{ae}\psi(g)_{bj} \dd \lambda(g) \\
        &\phantom{=}+ \int_\G \phi(g_1)_{ac} \phi(g_2)_{ae} \psi(g_1)_{bj}\psi(g_2)_{bj}
        \dd \lambda(g_1) \dd \lambda(g_2) \\
        &= k\, \delta_{ce} - \int_\G \tr(\psi(g)) ( \phi(g)_{ec} + \phi(g)_{ce})\dd \lambda (g) \\
        &\phantom{=}+ \int_\G \tr(\psi(g_1)^\top \psi(g_2)) (\phi(g_1)^\top \phi(g_2))_{ce} 
        \dd \lambda(g_1) \dd \lambda(g_2) 
    \end{align*}
    where we have used that the indices $b, j = 1,\dots, k$. Consider the final term
    \begin{align*}
        & \int_\G \tr(\psi(g_1)^\top \psi(g_2)) (\phi(g_1)^\top \phi(g_2))_{ce} 
         \dd \lambda(g_1) \dd \lambda(g_2) \\
         &=  \int_\G \tr(\psi(g_1^{-1}g_2))(\phi(g_1^{-1}g_2))_{ce} 
         \dd \lambda(g_1) \dd \lambda(g_2)  \\ 
         &= \int_\G \tr(\psi(g)) \phi(g)_{ce}\dd \lambda(g)  
    \end{align*}
    where we used that the representations are orthogonal, Fubini's theorem and
    that the Haar measure is invariant.
    Now we put things back together.
    To begin with
    \[
        \twine^A_{abcj} \twine^A_{abej} 
        = k\, \delta_{ce} - \int_\G \tr(\psi(g)) \phi(g^{-1})_{ce}\dd \lambda (g) 
    \]
    and putting this into \cref{eq:dagger} with $\zz \zz^\top = (\xx^\top \xx)^{+}$ gives
    \[
        \E[\tr(\twine^A (\zz \xxi)^\top \twine^A (\zz\xxi))]
        = \sigma_\xi^2 \left( k\, \delta_{ce} - \int_\G  
        \tr(\psi(g)) \phi(g^{-1})_{ce} \dd \lambda (g)\right) \E[(\xx^\top\xx)^+_{ce}] 
    \]
    where $c, e = 1, \dots, d$.
    Applying \Cref{lemma:expected-inv-wishart,lemma:expected-inv-wishart-singular}
    gives $\E[(\xx^\top\xx)^+_{ce}] = \sigma_X^{-2} r(n,d)\delta_{ce}$ where 
    \[
        r(n, d) = \begin{cases}
            {\frac{n}{d(d - n - 1)}} & n < d- 1\\
            (n - d - 1)^{-1} & n > d + 1 \\ 
            \infty & \text{otherwise}
        \end{cases}.
    \]
    When $n \in [d-1, d+1]$ it is well known that the expectation diverges, 
    see \Cref{sec:wishart}.
    Using the orthogonality of $\phi$ we arrive at
    \begin{align*}
        \sigma_X^2\E[\fnorm{\twine^A ((\xx^\top \xx)^+ \xx^\top \xxi)}^2] 
        &= \sigma_\xi^2 r(n, d) \left(dk - \int_\G  \tr(\psi(g)) \tr(\phi(g))
        \dd \lambda (g)\right) \\
        &= \sigma_\xi^2 r(n, d) \left(dk - \binner{\chi_\phi}{\chi_\psi}\right) 
    \end{align*}

    \paragraph{First Term}
    If $n \ge d$ then $(\xx^\top \xx)^+ \xx^\top \xx \Theta \eqas \Theta$ and 
    since $h_\Theta \in S$ the first term vanishes almost surely.
    This gives the case of equality in the statement.
    If $n < d$ we proceed as follows.
    Write $P_E = (\xx^\top \xx)^+ \xx^\top \xx$ which is the orthogonal projection onto
    the rank of $\xx^\top \xx$.
    By isotropy of $X$, $E \sim \unif \Gr_n(\R^d)$ with probability 1.\footnote{%
    Regarding \cref{remark:other-covariates},
    this is where absolute continuity with
    respect to the Lebesgue measure is required.}
    Recall that $\twine^A(\Theta) = 0$, which in components reads
    \begin{equation}\label{eq:null}
        \twine^A_{abce}\Theta_{ce} = 0 \qquad \forall a, b.
    \end{equation}
    Also in components, we have
    \[
        \E[ \fnorm{\twine^A((\xx^\top \xx)^+ \xx^\top \xx \Theta)}^2] 
        = \twine^A_{fhai}\twine^A_{fhcj} 
        \E[P_E \otimes P_E]_{abce} \Theta_{bi} \Theta_{ej}
    \]
    and using \cref{lemma:proj-variance} we get
    \begin{align}
        \begin{split}
         &\E[ \fnorm{\twine^A((\xx^\top \xx)^+ \xx^\top \xx \Theta)}^2] 
        \label{eq:linear-equivariance-bias}\\
        &= \frac{n(d-n)}{d(d-1)(d+2)}
         \twine^A_{fhai}\twine^A_{fhaj} \Theta_{bi} \Theta_{bj} \\
         &\phantom{=}+ \frac{n(d-n)}{d(d-1)(d+2)}
         \twine^A_{fhai}\twine^A_{fhbj} \Theta_{bi} \Theta_{aj} \\ 
        &\phantom{=} + \frac{n(d-n) + n(n-1)(d+2)}{d(d-1)(d+2)}
         \twine^A_{fhai}\twine^A_{fhcj} \Theta_{ai} \Theta_{cj}.
        \end{split}
    \end{align}
    The third term vanishes by \cref{eq:null}.
    Consider the remaining two terms separately. 
    Start with the first term of \cref{eq:linear-equivariance-bias},
    in which
    \[
         \twine^A_{fhai}\twine^A_{fhaj} \Theta_{bi} \Theta_{bj} 
          = (\Theta^\top \Theta)_{ij}\twine^A_{fhai}\twine^A_{fhaj}
    \]
    where
    \begin{align*}
        \twine^A_{fhai}\twine^A_{fhaj}
        &= 
        \left(\delta_{fa}\delta_{hi} - \int_\G \phi(g)_{fa}\psi(g)_{hi}\dd \lambda(g) \right) 
        \left(\delta_{fa}\delta_{hj} - \int_\G \phi(g)_{fa}\psi(g)_{hj}\dd \lambda(g) \right) 
        \\ 
        &= 
        d \delta_{ij} - \int_\G \tr (\phi(g)) \psi(g)_{ij} \dd \lambda(g)
        - \int_\G \tr (\phi(g)) \psi(g)_{ji} \dd \lambda(g)
        \\
        &\phantom{=}
        + \int_\G \phi(g_1)_{fa}\phi(g_2)_{fa} \psi(g_1)_{hi} \psi(g_2)_{hj} 
        \dd \lambda(g_1)\dd \lambda(g_2) \\ 
        &= d \delta_{ij} - \int_\G \tr (\phi(g)) \psi(g)_{ji} \dd \lambda(g)
    \end{align*}
    using the orthogonality of the representations and invariance of the Haar measure
    (the calculation is the same as for the first term).
    Therefore
    \begin{align*}
         \twine^A_{fhai}\twine^A_{fhaj} \Theta_{bi} \Theta_{bj} 
        &= 
        d \fnorm{\Theta}^2 - \int_\G \chi_\phi(g)\tr\left(\psi(g^{-1}) \Theta^\top \Theta\right)
        \dd \lambda(g) \\ 
        &= d \fnorm{\Theta}^2 - \int_\G \chi_\phi(g)\tr\left(\psi(g) \Theta^\top  \Theta\right) 
        \dd \lambda(g) .
    \end{align*}
    Now for the second term of \cref{eq:linear-equivariance-bias}
    \begin{align*}
        &\Theta_{bi} \Theta_{aj}\twine^A_{fhai}\twine^A_{fhbj}\\
        &= \Theta_{bi} \Theta_{aj} \left(\delta_{fa}\delta_{hi} 
        -  \int_\G  \phi(g)_{fa}\psi(g)_{hi}\dd \lambda(g) \right) 
        \left(\delta_{fb}\delta_{hj} 
        - \int_\G  \phi(g)_{fb}\psi(g)_{hj}\dd \lambda(g) \right)  
        \\ 
        &= \Theta_{bi} \Theta_{aj}\left(\delta_{ab} \delta_{ij} 
        - \int_\G \phi(g)_{ab}\psi(g)_{ij} \dd \lambda(g) \right) \\
        &= \fnorm{\Theta}^2 - \int_\G \tr( \Theta^\top \phi(g) \Theta \psi(g) ) 
        \dd \lambda(g) \\
        &= \fnorm{\Theta}^2 - \int_\G \tr(\psi(g^2) \Theta^\top \Theta ) \dd \lambda(g).
    \end{align*}
    Putting these together gives 
    \[
      \E[ \fnorm{\twine^A((\xx^\top \xx)^+ \xx^\top \xx \Theta)}^2] 
      = \frac{n(d-n)}{d(d-1)(d+2)} \left( (d+1) \fnorm{\Theta}^2- 
          \tr( J_\G \Theta^\top  \Theta) \right) 
    \]
    where $J_\G \in \R^{k \times k}$ is the matrix-valued function of $\G$, $\psi$ and $\phi$
    \[
        J_\G = \int_\G (\chi_\phi(g) \psi(g) + \psi(g^2)) \dd \lambda(g).
    \]
\end{proof}
\Cref{thm:equivariant-regression} is a direct generalisation of \Cref{thm:invariant-regression}. 
As we remarked in the introduction, $dk - \binner{\chi_\psi}{\chi_\phi}$ 
plays the role of $\dim A$ 
in \Cref{thm:invariant-regression} and is a measure of the significance of the
symmetry to the problem.
The dimension of $\wlin$ is $dk$, while
$\binner{\chi_\psi}{\chi_\phi}$ is 
the dimension of the space of equivariant maps.
In our notation $\binner{\chi_\psi}{\chi_\phi} = \dim S$.

Just as with \cref{thm:invariant-regression}, there is an additional term (the first)
in the overparameterised case $d > n + 1$ that represents the estimation error
in the noiseless setting $\xi \eqas 0$.
Notice that if $k=1$ and $\psi$ is trivial we find
\[
    J_\G = \int_\G \chi_\phi(g) \dd \lambda(g)  + 1 =  \binner{\chi_\phi}{1} + 1 = \dim S + 1
\]
which confirms that \cref{thm:equivariant-regression} reduces exactly to \cref{thm:invariant-regression}.

Interestingly, the first term in the $d > n+1$ case can be made independent of $\psi$,
since the equivariance of
$h_\Theta$ implies
\[
    \tr(J_\G \Theta^\top \Theta) = \tr(\Theta^\top J_\phi \Theta)
\]
where
\[
    J_\phi = \int_\G (\chi_\phi(g) \phi(g) + \phi(g^2)) \dd \lambda(g).
\]

\begin{remark}\label{remark:other-covariates}
    Versions of \cref{thm:equivariant-regression} are 
    possible for other probability distributions on $X$.
    For instance, a similar result would hold for any isotropic distribution that 
    is absolutely continuous with respect to the Lebesgue measure 
    and has finite variance.
    The isotropy implies the existence of a scalar $r$ 
    (which depends on $n$ and the distribution of $X$)
    such that $\E[(\xx^\top \xx)^+] = r I_d$
    where $\xx \in \R^{n \times d}$ are the row-stacked training inputs as defined in the proof.
\end{remark}

\chapter{Kernel Methods}\label{chap:kernels}
\section*{Summary}
In this chapter we apply and extend the results of \cref{chap:general-theory-i} to reproducing
kernel Hilbert spaces, focussing on invariance rather than equivariance.
We go a step further than the previous chapter on 
linear models, applying \cref{lemma:l2-decomposition}
to establish \cref{thm:kernel-invariance}, which gives a strict generalisation benefit
for invariance in kernel ridge regression when the symmetry is also present
in the target. 
As with \cref{chap:linear}, we study the random design setting.
This result is specialised to the linear kernel and the orthogonal group,
showing a fundamental connection to \cref{thm:invariant-regression}.
The other main developments of this chapter are more abstract.
\Cref{thm:kernel-invariance} gives a lower bound for the generalisation
of kernel ridge regression with an invariant target. By studying the result,
we uncover a condition on the kernel under which this lower
bound vanishes in the $n\to\infty$ limit. 
Assuming that this condition holds, we derive an independent
result in \cref{thm:rkhs-decomposition}
that provides a decomposition of a reproducing kernel Hilbert space into 
orthogonal $\G$-symmetric and $\G$-anti-symmetric parts,
analogous to \cref{lemma:l2-decomposition}.

\section{Background, Assumptions and Preliminaries}
A function $k: \X\times\X\to\R$ is \emph{positive definite} if, for all
$n$ and all
distinct
$x_1, \dots, x_n \in \X$ the matrix with components
$K_{ij} = k(x_i, x_j)$ is positive semi-definite.
For $\xx = (x_1, \dots, x_n) \in \X^n$
we write $f(\xx)\in \R^n$ as the vector with elements $f(\xx)_i = f(x_i)$.
For any Borel measure $\nu$ we write
$\supp \nu$ for the support of $\nu$, which is the smallest closed subset of 
$\X$ whose complement has measure 0.
We will say that a kernel $k:\X\times\X \to \R$ is invariant if 
$k(x, gx') = k(x, x')$
for all $g \in \G$ and all $x, x' \in \X$.
For any measurable $j: \X\times\X\to\R$ we define
\[
    \norm{j}^2_\Lmumu
    =
    \int_\X j(x, y)^2\dd\mu(x)\dd\mu(y).
\]

\subsection{The Basics of Reproducing Kernel Hilbert Spaces}\label{sec:rkhs-basics}
A \emph{Hilbert space} is an inner product space that is complete
with respect to the norm topology induced by the inner product.
We will only consider spaces over $\R$.
A \emph{reproducing kernel Hilbert space} (RKHS) $\H$ is a Hilbert space of real functions 
$f: \X \to \R$ on which the evaluation
functional $\delta_x : \H \to \R$ with $\delta_x[f] = f(x)$ is continuous 
for all $ x \in \X$ or, equivalently, is a bounded operator.
The Riesz representation theorem tells us that
there is a unique function $k_x \in \H$ such that 
$
    \delta_x[f] = \inner{f}{k_x}_\H
$
for all $f \in \H$,
where $\inner{\cdot}{\cdot}_\H: \H \times \H \to \R$ is the inner product on $\H$.
We will refer to the function $k_x$ as the \emph{representer (of evaluation at $x$)}.
We identify the function $k : \X \times \X \to \R$ with $k(x, y) = \inner{k_x}{ k_y}_\H$
as the \emph{(reproducing) kernel} of $\H$.
Using the inner product representation, one can see that $k$ is 
positive definite and symmetric.
Conversely, the Moore-Aronszajn theorem 
states that
for any positive definite and symmetric function $k$, there is a unique
RKHS with reproducing kernel $k$ \parencite{aronszajn1950theory}.
In addition, any Hilbert space of functions admitting a reproducing kernel is an RKHS.
Finally, another characterisation of $\H$ is as the completion (with respect to
the norm topology) of the space of linear combinations 
$
    f_c(x) = \sum_{i=1}^n c_i k(x, x_i)
$
for $c_1, \dots, c_n \in \R$ and $x_1, \dots, x_n \in \X$ with the inner product
between $f_{c}$ and $f_{\tilde{c}}$ being 
$\sum_{i, j=1}^n c_i \tilde{c}_jk(x_i, x_j)$.
For (many) more details, see \parencite[Chapter 4]{steinwart2008support}.

\subsection{Assumptions}\label{sec:technical}
In this chapter we make the additional assumption that $\supp \mu = \X$.
The output space will be $\Y = \R$.
Let $k: \X \times \X \to \R$ be a measurable kernel with RKHS $\H$
such that $k(\cdot, x):\X \to \R$ is continuous for all $x \in \X$.
Assume that $\sup_{x \in \X} {k(x, x)} = M_k < \infty$ and note that this implies that
$k$ is bounded since 
\[
    k(x, x') = \inner{k_x}{k_{x'}}_\H \le \norm{k_x}_\H \norm{k_{x'}}_\H 
    = \sqrt{k(x, x)}\sqrt{k(x', x')} \le M_k.
\]
Every $f \in \H$ is $\mu$-measurable, bounded and continuous
by \cref{lemma:rkhs-measurable,lemma:rkhs-continuous}.
Recall that $\X$ is assumed to be a Polish space, so it is separable
and $\H$ is separable using \cref{lemma:rkhs-separable}.

\subsection{Relating $\H$ to $\Lmu$}
Define the integral operator $S_k: \Lmu \to \H$
\[
    S_k f(x) = \int_\X k(x, x') f(x')\dd \mu (x').
\]
$S_k$ assigns a function in $\H$ to every element of $\Lmu$.
On the other hand, every $f \in \H$ is measurable and bounded so has $\norm{f}_\mu < \infty$
and belongs to some element of $\Lmu$.
We write $\iota: \H \to \Lmu$ for the inclusion map that sends
$f$ to the element of $\Lmu$ that contains $f$.
By \cref{thm:rkhs-integral-op-1}, $S_k$ is well-defined and
$\iota$ is its adjoint.
Both $\iota$ and $S_k$ are bounded with operator norms bounded
by $M_k$. This is immediate from \cref{thm:rkhs-integral-op-1} 
or follows directly from
\[
    \norm{\iota f}_\mu^2 = \int_\X f(x)^2 \dd\mu(x) 
    = \int_\X \inner{k_x}{ f}_\H^2 \dd\mu(x)
    \le \int_\X \norm{k_x}_\H^2\norm{f}_\H^2 \dd\mu(x)
    \le M_k \norm{f}_\H^2
\]
for all $f\in\H$ and 
$\opnorm{S_k} = \opnorm{\iota}$ by \cref{thm:adjoint}.
Define $T_k : \Lmu \to \Lmu$ by $T_k= \iota \circ S_k$
and it is easy to check that $T_k$ is self-adjoint,
$\opnorm{T_k} \le M_k$ and that $\inner{T_k f}{f} \ge 0$ for all
$f\in\Lmu$.

We will make use of the following fact, which is both standard
and easy to see.

\begin{lemma}\label{lemma:dense-image}
    The image of $\Lmu$ under $S_k$ is dense in $\H$
    and $\iota$ is injective, so any element of $\Lmu$ contains at 
    most one $f \in \H$.
\end{lemma}
\begin{proof}
    \Cref{thm:rkhs-integral-op-1} says that $S_k (\Lmu)$ is dense in $\H$
    if and only if $\iota$ is injective.
    Injectivity of $\iota$ is equivalent to the statement that
    for all $f, f'\in \H$ the set
    $
        A(f, f') = \{ x \in \X: f(x) \ne f'(x)\}
        $
    has $A \ne \emptyset \implies \mu(A) > 0$.
    Continuity implies that for all $f, f' \in \H$, either $f = f'$ pointwise or $A(f, f')$
    contains an open set.
    By the support of $\mu$ this implies $\mu(A) > 0$: if there was a non-empty open set
    $B$ with $\mu(B) = 0$ then $\X \setminus B$ is a closed proper subset of $\X$, 
    contradicting $\supp \mu = \X$.
    Thus, $\iota$ is injective.
\end{proof}

Note that we do not assume that $\X$ is compact.
This allows for application to common settings such as $\X = \R^n$
but prevents the application of Mercer's theorem.
We work around this, but for generalisations of Mercer's theorem 
see \parencite{steinwart2012mercer} and references therein.

\subsection{Orbit Averaging}
Recall the definition of the averaging operator for invariance
$\O : \Lmu \to \Lmu$ as
\[
    \O f(x) = \int_\G f(gx) \dd \lambda (g).
\]
Note that we have not yet defined $\O$ as operator on $\H$.
In \cref{sec:structure} we give an additional condition under which
$\O: \H \to \H$ is well defined and, as a consequence, an analogy 
of \cref{lemma:l2-decomposition} for RKHSs.
For all $x \in \X$ we define
$\kbar{x} = \O \iota k_x$ and $k_x^\sperp = \iota k_x - \kbar{x}$. 
We also define $\overline{k}:\X\times\X\to\R$ and $k^\sperp: \X\times\X\to\R$
by
\[
    \overline{k}(x, y) 
    %\coloneqq \kbar{x}(y)
    = \int_\G k(x, gy) \dd\lambda(g)
\]
and $k^\sperp(x, y) \coloneqq k(x, y) - \overline{k}(x, y)$. 
Note that the averaging is only over one argument.
The function $k_x[y]: \G\to\R$ with $k_x[y](g) = k(x, gy)$ is $\lambda$-measurable
by \cref{lemma:section} and a simple composition argument as in \cref{prop:q-well-defined};
it is also bounded so $\overline{k}$ exists, is finite and itself is $\mu$-measurable
by \cref{lemma:section}.
For all $x\in\X$ we have $\overline{k}(x, y) = \kbar{x}(y)$ $\mu$-almost-surely in $y$.
On the other hand, the functions $\overline{k}(x, \cdot)$ are not
necessarily in $\H$.

\section{Generalisation}\label{sec:kernel-generalisation}
In this section we apply the theory developed in \cref{chap:general-theory-i} to study
the impact of invariance on kernel ridge regression with an invariant target.
We analyse the generalisation benefit of invariance, finding
a strict benefit when the symmetry is present in the target.

\subsection{Kernel Ridge Regression}
Given observations $((x_i, y_i): i=1, \dots, n)$ where
$x_i \in \X$ and $y_i\in \R$, \emph{kernel ridge regression} (KRR) 
returns a predictor
that solves the optimisation problem
\begin{equation}\label{eq:krr}
    \argmin_{f \in \H} \;\sum_{i=1}^n (f(x_i) - y_i)^2 + \rho \norm{f}_\H^2
\end{equation}
and $\rho > 0$ is the regularisation parameter.
In general $\rho$ is a function of $n$, but we leave
this dependence implicit to save on notation.
KRR can be thought of as performing ridge regression in a
possibly infinite dimensional feature space $\H$.
The solution to this
problem is of the form
$
    f(x) = \sum_{i=1}^n \alpha_i k_{x_i}(x)
    $
where $\alpha \in \R^n$ solves 
\begin{equation}\label{eq:krr-least-squares}
    \argmin_{\alpha \in \R^n} \; \norm{\yy - K\alpha}_2^2 + \rho\alpha^\top K\alpha
    ,
\end{equation}
$\yy \in \R^n$ is the standard row-stacking of the training outputs with
$\yy_i = y_i$ and
$K$ is the kernel Gram matrix with $K_{ij} = k(x_i, x_j)$.
The form of the solution is an immediate consequence of the
representer theorem \parencite{kimeldorf1970correspondence,scholkopf2001generalized},
or can be seen directly by expanding
$f$ in terms of its components in and orthogonal
to the span of $\{k_{x_1}, \dots, k_{x_n}\}$ and 
substituting into \cref{eq:krr}.
We consider solutions of the form
$
    \alpha = (K + \rho I)^{-1}\yy
$
which results in the predictor
\begin{equation}\label{eq:krr-solution}
    f(x) = k_{\xx}(x)^\top (K + \rho I)^{-1}\yy
\end{equation}
where $k_{\xx}: \X \to \R^n$ has values $k_{\xx}(x)_i = k_{x_i}(x) = k(x_i, x)$.
It's easy to see that this solution is unique.
If $K$ is a positive definite matrix then of course $\alpha$ is the 
unique solution to \cref{eq:krr-least-squares}.
On the other hand suppose $K$ is degenerate and let $\beta$ be the component
of $\alpha$ in the null space of $K$, then 
$0= \beta^\top K \beta = \norm{\sum_{i=1}^n \beta_i k_{x_i}}_\H^2 $
so $f$ is independent of $\beta$.

In calculating the generalisation benefit of invariance, we will use as a 
comparator its averaged version
\[
    \bar{f}(x) = \O \iota f(x) =  \kbar{\xx}(x)^\top (K + \rho I)^{-1}\yy
\]
which is $\G$-invariant.
Once again, in the proof we will compare the risks of $f$ and $\bar{f}$,
recalling the definition
of the risk of $f$, for any random variables $(X, Y)$, as
\[
    R[f] = \E[\norm{f(X) - Y}_2^2]
\]
with the expectation conditional on $f$ if it is random.
Just as with \cref{lemma:generalisation}, the choice of $\bar{f}$
as a comparator costs us nothing and 
this choice is also natural: $\bar{f}$ is the closest invariant predictor to 
$f$ in $\Lmu$ (see \cref{prop:fa-least-squares}).

\subsection{Generalisation Benefit of Invariance}\label{sec:kernel-invariance}
In this section we give a characterisation
of the generalisation benefit of invariance in kernel ridge regression.
\Cref{thm:kernel-invariance} shows that invariance improves generalisation in kernel
ridge regression when the conditional mean of the target distribution is invariant.
In a sense, \cref{thm:kernel-invariance}
is a generalisation of \cref{thm:invariant-regression}
and we will return to this comparison later.

\begin{theorem}\label{thm:kernel-invariance}
    Let the training sample be $((X_i, Y_i): i=1, \dots, n)$ 
    \iid~with $X_i \sim \mu$ and
    $
        Y_i = \fopt(X_i) + \xi_i
    $
    where $\fopt\in\Lmu$
    is $\G$-invariant with
    $\E[\xi_i \mid{} X_1, \dots, X_n] = 0$ and 
    $\E[\xi_i\xi_j\mid{} X_1, \dots, X_n]=\sigma^2 \delta_{ij} < \infty$ 
    for $i,j=1, \dots, n$.
    Similarly, let $X\sim\mu$ and $Y = \fopt(X) +\xi$ where $\E[\xi\mid{}X] = 0$
    and $\E[\xi^2] = \sigma^2 < \infty$.
    Let $f$ be the solution to the KRR problem \cref{eq:krr} 
    given in \cref{eq:krr-solution} and let
    $\bar{f} = \O\iota f$ be its averaged version.
    Let $f'\in\Lmu$ be any predictor with risk not larger than $\bar{f}$,
    i.e., $R[f'] \le R[\bar{f}]$. Then 
    \[
        \E\left[R[f] - R[f']\right]
        \ge 
         \E[\norm{(\id - \O) \iota \bias \fopt}_\mu^2]
            + \sigma^2
        \frac{ 
        \norm{k^\sperp}^2_\Lmumu
        }{(\sqrt{n}M_k + \rho/\sqrt{n})^2 }
    \]
    where $\bias \fopt$ solves
    the corresponding noiseless problem
    \begin{equation}\label{eq:noiseless-krr}
        \argmin_{f\in\H}\; \sum_{i=1}^n(f(X_i) - \fopt(X_i))^2
        + \rho\norm{f}_\H^2.
    \end{equation}
\end{theorem}

\begin{proof}
    It is sufficient to set $f' = \bar{f}$, since if $R[f'] \le R[\bar{f}]$ we have
    \[
        R[f] - R[f'] = R[f] - R[\bar{f}] + R[\bar{f}] - R[f'] \ge R[f] - R[\bar{f}].
    \]
    Let $\pJ$ be the Gram matrix with components 
    $\pJ_{ij} = \inner{k^\sperp_{X_i}}{k^\sperp_{X_j}}_\mu$ and
    let $u \in \R^n$ have components $u_i = \fopt(X_i)$.
    We can use \cref{lemma:generalisation} to get 
    \[
        R[f] - R[\bar{f}]
        = \E[(k_{\xx}^\sperp(X)^\top(K + \rho I)^{-1}\yy)^2 \mid \xx, \yy]
    \]
    where $X \sim \mu$ and $k_{\xx}^\sperp: \X \to \R^n$ 
    with $k_{\xx}^\sperp(x)_i = k_{X_i}^\sperp(x)$.
    Let $\xxi \in \R^n$ have components $\xxi_i = \xi_i$ then
    one finds
    \begin{align}
        &\E[ R[f] - R[\bar{f}] \mid{} \xx] \nonumber\\
        &= 
        \E[(k_{\xx}^\sperp(X)^\top(K + \rho I)^{-1}u)^2 \mid{} \xx]
        + \E[(k_{\xx}^\sperp(X)^\top(K + \rho I)^{-1}\xxi)^2 \mid{} \xx] \nonumber\\ 
        &= 
        \E[(k_{\xx}^\sperp(X)^\top(K + \rho I)^{-1}u)^2 \mid{} \xx]
         +
        \sigma^2 \tr\left(\pJ(K + \rho I)^{-2}\right)\label{eq:kernel-risk-decomp}
    \end{align}
    which follows from the assumptions on $\xi_1,\dots,\xi_n$ 
    (by looking at components or via the trace trick).
    Applying the representer theorem, we find that for all $x\in\X$
    \[
        \bias \fopt(x) = k_{\xx}(x)^\top(K + \rho I)^{-1} \fopt(\xx)
    \]
    which allows us to recognise the first term in \cref{eq:kernel-risk-decomp}
    as $\E[\norm{(\id - \O)\iota\bias \fopt}_\mu^2]$.
    We now analyse the second term 
    in \cref{eq:kernel-risk-decomp}.
    Note that $J^\sperp$ is positive semi-definite because it is a
    Gram matrix, so we can apply \cref{lemma:trace-psd,lemma:op-norm}
    with the bound on the kernel to get
    \[
        \tr\left(\pJ(K + \rho I)^{-2}\right) 
        \ge \lmin\left( (K+\rho I)^{-2} \right ) \tr(\pJ)
        \ge \frac{\tr(\pJ)}{(M_k n + \rho)^{2}} .
    \]
    Taking expectations and using the fact that $X_1, \dots, X_n$ are 
    \iid~gives
    \[
        \E\left[ \tr\left(\pJ(K + \rho I)^{-2}\right)\right]
        \ge \frac{\E[\pJ_{11}]}{(M_k \sqrt{n} + \rho/\sqrt{n})^{2}}.
    \]
    The following completes the proof
    \[
    \E[\pJ_{11}] 
    = \E[\norm{k_X^\sperp}_\mu^2] 
    = \int_\X k^\sperp(x, y)^2 \dd\mu(x) \dd\mu(y)
    = \norm{k^\sperp}_\Lmumu^2.
    \]
\end{proof}

\begin{remark}
    The orthogonal projection 
    on to the span of representers
    $\set{k_{X_1}, \dots, k_{X_n}}$ can be obtained
    by defining $\Pi_n: \H \to \H$
    by $\Pi_n f = \lim_{\rho \to 0} \bias f$ for all $f \in \H$.
    In particular $\Pi_n f(x) = k_{\xx}(x)^\top K^+ f(\xx)$ 
    where $K^+$ is the Moore-Penrose pseudo-inverse of $K$,
    so 
    $\Pi_n^2 f(x) = k_{\xx}(x)^\top K^+ K K^+ f(\xx) = \Pi_n f(x)$
    and $\inner{\Pi_n f}{h}_\H 
    = \sum_{i, j = 1}^n h(X_i)K^+_{ij}f(X_j) 
    = \inner{f}{\Pi_n h}_\H$.
\end{remark}

\begin{corollary}
    Assume the setting and notation of \cref{thm:kernel-invariance}
    and its proof. If $\lmin(\pJ) \ge c $ almost surely
    then \cref{thm:kernel-invariance} reduces to
    \[
        \E\left[R[f] - R[f']\right]
        \ge 
        \frac{ 
        \norm{\fopt}_\mu^2 c
        + \sigma^2
        \norm{k^\sperp}^2_\Lmumu
        }{(\sqrt{n}M_k + \rho/\sqrt{n})^2 }.
    \]
\end{corollary}
\begin{proof}
    The bias term in \cref{eq:kernel-risk-decomp} can be written as
    \begin{align*}
        \E[(k_{\xx}^\sperp(X)^\top(K + \rho I)^{-1}u)^2 \mid{} \xx]
        &= u^\top (K + \rho I)^{-1}\pJ (K +\rho I)^{-1} u \\
        &\ge \norm{u}_2^2 \lmin\left((K + \rho I)^{-1}\pJ(K + \rho I)^{-1}\right)\\
        &\ge
        \frac{
            c \norm{u}_2^2
        }{(nM_k + \rho)^2}.
    \end{align*}
    Using $\E[\norm{u}_2^2] = n\norm{\fopt}_\mu^2$ gives the statement.
\end{proof}

\begin{remark}
   One can obtain the upper bound
    \[
        \E\left[R[f] - R[\bar{f}]\right]
        \le 
         \E[\norm{(\id - \O) \iota \bias \fopt}_\mu^2]
            + \sigma^2 \rho^{-2} \norm{k^\sperp}^2_\Lmumu
    \]
    by applying the upper bound in \cref{lemma:trace-psd}
    to \cref{eq:kernel-risk-decomp}. 
    %The bias term is exact
    %and the $O(1)$ variance term is tight by considering
    %the Dirac kernel where $k(x, y) = 1$ if $x=y$ and $0$ otherwise.
\end{remark}

\subsection{Discussion of \cref{thm:kernel-invariance}}
The first term in \cref{thm:kernel-invariance} 
corresponds to an approximation error or bias 
and the second is the variance term.
We examine these terms in the following sections.
At a high level, the first term 
represents the propensity of the learning
algorithm to produce a predictor with a large $\G$-anti-symmetric component
and can be interpreted as the capacity of the learning algorithm
to represent invariant functions in $\Lmu$.
The numerator in the second term can be viewed as the
complexity of the $\G$-anti-symmetric projection of $\H$, just as 
in \cref{thm:invariant-regression,thm:equivariant-regression}.

We note that as a byproduct of \cref{thm:kernel-invariance} we get a 
lower bound for KRR with random design. 
The variance term matches the 
$O(1/n)$ dependence on the number of examples
in the upper bound from \textcite[Section 3]{mourtada2022elementary}.

\subsubsection{Bias Term}\label{sec:kernel-bias-term}
The quantity $\norm{(\id - \O) \iota\bias \fopt}_\mu^2$
measures the extent to which $\bias \fopt$ is $\G$-invariant
and vanishes if and only if it is.
By invariance $\fopt = \O \fopt$ so we can write
\[
    (\id - \O) \iota\bias \fopt = [\O, \iota\bias] \fopt
\]
where $[A, B] = AB - BA$ denotes the commutator of operators $A$ and $B$.
The bias term
$\E[\norm{[\O, \iota\bias] \fopt}_\mu^2]$ measures
the extent to which, on average, solving \cref{eq:noiseless-krr}
maps invariant functions in $\Lmu$ to invariant functions in $\H$.
Note that $\Lambda_{n, n\rho}\fopt$ solves
\[
    \argmin_{f\in\H}\; \frac1n \sum_{i=1}^n(f(X_i) - \fopt(X_i))^2
    + \rho\norm{f}_\H^2.
\]
We see in \cref{thm:bias-concentration} that, under some mild additional
conditions on $\X$, $\Y$ and $\rho$, 
$
    \iota\Lambda_{n, n\rho}\fopt \to \iota\popbias \fopt
    $
in probability as $n\to\infty$ where for $\alpha > 0$
$\Lambda_\alpha\fopt$ solves 
\begin{equation}\label{eq:population-krr}
    \argmin_{f\in\H}\; \norm{ \iota f - \fopt}_\mu^2 + \alpha\norm{f}_\H^2.
\end{equation}
In this limit
the bias term in \cref{thm:kernel-invariance} becomes
\[
    \E[\norm{(\id - \O) \iota \bias \fopt}_\mu^2]
    \rightarrow
    \norm{[\O, \iota\popbias]\fopt}_\mu^2
\]
as $n\to\infty$.
Hence, under the conditions of \cref{thm:bias-concentration},
bias term vanishes in general as $n \to \infty$ 
if and only if
\[
    \lim_{n \to \infty}[\O,\iota \popbias]\fopt = 0
\]
for all invariant $\fopt \in \Lmu$,
where we have used the continuity of $\O$ from \cref{prop:q-well-defined}
(recall that $\rho$ can depend on $n$).
We interpret this as a fundamental requirement 
for learning invariant functions with kernel ridge regression:
the risk is strictly positive in the limit $n\to\infty$ 
if it is not satisfied.

The above motivates us to consider the operator 
$[\O,\iota \popbias]$ for arbitrary $\rho>0$.
In \cref{lemma:commutators-vanish,prop:commutator-to-kernel}, 
we translate this 
into a condition on the kernel,
finding that $[\O,\iota \popbias] =0$
is equivalent to the kernel satisfying
\[
    \int_\G k(gx, t)\dd\lambda(g) = \int_\G k(x, g^{-1}t)\dd\lambda(g)
\]
for all $x, t \in\X$.
In respect of the above asymptotic analysis,
this is a sufficient condition on $k$
such that KRR can approximate invariant targets
to arbitrary accuracy.%
\footnote{%
    What's missing in saying that it's also a necessary condition
    is that the asymptotic argument
    concludes that $[\O, \iota\popbias]$ should restrict to 0 on invariant functions,
    while the condition on the kernel is equivalent to the commutator vanishing
    altogether.%
    }
Note that the above does not imply that the kernel is invariant.
We will return to this condition in \cref{sec:structure} where we see that it
implies a version of \cref{lemma:l2-decomposition} for $\H$.
We also outline some examples of kernels that satisfy this condition.

Before \cref{prop:commutator-to-kernel} we need the following
result, from which the takeaway is that $[\O , \iota \popbias] = 0$
if and only if $[\O, T_k] = 0$.

\begin{lemma}\label{lemma:commutators-vanish}
    Let $\rho > 0$ and $f\in\Lmu$, then
    \[
        [\O, \iota\popbias]f = 0 \quad\text{ if and only if }\quad
        [\O, T_k](T_k + \rho \id)^{-1}f = 0
    \]
    which immediately implies
    that following are equivalent
    \begin{enumerate}[(a)]
        \item $\exists \rho > 0$ $[\O, \iota\popbias] = 0$
        \item $\forall \rho > 0$ $[\O, \iota\popbias] = 0$
        \item $[\O, T_k] = 0$.
    \end{enumerate}
\end{lemma}

\begin{proof}
    By \cref{lemma:tikhonov-solution}
    \[
        \iota \popbias = (T_k + \rho \id)^{-1} T_k.
    \]
    Also note that for linear operators $A$ and $B$ with $B$ invertible,
    \[
        B[A, B^{-1}]B = B(AB^{-1} - B^{-1}A)B = BA - AB = -[A, B].
    \]
    Then
    \begin{align*}
        \O \iota \popbias
        &= \O(T_k + \rho \id)^{-1} T_k \\ 
        &= [\O, (T_k + \rho \id)^{-1}] T_k 
        +  (T_k + \rho \id)^{-1}\O T_k \\ 
        &= -(T_k + \rho \id)^{-1}[\O, T_k](T_k + \rho \id)^{-1}  T_k 
        +  (T_k + \rho \id)^{-1}\O T_k \\ 
        &= -(T_k + \rho \id)^{-1}[\O, T_k]\iota \popbias
        +  (T_k + \rho \id)^{-1}\O T_k \\ 
        &= -(T_k + \rho \id)^{-1}[\O, T_k]\iota \popbias
        +  (T_k + \rho \id)^{-1}[\O, T_k] 
        +  \iota \popbias \O 
    \end{align*}
    so for any particular $\rho > 0$ and any $f\in\Lmu$,
    $[\O, \iota \popbias]f = 0$ is equivalent to
    \[
        [\O, T_k]\iota \popbias f
        =  [\O, T_k] f.
    \]
    Without loss of generality we can write $f = (T_k + \rho \id)h$ 
    for some $h \in \Lmu$, then the above is
    equivalent to
    \[
        [\O, T_k] (T_k+ \rho \id)^{-1} T_k (T_k + \rho \id) h
        = [\O, T_k](T_k + \rho \id)h
    \]
    which in turn is equivalent to
    \[
        \rho [\O, T_k]h = 0.
    \]
    The proof is complete.
\end{proof}

\begin{theorem}\label{prop:commutator-to-kernel}
    The following are equivalent
    \begin{enumerate}[(a)]
        \item $[\O, T_k] = 0$
        \item  $k$ satisfies
        \[
            \int_\G k(gx, t)\dd\lambda(g) = \int_\G k(x, g^{-1}t)\dd\lambda(g)
        \]
            for all $x, t\in\X$
        \item $\overline{k}$ is a kernel function 
            (it is symmetric and positive definite).
    \end{enumerate}
\end{theorem}
\begin{proof}
    The condition $[\O, T_k] = 0$ means 
    $\O T_k f = T_k \O f$ for all $f\in\Lmu$,
    which is captured by the following
    \begin{align*}
        \int_\G \int_\X k(gx, t)f(t) \dd\mu(t) \dd\lambda(g)
        &\eqas \int_\X k(x, t) \int_\G f(gt)\dd\lambda(g)\dd\mu(t)\\
        &= \int_\G \int_\X k(x, t) f(gt)\dd\mu(t)\dd\lambda(g)\\
        &= \int_\G \int_\X k(x, g^{-1}t) f(t)\dd\mu(t)\dd\lambda(g)
    \end{align*}
    where the first line is the condition for the commutators to vanish, 
    the second line uses Fubini's theorem and 
    the third line uses the invariance of $\mu$.
    So, applying Fubini's theorem again, 
    $[\O, T_k] = 0$ 
    is equivalent to
    \[
        \int_\X \left(\int_\G k(gx, t)\dd\lambda(g) 
        - \int_\G k(x, g^{-1}t)\dd\lambda(g)\right)
        f(t)\dd\mu(t) \eqas 0
    \]
    for all $f \in \Lmu$.
    In turn this is equivalent to 
    \[
        \int_\G k(gx, t)\dd\lambda(g) = \int_\G k(x, g^{-1}t)\dd\lambda(g)
    \]
    $(\mu\otimes\mu)$-almost-everywhere and indeed pointwise 
    by the injectivity of $\iota$.
    Moreover, one can check that $\overline{k}$ is a kernel function if and only if
    the above
    display holds, it is positive definite since for all $n\in\N$, $a\in \R^n$
    and $x_1, \dots, x_n \in \X$
    \[
        \sum_{i,j=1}^na_ia_j\overline{k}(x_i, x_j) 
        = \int_\G \sum_{i,j=1}^na_ia_jk(x_i, gx_j)\dd\lambda(g) \ge 0
    \]
    and symmetry is equivalent to the starred equality holding for all $x, x'\in \X$
    \[
        \overline{k}(x, x') 
        = \int_\G k(x, gx') \dd\lambda(g)
        \overset{\star}{=} \int_\G k(gx, x') \dd\lambda(g)
        = \int_\G k(x', gx) \dd\lambda(g)
        = \overline{k}(x', x).
    \]
\end{proof}

The following result leans very heavily on \parencite{devito2005learning}.
\begin{theorem}
    \label{thm:bias-concentration}
    Let $\X \subset \R^d$ be compact and $\Y \subset [-B, B]$.
    Let $\fopt \in \Lmu$.
    Let $\delta \in (0, 1)$, then with probability at least $1 - \delta$
    \[
        \norm{\iota\Lambda_{n, n\rho}\fopt - \iota \popbias\fopt}_\mu
        \le \frac{B M_k^2}{8\rho}
            q\left(\frac{8}{n} \log(4/\delta)\right)
        +
        \frac{B M_k}{4\sqrt{\rho}} q\left(\frac{8}{n} \log(4/\delta)\right).
    \]
    where $q(t) = \frac12(t + \sqrt{t^2 + 4t})$ and
    $q(t) = \sqrt{t} + o(t)$ as $t \to 0$.
    In particular if $\rho = \omega(n^{-l})$ for $l < 1/2$ then
    \[
        \iota\Lambda_{n, n\rho}\fopt \to \iota \popbias\fopt
    \]
    in probability as $n\to\infty$.
\end{theorem}

\begin{proof}
    The proof is pieced together from \parencite{devito2005learning}.
    Endow $\R^n$ with the inner product 
    $\inner{u}{v}_n = \frac1n u^\top v$ and write 
    the induced norm as $\norm{\cdot}_n$.
    Define the data dependent evaluation operator
    $\eval: \H \to \R^n$ by $(\eval f) = f(\xx)$.
    Its adjoint is $\eval^*: \R^n \to \H$ with values
    $\eval^* v (x) = \inner{k_{\xx}(x)}{v}_n$.
    The noiseless KRR problem in \cref{thm:kernel-invariance}
    solved by $\Lambda_{n, n\rho}\fopt$ can then be written
    \[
        \argmin_{f\in\H}\; \norm{\eval f - u}_n^2 + \rho \norm{f}_\H^2
    \]
    where $u=\fopt(\xx)$
    and then \cref{lemma:tikhonov-solution} shows that the unique solution is
    \[
        \fhat = (\eval^*\eval + \rho \id)^{-1}\eval^* u
    \]
    so $\fhat = \bias \fopt$. Analogously, the unique
    solution to \cref{eq:population-krr} is
    \[
        f = (S_k\iota +\rho\id)^{-1}S_k\fopt
    \]
    and $f= \popbias \fopt$. Then, following \parencite[Eq.~18]{devito2005learning},
    \begin{align}
       \fhat - f
        &=(\eval^*\eval + \rho \id)^{-1} \eval^* u
            -
            (S_k \iota + \rho\id)^{-1}S_k\fopt\nonumber\\
        &= \left[
        (\eval^*\eval + \rho \id)^{-1}
            -
            (S_k \iota + \rho\id)^{-1}
        \right]
        \eval^* u
        + (S_k\iota +\rho \id)^{-1}(\eval^* u - S_k\fopt)\nonumber\\
        &= (S_k \iota + \rho\id)^{-1}
        (S_k \iota - \eval^*\eval)
        (\eval^*\eval + \rho \id)^{-1}
        \eval^* u\\
        &\phantom{=}
        + (S_k\iota +\rho \id)^{-1}(\eval^* u - S_k\fopt).
        \label{eq:approximation-residual}
    \end{align}
    The derivations in \parencite[Eq.~19 and Eq.~20]{devito2005learning}
    (when corrected slightly) give
    \begin{align*}
        \opnorm{\iota(S_k\iota+ \rho \id)^{-1}} &\le \frac{1}{2\sqrt{\rho}}\\
        \opnorm{(\eval^* \eval+ \rho \id)^{-1}\eval^*} &\le \frac{1}{2\sqrt{\rho}}.
    \end{align*}
    Then with $\norm{u}_n\le B$, \cref{eq:approximation-residual} becomes
    \[
        \norm{\iota\fhat - \iota f}_\mu
        \le \frac{B}{4\rho}
        \opnorm{S_k \iota - \eval^*\eval}
        + \frac{1}{2\sqrt{\rho}}\norm{\eval^*u - S_k \fopt}_\H.
    \]
    We apply \cref{thm:op-norm-concentration} to get that
    with probability at least $1 - \delta$
    \[
        \norm{\iota\fhat - \iota f}_\mu
        \le \frac{B M_k^2}{8\rho}
            q\left(\frac{8}{n} \log(4/\delta)\right)
        +
        \frac{B M_k}{4\sqrt{\rho}} q\left(\frac{8}{n} \log(4/\delta)\right).
    \]
    The proof is complete.
\end{proof}

\begin{theorem}[{\parencite[Theorem 3]{devito2005learning}}]%
    \label{thm:op-norm-concentration}
    Assume the setting and notation of \cref{thm:bias-concentration}
    and its proof.
    With probability at least $1 - \delta$ we have, simultaneously,
    \begin{align*}
        \opnorm{S_k \iota - \eval^* \eval} 
        &\le \frac{M_k^2}{2} q\left(\frac{8}{n} \log(4/\delta)\right)\\
        \norm{\eval^*u - S_k \fopt}_\H
        &\le \frac{B M_k}{2} q\left(\frac{8}{n} \log(4/\delta)\right)
    \end{align*}
    where $q(t) = \frac12(t + \sqrt{t^2 + 4t})$ and
    $q(t) = \sqrt{t} + o(t)$ as $t \to 0$.
\end{theorem}

\subsubsection{Variance Term}
We write $N[k^\sperp] = \norm{k^\sperp}_\Lmumu^2$
and make the same definition for $k$ and $\overline{k}$.
\Cref{thm:kernel-invariance} shows that 
the solution $f$ to \cref{eq:krr} can only learn to be invariant
if the number of examples dominates $N[k^\sperp]$.
Intuitively, we think of $N[k^\sperp]/M_k^2$ as an effective dimension of the space of
$\G$-anti-symmetric functions in $\H$.
With this interpretation, the variance term in \cref{thm:kernel-invariance}
plays the same role as the $\dim A$ term in \cref{thm:invariant-regression}.
Taking the ridgeless limit $\rho \to 0$, 
we view \cref{thm:kernel-invariance}
as a generalisation of \cref{thm:invariant-regression}.

This interpretation of $N[k]$ as a measure of dimension or capacity
will appear again for the linear kernel in \cref{sec:linear-kernel}.
In \cref{sec:structure} we associate $\G$-symmetric and $\G$-anti-symmetric
RKHSs with $\overline{k}$ and $k^\sperp$ respectively, which somewhat solidifies 
this perspective.
In particular, the decomposition in \cref{lemma:N-decomposition} below
will correspond to a decomposition of RKHSs $\H = \O \H \oplus (\id - \O)\H$,
the same as \cref{lemma:l2-decomposition} but with the RKHS inner product.
However, the interpretation of $N[k]$ as a dimension
requires some refinement: $N[k]$ depends on the scale of $k$ while
any capacity measure of $\H$ should not, 
and the scale invariant $N[k] / M_k$ does not give
the decomposition in \cref{lemma:N-decomposition}.

\begin{lemma}\label{lemma:N-decomposition}
    \[
        N[k] = N[\overline{k}] + N[k^\sperp]
    \]
\end{lemma}
\begin{proof}
    \begin{align*}
        N[k] 
        &= \norm{k}_\Lmumu^2\\
        &= \int_\X k(x, y)^2 \dd\mu(x)\dd\mu(y)   \\
        &= \int_\X \norm{k_x}_\mu^2 \dd\mu(x) \\
        &= \int_\X \norm{\kbar{x}}_\mu^2 \dd\mu(x) 
        +
        \int_\X \norm{k_x^\sperp}_\mu^2 \dd\mu(x) \\
        &= N[\overline{k}] + N[k^\sperp]
    \end{align*}
\end{proof}

\subsection{Special Case: The Linear Kernel}\label{sec:linear-kernel}
In this section we explore the relation of \cref{thm:kernel-invariance}
to \cref{thm:invariant-regression} by specialising to the linear kernel.
We begin with a calculation $N[k]$ for the linear kernel, which
agrees with the interpretation as a dimension from the previous section.
For the rest of this section we refer to the action $\phi$ of $\G$ explicitly,
writing $\phi(g)x$ instead of $gx$.

\begin{example}[Linear Kernel]\label{example:linear-kernel-dim}
    Let $X$ and $Y$ be mean zero isotropically distributed random 
    vectors $\R^d$ whose co-ordinates have unit variance 
    and let $k$ be the linear kernel on $\R^d$ with
    $
        k(x, y) = x^\top y
        $,
    then
    \[
        N[k] = \E[k(X, Y)^2] = \E[X_iX_jY_iY_j] = d.
    \]
    Recall the matrix $\proj = \int_\G \phi(g) \dd\lambda(g)$ defined 
    in \cref{sec:invariant-regression} along with the $\G$-symmetric and $\G$-anti-symmetric
    subspaces $S$ and $A$ of $\R^d$ respectively.
    We have
    \[
        \overline{k}(x, y) 
        = \int_\G x^\top \phi(g) y \dd\lambda(g)
        = x^\top \proj y
    \]
    so
    \[
        N[\overline{k}] 
        = \E\left[(X^\top \proj Y)^2\right] = \fnorm{\proj}^2 = \dim S
    \]
    and $N[k^\sperp] = d - \dim S = \dim A$ by \cref{lemma:N-decomposition}
\end{example}

\begin{theorem}\label{prop:kernel-linear}
    Assume the setting and notation of \cref{thm:kernel-invariance} with $\sigma^2=1$.
    In addition, let $\X = \mathbb{S}_{d-1}(\sqrt{d})$ be the $(d-1)$-sphere of radius
    $\sqrt{d}$ with $d>1$ and let $\mu = \unif\X$.
    Let $\G$ act via an orthogonal representation $\phi$ on $\X$ and define
    the matrix 
    $
        \proj = \int_\G \phi(g) \dd\lambda(g)
        $.
    Let $k(x, y) = x^\top y$ be the linear kernel and suppose $\fopt(x) = \theta^\top x$
    for some $\theta \in \R^d$.
    Let $K$ be the kernel Gram matrix
    $K_{ij} = k(X_i, X_j)$ and
    let $\gamma_1, \dots, \gamma_n$ be its eigenvalues.
    Define
    \[
        \zeta_1(\rho) = \E\left[\sum_{i=1}^n
        \frac{\gamma_i^2}{(\gamma_i + \rho)^2}\right]
        \quad\text{ and }\quad
        \zeta_2(\rho) = \E\left[\left(
        \sum_{i=1}^n \frac{\gamma_i}{\gamma_i + \rho}\right)^2\right].
    \]
    Then the bound in \cref{thm:kernel-invariance} becomes
    \[
        \E[R[f] - R[f']]
        \ge 
        \frac{
            \norm{\theta}_2^2 
            (d \zeta_1(\rho) - \zeta_2(\rho))
            (d - \fnorm{\proj}^2)
        }{d(d+2)(d-1)}
        +
        \frac{ 
        d - \fnorm{\proj}^2
        }{(d\sqrt{n} + \rho/\sqrt{n})^2 } 
    \]
    where the first and second terms are exactly the bias and variance
    terms in \cref{thm:kernel-invariance} respectively and 
    each term is non-negative.
\end{theorem}

\begin{proof}
    We will calculate each term in the bound in \cref{thm:kernel-invariance}
    separately.
    We make use of the Einstein summation convention, implicitly
    summing over repeated indices.
    The variance term is straightforward.
    We know from \cref{example:linear-kernel-dim} that%
    \footnote{%
        As a sanity check, direct calculation gives
        \[
            \fnorm{\proj}^2
            = \tr(\proj^\top \proj)
            = \int_\G \tr(\phi(g_1)^\top \phi(g_2))\dd\lambda(g_1)\dd\lambda(g_2)
            \le \left( \int_\G\fnorm{\phi(g)} \dd\lambda(g)\right)^2
            = d
        \]
        using Cauchy-Schwarz on the Frobenius inner product and with
        $\fnorm{\phi(g)} = \sqrt{d}$ because the representation is orthogonal.
        }%
    \[
        0 \le \norm{k^\sperp}_\Lmumu^2 = d - \fnorm{\proj}^2
    \]
    and the denominator comes from $M_k = \sup_{x} k(x, x) = \norm{x}_2^2= d$.
    The rest of the work is on the bias term. 
    Let $\xx\in\R^{n\times d}$ have components
    $\xx_{ij} = (X_i)_j$. Then 
    \[
        K_{ij} \coloneqq k(X_i, X_j) = X_i^\top X_j 
        = (X_i)_l(X_j)_l
        = (\xx\xx^\top)_{ij}
    \]
    and 
    \[
        k_{\xx}(y)_i = k_{X_i}(y) = X_i^\top y 
        = \xx_{ij}y_j
        = (\xx y)_i
    \]
    along with $\fopt(\xx)_i = \fopt(X_i) = X_i^\top\theta =(\xx \theta)_i$.
    So
    \begin{align*}
        \bias \fopt(y) 
        &= k_{\xx}(y)^\top(K +\rho I_n)^{-1} \fopt(\xx) \\ 
        &= (\xx y)^\top (\xx \xx^\top + \rho I_n)^{-1} \xx \theta \\ 
        &= y^\top\xx^\top (\xx \xx^\top + \rho I_n)^{-1} \xx \theta 
    \end{align*}
    where for $m\in\N$ we write $I_m$ for the $m\times m$ identity matrix.
    With $\proj^\sperp = I_d - \proj$ we get (by linearity)
    \begin{align*}
        (\id - \O)\bias \fopt(y) 
        &= y^\top\proj^\sperp \xx^\top (\xx \xx^\top + \rho I_n)^{-1} \xx \theta\\
        &= y^\top\proj^\sperp (\xx^\top \xx + \rho I_d)^{-1} \xx^\top\xx \theta
    \end{align*}
    which follows from
    \begin{align*}
        \xx^\top (\xx \xx^\top + \rho I_n)^{-1} \xx
        &=
        (\xx^\top\xx + \rho I_d)^{-1}
        (\xx^\top\xx + \rho I_d)
        \xx^\top (\xx \xx^\top + \rho I_n)^{-1} \xx \\
        &= 
        (\xx^\top\xx + \rho I_d)^{-1}
        \xx^\top( \xx\xx^\top + I_n) ( \xx\xx^\top + I_n)^{-1} \xx\\
        &= 
        (\xx^\top\xx + \rho I_d)^{-1}
        \xx^\top\xx.
    \end{align*}
    Write $P = (\xx^\top\xx + \rho I_d)^{-1} \xx^\top\xx$, so
    \[
        \norm{(\id - \O)\bias \fopt(y) }_\mu^2
        = \norm{\proj^\sperp P \theta}_2^2
    \]
    and
    \[
        \E[\norm{(\id - \O)\bias \fopt(y) }_\mu^2]
        = \theta_i \theta_j \proj^\sperp_{lq}\proj^\sperp_{mq}\E[ P_{il}P_{jm}]
    \]
    We show that the 4-tensor $\E[ P_{ab}P_{ce}]$ is isotropic.
    For any rotation matrix $R\in\SO_d$
    \begin{align*}
        R_{\alpha a}
        R_{\beta b}
        R_{\gamma c}
        R_{\epsilon e}
        \E[ P_{ab}P_{ce}]
        = \E[(RPR^\top)_{\alpha\beta} (RPR^\top)_{\gamma\epsilon}]
    \end{align*}
    then isotropy of the covariates implies 
    $\xx^\top \xx \eqdist R\xx^\top \xx R^\top $ for all $R\in \SO_d$
    and hence
    \[
        P \eqdist (R \xx^\top \xx R^\top+ \rho I)^{-1} R\xx^\top \xx R^\top
        = RPR^\top.
    \]
    The general form of a real, isotropic 4-tensor is
    \[
        \E[P_{ab}P_{ce}]
        = \alpha \delta_{ab} \delta_{ce} 
        + \beta \delta_{ac}\delta_{be} 
        + \gamma \delta_{ae} \delta_{cb}
    \]
    for $\alpha, \beta, \gamma \in \R$ \parencite{hodge1961}.
    We can ignore $\alpha$ as will be clear in a moment
    and $P^\top = P$ so $\beta = \gamma$.
    We have
    \begin{align*}
        \zeta_1(\rho) &\coloneqq \E[\tr(P^2)] = d \alpha + d(d+1)\beta \\
        \zeta_2(\rho) &\coloneqq \E[\tr(P)^2] = d^2\alpha  + 2d \beta 
    \end{align*}
    so 
    \[
        \beta = \frac{d \zeta_1(\rho) - \zeta_2(\rho)}{d(d+2)(d-1)}
    \]
    and the bias term is
    \begin{align*}
        \E[\norm{(\id - \O)\bias \fopt(y) }_\mu^2]
        & = \theta_a \theta_c \proj^\sperp_{bq}\proj^\sperp_{eq}\E[ P_{ab}P_{ce}]\\
        & = 
        (\alpha + \beta) \norm{\proj^\sperp \theta}_2^2
        + \beta \norm{\theta}_2^2 \fnorm{\proj^\sperp}^2\\
        & = 
        \frac{
            \norm{\theta}_2^2 \fnorm{\proj^\sperp}^2
        (d \zeta_1(\rho) - \zeta_2(\rho))}{d(d+2)(d-1)}
    \end{align*}
    where in the final line we used $\proj^\sperp \theta = 0$
    by assumption that $\fopt$ is invariant.
    Note that $\fnorm{\proj^\sperp}^2 = d - \fnorm{\proj}^2$.
    Finally, it's immediate by diagonalising that
    \[
        \zeta_1(\rho) = \E\left[\sum_{i=1}^n
        \frac{\gamma_i^2}{(\gamma_i + \rho)^2}\right]
    \]
    and
    \[
        \zeta_2(\rho) = \E\left[\left(
        \sum_{i=1}^n \frac{\gamma_i}{\gamma_i + \rho}\right)^2\right].
    \]
    The inequality $\zeta_2(\rho) \le \min\set{n,d}\zeta_1(\rho)$ follows from
    $\norm{v}_1^2 \le m \norm{v}_2^2$ for all $v\in\R^m$ and the
    number of non-zero eigenvalues of $K = \xx \xx^\top$.
\end{proof}

\Cref{prop:kernel-linear} is a special case of \cref{thm:kernel-invariance}
with the linear kernel.
From \cref{example:linear-kernel-dim} we have $d - \fnorm{\proj}^2 = \dim A$,
highlighting a similarity between \cref{prop:kernel-linear,thm:invariant-regression}.
These results are similar but not quite the same: the input spaces 
and distributions are different, while \cref{prop:kernel-linear}
has a ridge penalty but \cref{thm:invariant-regression} does not.
Furthermore, the linear kernel is unbounded on $\R^d$, so a direct application 
of \cref{thm:kernel-invariance} to the setting of \cref{thm:invariant-regression} is
not possible.

On the other hand, in the ridgeless limit 
$\rho \to 0$ \cref{prop:kernel-linear} becomes
\[
    \E[R[f] - R[f']]
    \ge 
    \frac{
        \norm{\theta}_2^2 
        (d \min\set{n, d} - \min\set{n, d}^2)
        \dim A
    }{d(d+2)(d-1)}
    +
    \frac{ 
    \dim A
    }{(d\sqrt{n} + \rho/\sqrt{n})^2 }.
\]
So the bias term turns out to match \cref{thm:invariant-regression}
exactly: when $n \ge d$ it vanishes and 
when $n < d$ we recover the $n<d-1$ case in \cref{thm:invariant-regression}.
This is natural, as the bias term was calculated exactly
in each case.
The variance term in \cref{thm:kernel-invariance}
is estimated which means there is no opportunity for double
descent behaviour to appear in \cref{prop:kernel-linear}.
One would hope that as $d\to\infty$
the two results look qualitatively similar
because in this limit the normally distributed covariates
in \cref{thm:invariant-regression}
become uniform on the sphere as in \cref{prop:kernel-linear}.
Indeed if $d\to\infty$ and $n$ remains fixed then the results match,
but if $n, d\to\infty$ with $r = n/d$ finite the variance term
above
looks like $\frac1r \frac{\dim A }{d^3}$ while the variance term
in \cref{thm:invariant-regression}, which gives an equality,
goes like $\frac{r}{1-r}\frac{\dim A}{d}$ when $r < \frac{d-1}{d}$ and
$\frac{1}{r-1}\frac{\dim A}{ d}$ when $r > \frac{d+1}{d}$.
This discrepancy is due to the $M_k$ in the denominator.
It's possible that it can be improved with a more sophisticated approach in the 
proof of \cref{thm:kernel-invariance}.

\section{Structure of $\H$}\label{sec:structure}
In this section we present \cref{thm:rkhs-decomposition},
which is a version of \cref{lemma:l2-decomposition} for RKHSs.
If $k$ satisfies, for all $x, x' \in \X$,
\begin{equation}\label{eq:kernel-switch}
    \int_\G k(gx, x') \dd \lambda (g) = \int_\G k(x, gx') \dd \lambda (g).
\end{equation}
then $\H$ is an orthogonal direct sum of two subspaces, each of which is an RKHS,
one of invariant functions and another of those that vanish when averaged
over $\G$.
Moreover, the kernels turn out to be $\overline{k}$ and $k^\sperp$ respectively.

For \cref{eq:kernel-switch} to hold,
it is sufficient to have $k(gx, y) = k(x, g^{-1}y)$ and
use the change of variables $g^{-1} \mapsto g$.
Any kernel for which this stronger condition holds is known as a
\emph{unitary kernel}.
Highlighting two special cases:
any inner product kernel $k(x, x') = \kappa(\inner{x}{x'})$
where the action of $\G$ is unitary with respect to $\inner{\cdot}{\cdot}$
satisfies \cref{eq:kernel-switch}, as does any stationary kernel
$k(x, x') = \kappa(\norm{x - x'})$ with a norm that is \emph{preserved} by
$\G$ in the sense that $\norm{gx - gx'} = \norm{x - x'}$ for all
$g\in\G$ and all $x, x' \in \X$.
Respectively, examples are the Euclidean inner product with orthogonal representations
and permutations with any $p$-norm.

As it happens, if the kernel satisfies \cref{eq:kernel-switch} then
$\overline{k}$ qualifies as a Haar integration kernel,
introduced by \textcite{haasdonk05invariancein} and defined by
\[
    \tilde{k}(x, x') = \int_\G k(gx, g'x') \dd\lambda(g) \dd\lambda(g')
\]
for any kernel $k$. Simply apply \cref{eq:kernel-switch} and use the invariance
of $\lambda$ to get $\overline{k}$.
Note that this means that \cref{eq:kernel-switch} implies that
$\overline{k}$ is $\G$-invariant in both arguments.

Before we go further, for all $h\in\H$ we define the function
$\O h$ by
\[
    \O h(x) = \int_\G h(gx) \dd\lambda(g).
\]
Implicit in \cref{thm:rkhs-decomposition} is that
$\O: \H \to \H$ is well defined.
Note the slight abuse of notation,
simultaneously writing $\O$ for the operator on different spaces.
The map $\O : \H \to \H$ is not well-defined in general.
Indeed there are trivial examples where it is not, 
such as \cref{example:o-not-well-defined}.

\begin{example}\label{example:o-not-well-defined}
    Define $k: \R^2 \times \R^2 \to \R$ by $k(x, y) = x^\top Ay$ with
    $A = \begin{psmallmatrix}
        1 & 0 \\
       0 & 0 
    \end{psmallmatrix}$.
    It's easy to check that $k$ is symmetric and positive definite,
    so there is a (unique) RKHS $\H$ with
    reproducing kernel $k$ by the Moore-Aronszajn theorem.
    The RKHS consists of linear functions of the following form
    and the completion thereof
    \[
        f(x) =  \sum_{i=1}^n \alpha_i k(z_i, x) 
        = \left(\sum_{i=1}^n \alpha_i  Az_i\right)^\top x
    \]
    for $\alpha_1, \dots, \alpha_n \in \R$, $z_1, \dots, z_n \in \R^2$ and
    some $n\in\N$.
    Any $f \in \H$ has $f(e_2) = 0$ where $e_2 = (0, 1)^\top$.
    Now consider $\G = \sym_2$ acting on $\R^2$ by permutation of the co-ordinates.
    Then $\O f (x) = \frac12 f(x) + \frac12 f(Bx)$ where 
    $B = \begin{psmallmatrix}
        0 & 1\\
        1 & 0
    \end{psmallmatrix}$.
    If $f(x) = e_1^\top Ax$ where $e_1 = (1, 0)^\top$,
    then $\O f(e_2) = \frac12$ so $\O f \not\in \H$.
    It's easy to check that $k$ does not satisfy \cref{eq:kernel-switch}.
\end{example}

\begin{theorem}\label{thm:rkhs-decomposition}
   Suppose the kernel satisfies \cref{eq:kernel-switch}
   and let $\Hs  = \{f \in \H : f \text{ is $\G$-invariant}\}$ and
   $\Ha = \{f \in \H: \O f = 0\}$, then:
    \begin{itemize}
        \item $\H$ admits the orthogonal decomposition $\H = \Hs \oplus \Ha$
        \item $\Hs$ is an RKHS with kernel 
            $
                \overline{k}(x, y) = \int_\G k(x, gy) \dd\lambda(g)
                $
        \item $\Ha$ is an RKHS with kernel 
            $ k^\sperp(x, y) = k(x, y) - \overline{k}(x, y) $
   \end{itemize}
\end{theorem}
\begin{proof}
    First we show that $\O: \H \to \H$ is well defined.
    Let the image of $\Lmu$ under $S_k$ be $\H_2$.
    By \cref{lemma:dense-image} the completion of $\H_2$ in 
    $\norm{\cdot}_\H$ is $\H$. 
    Then for all $f\in\Lmu$
    \begin{align*}
        \O S_k f (x)
        &= \int_\G \int_\X k(gx, t)f(t) \dd\mu(t) \dd\lambda(g)\\
        &= \int_\X\int_\G  k(gx, t)f(t) \dd\lambda(g)\dd\mu(t) \\
        &= \int_\X\int_\G  k(x, gt)f(t) \dd\lambda(g)\dd\mu(t) \\
        &= \int_\G\int_\X  k(x, gt)f(t) \dd\mu(t) \dd\lambda(g)\\
        &= \int_\G\int_\X  k(x, t)f(g^{-1}t) \dd\mu(t) \dd\lambda(g)\\
        &= \int_\X  k(x, t)\int_\G f(g^{-1}t) \dd\lambda(g) \dd\mu(t) \\
        &= \int_\X  k(x, t)\int_\G f(gt) \dd\lambda(g) \dd\mu(t) \\
        &= S_k \O f (x)
    \end{align*}
    where we used \cref{eq:kernel-switch},
    Fubini's theorem, the invariance of $\mu$
    and the compactness of $\G$ to substitute $g^{-1}\mapsto g$.
    This shows that $\O: \H_2 \to \H_2$ is well defined and
    that $\O$ and $S_k$ commute.
    Let $a,b \in \H_2$ with preimages $a', b' \in \Lmu$ such that
    $a = S_ka'$ and $b = S_kb'$, then, making use of the fact that 
    $\O$ is self-adjoint on $\Lmu$ from \cref{prop:self-adjoint} and
    that $[T_k, \O]=0$ by \cref{prop:commutator-to-kernel},
    \begin{align*}
        \inner{\O a}{b}_\H 
        &= \inner{\O S_ka'}{S_kb'}_\H
        = \inner{S_k\O a'}{S_kb'}_\H
        = \inner{\O a'}{T_kb'}_\mu\\
        &= \inner{a'}{\O T_k b'}_\mu
        = \inner{a'}{T_k \O b'}_\mu
        = \inner{S_k a'}{ S_k \O b'}_\H\\
        &= \inner{S_k a'}{ \O S_k b'}_\H
        = \inner{a}{\O b}_\H.
    \end{align*}
    From this it follows that, for all $f\in \H_2$,
    \[
    \norm{\O f}_\H^2 = \inner{\O f}{\O f}_\H 
    = \inner{f}{\O f}_\H \le \norm{f}_\H\norm{\O f}_\H
    \]
    so $\norm{\O f}_\H \le \norm{f}_\H$. 
    We can use this to show that $\O: \H \to \H$ is well-defined.
    Since $\H_2$ is dense in $\H$, for any $f \in \H$
    there is a sequence $\{f_n\} \subset \H_2$ converging to $f$ in $\norm{\cdot}_\H$.
    We need to show that $\O f \in \H$.
    From above
    $
        \norm{\O f_n - \O f_m}_\H \le \norm{f_n - f_m}_\H
        $
    so the sequence $\{\O f_n\}$ is Cauchy in $\H$, so must converge to 
    some $\tilde{f}\in \H$ by completeness.
    Note that convergence in $\norm{\cdot}_\H$ implies uniform convergence
    since for all $h_1,h_2 \in \H$
    \[
    \abs{h_1(x) - h_2(x)} = \abs{\inner{k_x}{h_1 - h_2}_\H}
    \le \sqrt{M_k}\norm{h_1 - h_2}_\H.
    \]
    Then for all $x\in\X$ and all $n\in\N$
    \begin{align*}
        \abs{\O f(x) - \tilde{f}(x)}
        &\le 
        \abs{\O f(x) - \O f_n(x)}
        + \abs{\O f_n(x) - \tilde{f}}\\
        &\le 
        \abs{\O f(x) - \O f_n(x)}
        + \sqrt{M_k}\norm{\O f_n - \tilde{f}}_\H
    \end{align*}
    and the first term can be controlled by
    \[
        \abs{\O f(x) - \O f_n(x)}
        \le \int_\G \abs{f(gx) - f_n(gx)}\dd\lambda(g)
        \le \sup_{t\in\X}\abs{f(t) - f_n(t)}
        \le \sqrt{M_k} \norm{f - f_n}_\H
    \]
    so $\tilde{f} = \O f$ and $\O f \in \H$.
    
    Now we prove the orthogonal decomposition of $\H$.
    We show that $\O$ is self-adjoint on $\H$.
    Let $f, h \in \H$, then there are sequences $\{f_m\}$ and $\{h_n\}$
    in $\H_2$ with limits $f$ and $h$ respectively.
    We just showed that $\O$ commutes with taking limits of these 
    sequences. So, taking $m\to\infty$ first,
    \begin{align*}
        \inner{\O f}{h}_\H 
        &= \inner{\O \lim_{m\to\infty}f_m}{\lim_{n\to\infty}h_n}_\H 
        = \lim_{n\to\infty}\lim_{m\to\infty}\inner{\O f_m}{h_n}_\H \\
        &= \lim_{n\to\infty}\lim_{m\to\infty}\inner{f_m}{\O h_n}_\H 
        = \inner{h}{\O f}_\H.
    \end{align*}
    Now we know that $\O: \H \to \H$ is self-adjoint,
    it's easy to check that it's idempotent so is
    an orthogonal projection on $\H$ (see \cref{sec:proof-l2-decomposition}).
    Let $h_S$ have eigenvalue $1$ and $h_A$ have eigenvalue $0$ under $\O$,
    then 
    $
        \inner{h_S}{h_A}_\H = \inner{\O h_S}{h_A}_\H = \inner{h_S}{\O h_A}_\H = 0
        $.
    Therefore, by linearity, for all $f \in \H$ we can write 
    $
        f = \bar{f} + f^\sperp
        $
    where $\bar{f} = \O f \in \Hs$ is $\G$-invariant and $f^\sperp = f - \O f \in \Ha$
    and these terms are orthogonal.
    
    We conclude by showing that $\Hs$ and $\Ha$ are RKHSs with
    kernels $\overline{k}$ and $k^\sperp$ respectively.
    By the linearity of $\O$, $\Hs = \O \H \subset \H$ is a subspace,
    so also an inner product space with $\inner{\cdot}{\cdot}_\H$.
    We check the completeness of $\Hs$. Above we already showed that
    if $\{f_n\}$ be a sequence in $\Hs$
    which has limit $f\in\H$ by completeness of $\H$, then
    $\lim_{n\to\infty}\O f_n = \O \lim_{n\to\infty}f_n = \O f \in \Hs$,
    which shows that $\Hs$ is complete. The same argument works to show
    that $\Ha$ is complete. The evaluation functional is continuous
    on each of these subspaces so each is an RKHS.
    Now for all $h_S \in \Hs$ we have
    \[
        h_S(x) = \inner{h_S}{k_x}_\H = \inner{h_S}{\O k_x}_\H 
    \]
    and the uniqueness afforded by the  
    Riesz representation theorem for Hilbert 
    spaces \parencite[Theorem 4.12]{rudin1987realandcomplex}
    tells us that the reproducing kernel for $\Hs$
    is 
    \[
        \inner{\O k_x}{\O k_y}_\H
        = \inner{k_x}{\O k_y}_\H
        = \int_\G k(x, gy)\dd\lambda(g)
        = \overline{k}(x, y).
    \]
    From the above it can be seen that $\overline{k}$ is positive definite
    and symmetric. The same steps using $\id - \O$ show that $k^\sperp$
    is the reproducing kernel of $\Ha$ and also verify that it is positive definite
    and symmetric.
\end{proof}

\begin{remark}
    It is straightforward to use the orthogonality in \cref{thm:rkhs-decomposition}
    to derive an invariant version of the representer theorem. That is, that any
    invariant minimiser of the of the appropriate risk functional is an element
    of the linear span of the evaluations of $\overline{k}$ at the training
    data. The same statement but for equivariance was shown by
    \textcite{reisert2007learning}.
\end{remark}

\chapter{General Theory II: Orbit Representatives}\label{chap:general-theory-ii}

\section*{Summary}
In previous chapters we studied invariance and equivariance by
considering the action of certain averaging operators 
on a function space. 
In this chapter we take a different approach, leveraging the observation
that an invariant function can be specified by its values on one representative
from each orbit of $\X$ under $\G$.
We show rigorously how learning 
with invariant or equivariant hypotheses
reduces to learning on a set of orbit representatives.
In addition, we show how to use these equivalences to derive a sample complexity
bound for learning invariant/equivariant classes with
empirical risk minimisation
in terms of the geometry of the input and output
spaces.

\section{Preliminaries}\label{sec:preliminaries}
In this chapter, $\Y$ is not restricted to be $\R^k$
and $(\Y, \S_\Y)$ can be any standard Borel space.
The action $\psi$ is an arbitrary action of $\G$ on $\Y$
and not necessarily a linear representation unless explicitly stated.
We will write $\Z = \X \times \Y$ and
the action of $\G$ on $\Z$ is defined by $g(x, y) = (gx, gy)$.
A set of functions $\F$ is \emph{invariant} or \emph{equivariant}
if all its elements are invariant or equivariant respectively.
We say the action of $\G$ on $\X$
is \emph{free} if the non-trivial elements of $\G$ have no fixed 
points, that is $\forall x\in \X:$ $gx = x \implies g = e$ the identity 
element. If the action is free then $\phi$ is injective
as a function from $\G$ to bijections on $\X$.
Let $P$ and $Q$ be random variables, we use the notation
$A \sim (P, Q)^n$ to mean the tuple $A = ((P_1, Q_1), \dots, (P_n, Q_n))$
where the $(P_i, Q_i)$ are \iid~and distributed 
independently of and identically to $(P, Q)$.

\begin{definition}\label{def:task}
    A \emph{task} is a tuple $\T = (X, Y, \ell)$ where $X$ is a random
    element of $\X$, $Y$ is a random element of $\Y$ and $\ell: \Y \times\Y \to \R_+$
    is an integrable function called the \emph{loss function}.
\end{definition}
We will assume that for any function class $\F$ and task $\T = (X, Y, \ell)$ considered
in this chapter $\E[\ell(f(X), Y)] < \infty$ $\forall f\in\F$.

\subsection{Learning}
The PAC framework \parencite[Definition 3.3]{shalev2014understanding},
originally due to \textcite{valiant1984theory},
provides a precise definition of learning from data.
We use a distribution dependent relaxation of 
the agnostic formulation from \textcite{haussler1992decision}.
What we arrive at can be considered a form of uniform learning \parencite{vapnik2015uniform}.
A \emph{hypothesis class} $\F$ is a set of functions from $\X \to \Y$, its elements
are \emph{hypotheses}.
A \emph{learning algorithm} $\alg: \cup_{i \in \N} \Z^i \to \F$ is a map that associates
with any tuple of observations an element of $\F$.
We say that $\alg$ \emph{learns} $\F$ with respect to a task 
$\T = (X, Y, \ell)$ if $\exists m: (0, 1)^2 \to \N$
such that $\forall \epsilon, \delta \in (0, 1)$,
if $n \ge m(\epsilon, \delta)$ then
\[
    \P\left( \E[\ell(f_S(X), Y) \mid{} S] 
    \le \inf_{f \in \F} \E[\ell(f(X), Y)] + \epsilon\right)
    \ge 1 - \delta
\]
where $f_S = \alg(S)$ and $S \sim (X, Y)^n$.
Throughout this chapter, we assume that $\T$, $\F$ and $\alg$ are such that 
the expectations in the above exist and are finite.
Suppose $\alg$ learns $\F$ with respect to $\T$ and let 
the set of all $m$ satisfying the above be $\mathcal{M}$, we
define the \emph{sample complexity} of $\alg$ on $\D$ as the pointwise minimum
$m_{\alg, \T}(\epsilon, \delta) = \min_{m \in \mathcal{M}}m(\epsilon, \delta)$.

\subsection{Invariant learning algorithms}
We say that a learning algorithm $\alg$ is \emph{$\G$-invariant} 
if 
\[
    \alg((g_1 x_1, g_1y_1), \dots, (g_n x_n, g_ny_n)) 
    = \alg((x_1, y_1), \dots, (x_n, y_n))
\]
$\forall (x_i, y_i) \in \Z$, $\forall g_i \in \G$ and $\forall n \in \N$.
Often we will just say \emph{invariant}.
If the hypothesis class is invariant and $\G$ acts trivially on $\Y$,
then any learning algorithm that depends on the training inputs only through the
values of the hypotheses satisfies this property. 
For instance, any form of empirical risk minimisation is covered by this.
Similarly, if the hypothesis class is equivariant and the learning algorithm 
depends on the data
only through a loss that is \emph{preserved} by $\G$,
meaning $\ell(gy, gy')= \ell(y, y')$ $\forall g\in\G$
$\forall y, y'\in\Y$, then it will be an invariant learning algorithm.
Once again, any form of empirical risk minimisation is covered.
Highlighting two cases: if $\G$ acts via
a unitary representation, then 
$\ell(y, y') = l(\inner{y}{y'})$ is preserved by $\G$ for any
$l$. The same goes for $\ell(y, y') = l(\norm{y - y'}_2)$.

Most of this chapter is about establishing the equivalence of learning problems 
when the hypothesis class is invariant or equivariant.
With this in mind, we clarify what we mean by equivalence.
Intuitively, two tasks are \fgequivalent{} if it's equally difficult to learn
$\F$ with respect to them using any $\G$-invariant learning algorithm.

\begin{definition}\label{def:fgequivalent}
    We say the tasks $\T$ and $\T'$ are \emph{\fgequivalent{}}
    if, for all $\G$-invariant learning algorithms $\alg$, 
    $\alg$ learns $\F$ with respect to $\T$ if and only if
            $\alg$ learns $\F$ with respect to $\T'$, and
    the sample complexities are equal, i.e., $m_{\alg, \T} = m_{\alg, \T'}$.
\end{definition}

\subsection{Orbit Representatives}

\begin{definition}[{\parencite[Definition 4.1]{eaton1989group}}]%
    \label{def:cross-section}
    Let $\G$ be a group acting measurably on a Borel space $(\X, \S_\X)$.
    The set $\X_\pi \subseteq \X$ is a 
    \emph{measurable cross-section} of $\X$ with respect to $\G$
    if the following conditions hold:
    \begin{enumerate}[1)]
        \item $\X_\pi$ is measurable.
        \item $\X_\pi$ contains exactly one element from each orbit of each 
            point $x \in \X$, say $x_\pi$.
        \item The function $\pi: \X \to \X_\pi$ defined by $\pi(x) = x_\pi$ 
            is $\S_\X$-measurable when $\X_\pi$ has the $\sigma$-algebra 
            $\{B \cap \X_\pi : B \in \S_\X\}$.
            We call $\pi$ the \emph{projection}.\label{cond:orbit-mapping}
    \end{enumerate}
\end{definition}

A measurable cross-section provides a natural way of identifying a set
of orbit representatives with suitable measurability properties.
%If $\G$ is compact and acts measurably on $\X$ with respect to $\S_\X$,
%then a measurable cross-section always exists \parencite{bloem2020probabilistic}.
Measurable cross-sections are not necessarily unique.

We maintain some measurable
cross-section $\X_\pi$ with projection $\pi$
in the background throughout this chapter.
It is arbitrary insofar as any additional assumptions are satisfied.

\subsection{Assumption on the Law of $X$}\label{sec:assumption-marginal}
Departing from the rest of this thesis, in this chapter we do not assume that $X$ is 
invariant in distribution unless explicitly stated.
However, we will often make the following (weaker) assumption.

\begin{assumption}\label{assumption:marginal}
We assume $X$ is such that there exists a probability measure $\nu$ on $\G$ such that 
$
    X \eqdist G X_\pi
$
where $X_\pi = \pi(X)$, $G \sim \nu $ and $G \indep X_\pi$.
\end{assumption}

Essentially, \cref{assumption:marginal} says that the orbit that $X$
belongs to and where it is in the 
orbit are independent.
For intuition, if $\G = \SO_2$ acts by rotation on $\R^2$ then any
density $f(r, \theta)$ that is separable in polar co-ordinates as 
$f(r, \theta) = f_{\texttt{rad}}(r) f_{\texttt{ang}}(\theta)$ 
gives the required independence.    

The distribution $\nu$ and the choice of cross-section are not independent
in general.
In particular, if the action is free, 
under the map $\X_\pi \mapsto g\X_\pi$ we have $\nu(A) \mapsto \nu(Ag)$
for all measurable $A\subset \G$.\footnote{%
    Let $g\in\G$ and $\X_{\pi'} = g\X_\pi$. Using \cref{assumption:marginal}
    we can write $X = G \pi(X) = G' \pi'(X) = G'g\pi(X)$
    so, because the action is free, for all measurable $A\subset \G$
    \begin{align*}
        G'^{-1} (A)
        = \set{\omega \in \Omega : G'(\omega) \in A} 
        = \set{\omega \in \Omega : G(\omega)g^{-1} \in A} 
        = \set{\omega \in \Omega : G(\omega) \in Ag}
        = G^{-1} (Ag).
    \end{align*}
    }
The dependence goes away only if $\nu$ is right-invariant,
which by compactness of $\G$ and the uniqueness from 
\cref{thm:bg-haar-measure}
means $\nu = \lambda$, where $\lambda$ is the Haar measure on 
$\G$.
Conversely, there is the following result.

\begin{theorem}[{\parencite[Theorem 4.4]{eaton1989group}}]
    Let $G \sim \lambda$ and let $\X_\pi$ be a measurable cross-section of
    $\X$ with respect to $\G$, then the following are equivalent:
    \begin{enumerate}[(i)]
        \item $X \eqdist g X$ for all $g\in\G$.
        \item There exists a random variable $X_\pi$ taking values on $\X_\pi$
            such that $X \eqdist G X_\pi$ and $G \indep X_\pi$.
    \end{enumerate}
\end{theorem}

\section{Learning with Invariant Models}
In this section we present \cref{thm:invariance}, 
which is a rigorous version of the common intuition that
learning with an invariant model is equivalent to learning on a space of
orbit representatives.
We show that these learning problems have the same sample complexity.
The class of orbit representatives may have much smaller dimension or complexity than the
input space, which could result in a reduction in sample complexity for invariant models.

\begin{theorem}\label{thm:invariance}
    Suppose $\G$ acts measurably on $\X$ and trivially on $\Y$.
    Let $\X_\pi$ be a measurable cross-section of $\X$ 
    with projection $\pi$.
    Let $\F$ be a hypothesis class of $\G$-invariant functions.
    Let $\T = (X, Y, \ell)$ be any task, write $X_\pi = \pi(X)$
    and define $\T_\pi = (X_\pi, Y, \ell)$.
    Let $X$ satisfy \cref{assumption:marginal}.
    Then $\T$ and $\T_\pi$ are \fgequivalent{}.
\end{theorem}
\begin{proof}
    For any $S= ((X_1,Y_1),\dots, (X_n, Y_n))$
    define $S_\pi = ((\pi(X_1),Y_1),\allowbreak\dots,\allowbreak (\pi(X_n), Y_n))$.
    If $S \sim (X, Y)^n$ then $S_\pi \sim (X_\pi, Y)^n$.
    Let $\alg$ be a $\G$-invariant learning algorithm.
    Set $f_S = \alg(S)$ and $f_{S_\pi} = \alg(S_\pi)$.
    The invariance of $\alg$ implies $f_S  =f_{S_\pi}$.
    By the invariance of $\F$, $f(X) = f(X_\pi)$ for all $f\in \F$.
    Together with the independence in \cref{assumption:marginal}, this means that
    $
        \E[\ell(f_S(X), Y) \mid S] = \E[\ell(f_{S_\pi}(X_\pi), Y) \mid S_\pi]
        $. 
    We can then conclude that
    \begin{align*}
        &\P\left( \E[\ell(f_S(X), Y)\mid S] 
        \le \inf_{f\in \F} \E[\ell(f(X), Y)] + \epsilon\right)\\
        &= 
        \P\left(\E[\ell(f_{S_\pi}(X_\pi), Y)\mid S_\pi] 
        \le \inf_{f\in \F} \E[\ell(f(X_\pi), Y)] + \epsilon\right)
    \end{align*}
    which completes the proof.
\end{proof}

\begin{example}
    Let $\X = \R{}^2$ and $\Y = \{0, 1\}$. \Cref{thm:invariance} tells us that learning with
    a rotationally invariant hypothesis class, such as discs about the origin
    $
        \F = \{(x, y) \mapsto \1{x^2 + y^2 \le r} : r \in \R{}_+\}
        $,
    is equivalent to learning on the reduced space $\X = \R_+$.
\end{example}

\begin{example}[Deep Sets, \parencite{zaheer2017deep}]
    \textcite{zaheer2017deep,bloem2020probabilistic}
    consider learning functions $f: [0, 1]^d \to \R{}$ that are
    $\sym_d$-invariant, where $\sym_d$ is the group of permutations on $d$ elements.
    It is shown that any such continuous function must be of the form 
    $
        f(T) = f_1\left(\sum_{t \in T} f_2(t)\right)
        $
    where $T \in [0, 1]^d$ and $f_1, f_2: \R \to \R$.
    By \cref{thm:invariance} we can see, as was shown by \textcite{sannai2021improved},
    that learning permutation invariant functions
    is equivalent to learning the same class of functions restricted to the domain
    $
        \{x_1 \ge x_2 \ge \dots \ge x_d; x_i \in [0, 1]\}.
        $
    In high dimensions this is a much smaller space than $[0, 1]^d$,
    it is a factor of $d!$ smaller in volume.
\end{example}

\begin{example}[G-CNN \parencite{cohen2016group}]
    \textcite{cohen2016group} present 
    a convolutional layer that is equivariant to
    various discrete groups of transformations.
    These layers are used to generate features for an invariant
    classifier. Consider the simple case of the group $p4$ of rotations
    about the origin in $\R^2$ through an angle of 
    $\frac{\pi}{2}$. \Cref{thm:invariance} shows that training a 
    $p4$-invariant G-CNN based classifier
    (e.g., by gradient descent on the empirical loss)
    is equivalent to learning the
    network restricted to a single quadrant of the plane.
\end{example}

For the final corollary we need an additional definition.
\begin{definition}[Packing, packing number]
    Let $(T, \tau)$ be a metric space and 
    $\epsilon > 0$. $E \subset T$ is an \emph{$\epsilon$-packing} of $T$
    (with respect to $\tau$) if for all $x, y \in E$, $\tau(x, y) > \epsilon$.
    The \emph{$\epsilon$-packing} number is the largest cardinality of all 
    the $\epsilon$-packings, 
    $\pack(T, \tau, \epsilon) = \sup_{E \in \mathcal{E}} \abs{E} $ 
    where $\mathcal{E}$ is the set of all $\epsilon$-packings.
    If the supremum doesn't exist, then we say the packing number is infinite.
\end{definition}

\begin{corollary}
    Let $(\X, \tau)$ be a compact metric space,
    let $\Y = \{0, 1\}$ and
    let $\F$ be a hypothesis class (a set of binary classifiers).
    Assume that the functions in $\F$ are \emph{$\gamma$-robust},
    meaning $\forall f \in \F:$ $f(x) \ne f(x') \implies \tau(x, x') \ge \gamma$.
    This implies that the VC-dimension of $\F$ is bounded 
    $\vc{\F} \le \pack(\X, \tau, \gamma )$.
    If in addition we assume that $\F$ is invariant then we know from \cref{thm:invariance} 
    that
    $\vc{\F} \le \pack(\X_\pi, \tau, \gamma ) \le \pack(\X, \tau, \gamma )$,
    suggesting a distribution independent sample complexity improvement for
    invariant hypotheses.
\end{corollary}

\section{Learning with Equivariant Models}\label{sec:equivariance}

Before we present our result for equivariant learning problems,
we introduce an additional assumption.
We relate \cref{assumption:targets} to the work 
of \textcite{bloem2020probabilistic} in \cref{sec:discussion-of-assumptions}.

\begin{assumption}\label{assumption:targets}
    We assume that there exists a measurable $f: \X\times[0, 1] \to \Y$ such that
    $Y \eqdist f(X, \eta)$, where $\eta \sim \unif[0,1]$, $\eta \indep X$
    and $f$ satisfies the equivariance property 
    $f(gX, \eta) \eqdist gf(X, \eta)$.
\end{assumption}
One can verify that, with reference to \cref{assumption:marginal},
$\eta\indep G$ and $\eta \indep X_\pi$.

A special case of \cref{assumption:targets} is an equivariant target 
function with independent additive noise
$f(X, \eta) = f^*(X) + \xi$ where
$\xi$ is such that $g\xi \eqdist \xi$ for all $g \in \G$ (e.g., orthogonal representation 
on $\Y$ and Gaussian noise).
\Cref{assumption:targets} is more general and allows for stochastic equivariant
functions and noise corruption that is not necessarily additive.
This is inspired by \emph{noise outsourcing}, 
which typically appears with almost sure equality rather
than equality in distribution. It is also known as 
\emph{transfer} \parencite[Theorem 6.10]{kallenberg2006foundations}.
See \parencite{bloem2020probabilistic} for an application to invariance/equivariance.

\begin{theorem}\label{thm:equivariance}
    Let $\G$ act measurably on both $\X$ and $\Y$.
    Let $\F$ be a $\G$-equivariant hypothesis class.
    Let $(X, Y)$ satisfy \cref{assumption:marginal} and \cref{assumption:targets}.
    Then the tasks $T = (X, Y, \ell)$ and $T_\pi = (X_\pi, Y_\pi, \bar{\ell})$ 
    are \fgequivalent{}, 
    where $X_\pi = \pi(X)$, $Y_\pi = f(X_\pi, \eta)$ with $f$ and $\eta$ 
    as in \cref{assumption:targets}
    and
    $
        \bar{\ell}(y, y') = \int_\G \ell(gy, gy') \dd \nu(g)
        $.
\end{theorem}
\begin{proof}
    Let $S \sim (X, Y)^n$ and $S_\pi \sim (X_\pi, Y_\pi)^n$.
    We have from \cref{assumption:marginal} that 
    $X \eqdist GX_\pi$ and from \cref{assumption:targets}
    that
    \[
        Y \eqdist f(X, \eta) \eqdist f(GX_\pi, \eta) \eqdist Gf(X_\pi, \eta) = GY_\pi.
    \]
    Let $\alg$ be an invariant learning algorithm, then
    we have $f_S \coloneqq \alg(S) = \alg(S_\pi) \eqqcolon f_{S_\pi}$
    and hence by \cref{assumption:marginal} we have
    $\E[\ell(f_S(X), Y) \mid{} S] = \E[\ell(f_{S_\pi}(X), Y)\mid{}S_\pi]$.
    Then, for all 
    $h\in\F$, \cref{assumption:marginal}, \cref{assumption:targets} and
    equivariance gives
    \begin{align*}
        \E[\ell(h(X), Y)] 
        &= \E[\ell(h(GX_\pi), f(GX_\pi, \eta))]\\
        &= \E[\ell(Gh(X_\pi), Gf(X_\pi, \eta))]\\
        &= \E[\bar{\ell}(h(X_\pi), Y_\pi)]
    \end{align*}
    where we applied Fubini's theorem.
    The function $(g, y, y') \mapsto \ell(gy, gy')$ is measurable by composition
    (see the proof of \cref{prop:q-well-defined} for analogous details)
    so every $(\Y\times \Y)$-section $\ell_{y, y'}(g) = \ell(gy, gy')$ 
    is $\nu$-measurable by \cref{lemma:section},
    so $\bar{\ell}$ exists.
    Furthermore, all of the above expectations must be finite because the left
    hand side is finite by assumption.
    Putting everything together concludes the proof
    \begin{align*}
        &\P\left(
        \E[\ell(f_S(X), Y) \mid S] \le \inf_{f\in \F} \E[\ell(f(X), Y)] + \epsilon\right) \\
        &= 
        \P\left(
        \E[\bar{\ell}(f_{S_\pi}(X_\pi), Y_\pi)\mid{} S_\pi] \le 
        \inf_{f \in \F} \E[\bar{\ell}(f(X_\pi), Y_\pi)] + \epsilon\right).
    \end{align*}
\end{proof}

The loss function $\bar{\ell}$ is the average of $\ell$ over the orbits of $\G$, weighted
by the probability of each $g\in\G$.
If $\ell$ is preserved by $\G$, that is $\ell(gy, gy') = \ell(y, y')$ 
for all $y, y' \in \Y$,
we have $\bar{\ell} = \ell$.
One can think of $Y_\pi$ as the canonical target
corresponding to the canonical input $X_\pi$.
The task $(X_\pi, Y_\pi, \ell)$ can be thought of as a canonical version of
the original task, with any nuisance information from the group transformations
removed. This canonical task depends the choice of cross-section.

\begin{example}[Deep Sets, \cite{zaheer2017deep}]
    Returning to \cite{zaheer2017deep}, the authors consider neural network layers
    $f: \R^d \to \R^d$ with $f(x) = \sigma(\Theta x)$,
    where $\Theta \in \R^{d \times d}$ and $\sigma$ is an element-wise
    non-linearity. It is shown that $f$ is $\sym_d$-equivariant if and only if 
    $
        \Theta_{ij} = a\delta_{ij} + b
    $
    for scalars $a, b \in \R$.
    In this case $\X = \Y = \R^d$ and $\sym_d$ acts on each
    space by permutation.
    Assume that the marginal distribution on the inputs is
    exchangeable,
    so $\nu$ exists and is uniform on $\sym_d$ as outlined at the end 
    of \cref{sec:assumption-marginal},
    then \cref{thm:equivariance} says that 
    learning the class of equivariant $f$
    is equivalent to learning on the restricted domain
    $
        \{x_1 \ge x_2 \ge \dots \ge x_d; x_i \in \X \}
     $
    with the averaged loss function
    $
        \bar{\ell}(y, y') = \frac{1}{d!}\sum_{s \in \sym_d} \ell(y_s, y_s')
    $
    where $(y_s)_i = y_{s(i)}$ and the same for $y'$.
\end{example}

\subsection{Discussion of \cref{assumption:targets}}\label{sec:discussion-of-assumptions}
It is known that, under certain conditions, a functional representation for $Y$
similar to the one in \cref{assumption:targets}
is equivalent to the conditional equivariance of $Y$, in the sense that 
$gY \mid{} gX \eqdist Y \mid{} X$ $\forall g \in \G$ \parencite{bloem2020probabilistic}.
We describe this setting below.

Assume that the distribution of $X$ is $\G$-invariant,
so $X \eqdist g X$ $\forall g\in\G$. 
We adapt the following definition from \textcite{bloem2020probabilistic}.

\begin{definition}[Representative Equivariant]\label{def:representative-equivariant}
    Let $\G$ be a group acting freely on a set $\X$.
    A \emph{representative equivariant} is an equivariant function $\tau: \X \to \G$.
    That is, $\tau(gx) = g\tau(x)$ $\forall g\in\G$ $\forall x \in \X$.
\end{definition}

The following result from \textcite{bloem2020probabilistic} then gives us
\cref{assumption:targets}.

\begin{theorem*}[{\textcite[Theorem 9]{bloem2020probabilistic}}]
    Let $\G$ be a compact group acting measurably on Borel spaces $\X$ and $\Y$
    such that there exists a measurable representative equivariant $\tau: \X \to \G$.
    Let $X$ be a $\G$-invariant random element of $\X$.
    Then $Y$ is conditionally $\G$-equivariant if and only if there's a measurable
    $\G$-equivariant function $f: \X \times [0, 1] \to \Y$ such that 
    $Y \eqas f(X, \eta)$ where $\eta \sim \unif[0, 1]$ and $\eta \indep X$.
\end{theorem*}

\section{Implication for Empirical Risk Minimisation}\label{sec:sample-complexity}
We apply the equivalences derived in previous
sections to derive improved sample complexity guarantees for equivariant
models when learning with empirical risk minimisation.
In particular, \cref{thm:sample-complexity} links the \emph{generalisation error},
defined to be the difference between the risk and the empirical risk, 
to the geometries of the input and output spaces.
Using invariant or equivariant hypotheses reduces the learning
problem to one on a cross-section $\X_\pi$. 
Often, $\X_\pi$ will be smaller than $\X$ and,
because the sample complexity depends a notion of the size of the input space,
we get a reduction for invariant/equivariant models.
We use covering numbers for this notion of size.

\begin{definition}[Covering, covering number]
    Let $(T, \tau)$ be a metric space and let $U\subset T$.
    $K\subset T$ is an \emph{$\epsilon$-cover} of $U$ 
    if $\forall u \in U$ $\exists k \in K$
    with $\tau(u, k) \le \epsilon$.
    Let $\mathcal{K}$ be the set of all $\epsilon$-covers of $U$,
    the \emph{$\epsilon$-covering number} of $U$ is 
    the smallest cardinality of all the $\epsilon$-covers
    $\cover(U; \tau, \epsilon) = \inf_{K \in \mathcal{K}} \abs{K} $. 
\end{definition}

\Cref{thm:sample-complexity} relies upon \cref{prop:concentration}
which is deferred to \cref{sec:concentration}.
The basic idea of this proof is 
to use the Lipschitz property to bound coverings of the hypothesis class in 
terms of coverings of the input space.
Similar ideas appeared in \parencite{sokolic2017generalization,sannai2021improved}.

\begin{theorem}\label{thm:sample-complexity}
    Let $(\X, \tau)$ be a metric space
    and let $\Y \subset \R^d$ be convex.
    Let $\F$ be a hypothesis class of functions $f: \X \to \Y$ that are 
    $L$-Lipschitz 
    $\norm{f(x) - f(x')}_\infty \le L\tau(x, x')$ $\forall x, x'\in\X$.
    Let $\ell(y, y'): \Y\times\Y\to\R_+$ be measurable loss function 
    that's bounded $\ell(y, y')\le M$ and Lipschitz in its first argument
    $\abs{\ell(y_1, y') - \ell(y_2, y')} \le C_\ell \linf{y_1 - y_2}$ 
    for some
    $C_\ell \in \R_+$ that's independent of $y_1$, $y_2$ and $y'$.
    Let $S \sim (X, Y)^n$ be a training sample. 
    Then, for all $\epsilon \in (0, 1)$,
    \begin{align*}
        &\P\left( 
        \sup_{f \in \F}\labs{\E[\ell(f(X), Y)] 
        - \frac1n \sum_{i=1}^n \ell(f(X_i), Y_i) }\ge \epsilon \right)\\
        &\le 
        2 \inf_{\alpha \in (0, 1)}\exp\left(
        D_{\alpha \epsilon}(\X, \F)
        -\frac18 
        (1 - \alpha)^2 n\epsilon^2 M^{-2}
        \right)
    \end{align*}
    where 
    \[
    D_{t}(\X, \F) = 
    \cover\left(\X, \tau, \frac{t}{12LC_\ell}\right) 
    \sup_{x \in \X}
    \log 
    \left(\cover{}\left(\F(x),
    \norm{\cdot}_\infty,\frac{t}{12L^2C_\ell}\right)\right)
    \]
    and $\F(x) = \{ f(x): f\in\F\} \subset \Y$.
\end{theorem}

\begin{proof}
    We estimate the covering number of $\F$ and plug it 
    into \cref{prop:concentration}.
    Let $E$ be a minimal $\delta$-cover for $\X$.
    We may assume that $\abs{E} < \infty$, because otherwise
    the result is trivial.
    For all $x\in\X$ 
    let $F_x$ be a minimal $\kappa$-cover of $\{f(x): f \in \F\}\subset \Y$
    in the vector norm $\norm{\cdot}_\infty$ on $\Y$.
    Let $H_E$ be the set of 
    all functions $h_E: E \to \Y$ such that $\forall x \in E$, $h_E(x) \in F_x$.
    For all $x \in \X$ let the set of closest elements in $E$ be
    $A(x) = \{x' \in E: \tau(x, x') = \min_{\tilde{x}\in E} \tau(x, \tilde{x})\}$.
    For any $h_E \in H_E$ define its extension $h :\X \to \Y$ by
    \[
        h(x) = 
        \begin{cases}
            h_E(x) \quad x\in E \\
            \frac{1}{\abs{A(x)}} \sum_{x' \in A(x)} h_E(x') \quad x\not\in E
        \end{cases}
    \]
    and let $H$ be the set of all such extensions for $h_E\in H_E$.
    Let $f\in \F$, $x \in \X$, $x' \in A(x)$ and $h\in H$ then
    \begin{align*}
        \norm{f(x) - h(x)}_\infty
        &= \norm{ f(x) - f(x') + f(x') - h(x') + h(x') - h(x)}_\infty\\
        &\le L\tau(x, x') + \norm{f(x') - h(x')}_\infty + \norm{h(x') - h(x)}_\infty.
    \end{align*}
    By assumption there's an $h_f\in H$ such that $\norm{f(x') - h_f(x')}_\infty \le \kappa$
    because $x' \in E$. Any such $h_f$ gives
    \begin{equation}\label{eq:inf-norm-bound}
        \norm{f(x) - h_f(x)}_\infty
        \le L\delta + \kappa + \norm{h_f(x') - h_f(x)}_\infty.
    \end{equation}
    We will make use of the following fact about $h_f$. Let $x_1, x_2 \in A(x)$, then
    \begin{align*}
        \norm{h_f(x_1) - h_f(x_2)}_\infty
        &= \norm{h_f(x_1) - f(x_1) + f(x_1) - f(x_2) + f(x_2) - h_f(x_2)}_\infty\\
        &\le 2 \kappa + L \tau(x_1, x_2)\\
        &\le 2 \kappa + 2L\delta
    \end{align*}
    because $\tau(x_1, x_2) \le \tau(x_1, x) + \tau(x_2, x) \le 2\delta$.
    This then gives
    \begin{align*}
        \norm{h_f(x') - h_f(x)}_\infty
        &= \lnorm{h_f(x') - 
            \frac{1}{\abs{A(x)}} \sum_{\tilde{x} \in A(x)} 
            h_f(\tilde{x})}_\infty\\
        &\le \frac{1}{\abs{A(x)}} 
            \sum_{\tilde{x} \in A(x) \setminus \{x'\}} 
            \norm{h_f(\tilde{x}) - h_f(x')}_\infty\\
        &\le 2 \kappa + 2L\delta.
    \end{align*}
    Putting this into \cref{eq:inf-norm-bound} gives
    \[
        \norm{f(x) - h_f(x)}_\infty
        \le 3 \kappa + 3L\delta
    \]
    which shows that $H$ is a $(3 \kappa + 3L\delta)$-cover for $\F$ in the
    function norm $\linf{\cdot}$. So setting $\kappa = L\delta$ gives
    \[
        \cover{}(\F, \linf{\cdot}, 6L\delta)
        \le \abs{H}
        = \prod_{x\in E} \abs{F_x}
        \le 
        \sup_{x \in \X} \cover{}(\F(x), \norm{\cdot}_\infty,\delta/L)^{\cover(\X, \tau, \delta)}.
    \]
    Using this in \cref{prop:concentration} gives the statement.
\end{proof}

The above result says, taking $\alpha=1/2$ for simplicity, that 
\[
    n = \Omega\left( \frac{ D_{\frac{\epsilon}{2}}(\X, \F) + \log(1/\delta)}{\epsilon^2}\right)
\]
examples are sufficient to have generalisation error at most $\epsilon$ with
probability at least $1-\delta$.
The significance of this for equivariant models is as follows.
Suppose that $\G$ acts on $\X$ and $\Y$, that the loss is preserved by $\G$,
e.g., squared-error loss and an orthogonal representation on $\Y$,
and that $(X, Y)$ satisfy \cref{assumption:marginal} and \cref{assumption:targets}.%
\footnote{%
    If $\Y = B_r(d)$ the closed Euclidean ball in $\R^d$ and 
    $\ell(y, y') = \norm{y - y'}^2_2$ is the squared-error loss,
    then this satisfies the statement of \cref{thm:sample-complexity}
    with $M = r^2$ and $C_\ell = 4r\sqrt{d}$.%
    }
In this setting, we can consider the change to the bound in \cref{thm:sample-complexity} 
that arises from taking $\F$ to be $\G$-equivariant.
Specifically, \cref{thm:equivariance} tells us
that the sample complexity of learning on the task $\T = (X, Y, \ell)$ is the same as 
learning on the task $\T' = (X_\pi, Y_\pi, \ell)$. This means that 
\[
    n = \Omega\left( \frac{ D_{\frac{\epsilon}{2}}(\X_\pi, \F) + \log(1/\delta)}%
    {\epsilon^2}\right)
\]
examples are sufficient for an equivariant model to have generalisation error at most $\epsilon$,
potentially much less than above.
The quantity $D_t(\X, \F)$
is, in a sense, simultaneously measuring the size of $\X$ and the outputs of $\F$ at a 
scale $t$.
The benefit of equivariance depends on the geometry of $\X$ and $\X_\pi$. 
Since $\X_\pi \subset \X$ we get $D_t(\X_\pi, \F) \le D_t(\X, \F)$ and, informally, 
the reduction will be large if $\X_\pi$ is much smaller than $\X$.
If $\F(\X_\pi)$ is much smaller than $\F(\X)$ then the same applies.

\subsection{Concentration of Measure}\label{sec:concentration}
The following is adapted from \parencite[Exercise 3.31]{mohri2018foundations}.
\begin{proposition}\label{prop:concentration}
    Let $\X$ be a set and let $\Y \subset \R^d$ be convex.
    Let $\F$ be a class of measurable functions $f: \X \to \Y$.
    Let $\ell(y, y'): \Y\times\Y\to\R_+$ be measurable loss function 
    that's bounded $\ell(y, y')\le M$ and Lipschitz in its first argument
    $\abs{\ell(y_1, y') - \ell(y_2, y')} \le C_\ell \linf{y_1 - y_2}$ 
    for some
    $C_\ell \in \R_+$ that's independent of $y_1$, $y_2$ and $y'$.
    Let 
    $S =((X_1, Y_1),\allowbreak \dots,\allowbreak (X_n, Y_n)) \sim (X, Y)^n$
    be a training sample 
    where $X$ and $Y$ are arbitrary random elements of $\X$ and $\Y$ respectively.
    Then, 
    \begin{align*}
        &\P\left( 
        \sup_{f \in \F}\labs{\E[\ell(f(X), Y)] 
        - \frac1n \sum_{i=1}^n \ell(f(X_i), Y_i) }\ge \epsilon \right)\\
        &\le 
        2 \inf_{\alpha \in (0, 1)}
        \cover\left(\F, \linf{\cdot}, \frac{\alpha\epsilon}{2 C_\ell}\right)
            \ee^{-\frac18 (1 - \alpha)^2 n\epsilon^2 M^{-2} }.
    \end{align*}
\end{proposition}

\begin{proof}
    For any probability measure $\varsigma$ on $\X \times \Y$ define
    $
        R_\varsigma[f] = \E[\ell(f(A), B)] 
        $
    where $(A, B) \sim \varsigma$.
    Then for all $f, f' \in \F$
    \begin{align*}
        R_\varsigma[f] - R_\varsigma[f']
        &= \E[\ell(f(A), B) - \ell(f'(A), B)] \\
        &\le C_\ell \E[\linf{f(A) - f'(A)}]\\
        &\le C_\ell \linf{f - f'}
    \end{align*}
    Now let 
    \[
        L_S(f) = \frac1n \sum_{i=1}^n \ell(f(X_i), Y_i) - \E[\ell(f(X), Y)]
    \]
    which has mean 0 for all $f$.
    By setting $\varsigma$ as the empirical measure on 
    $S$ and then as the distribution of $(X, Y)$,
    one finds that 
    \begin{equation}\label{eq:loss-bound}
        \abs{L_S(f) - L_S(f')} \le 2C_\ell \linf{f - f'}.
    \end{equation}
    Now let $\mathcal{K}$ be a $\kappa$-cover of $\F$ in $\linf{\cdot}$.
    Define the sets $D(k) = \{f \in \F: \linf{f-k} \le \kappa\}$.
    Then
    \begin{align*}
        \P\left(\sup_{f\in\F}\abs{L_S(f)} \ge \epsilon\right)
        &\le \P\left(\bigcup_{k\in \mathcal{K}} 
        \left\{\sup_{f \in D(k)} \abs{L_S(f)} \ge \epsilon\right\}\right) \\
        &\le \sum_{k \in \mathcal{K}} \P\left(\sup_{f \in D(k)} \abs{L_S(f)} \ge \epsilon\right).
    \end{align*}
    Set $\kappa = \frac{\alpha\epsilon}{2C_\ell}$ for $0 < \alpha < 1$.
    Using \cref{eq:loss-bound}, for all $f \in D(k)$ we have 
    \[
        L_S(f) \le 2C_\ell\kappa + L_S(k) = \alpha\epsilon + L_S(k),
    \]
    hence
    \[
        \P\left(\sup_{f\in\F}\abs{L_S(f)} \ge \epsilon\right)
        \le \sum_{k \in \mathcal{K}} \P(\abs{L_S(k)} \ge (1 - \alpha)\epsilon).
    \]
    Then Hoeffding's inequality gives
    \[
        \P\left(\sup_{f\in\F}\abs{L_S(f)} \ge \epsilon\right)
        \le 2\abs{\mathcal{K}} \exp\left(-\frac{2 (1 - \alpha)^2 n \epsilon^2}{16M^2}\right)
    \]
    where we used the bound on $\ell$ in the statement.
\end{proof}

\chapter{Connections and Extensions}\label{chap:outlook}

\section*{Summary}
This chapter is more high-level than the others.
In \cref{sec:connection} we relate the orbit averaging viewpoint on invariance
from \cref{chap:general-theory-i} to the orbit representative viewpoint 
from \cref{chap:general-theory-ii}.
In \cref{sec:neural-networks} we apply some of the ideas in this thesis to neural networks.
We outline some connections to other works in \cref{sec:other-work-comparison}.
Finally, in \cref{sec:future-work}, we give some suggestions for future work.

\section{Connections between Orbit Averaging and Orbit Representatives}\label{sec:connection}
\Cref{chap:general-theory-ii} operates with a different perspective on
invariant functions than developed in \cref{chap:general-theory-i}.
The former expresses invariance of functions in
terms of their dependence only on their values on a cross-section 
(one element from each orbit of $\X$ under the action of $\G$),
while the latter uses orbit averaging. 
The purpose of this section is to give an intuitive connection between
these viewpoints in the case of invariance. 
We leave equivariance to future work.

Consider $f\in\Lmu$.
By \cref{prop:sym-cpt}, $f$ is invariant if and only if
$\O f = f$. On the other hand, fixing a measurable cross-section $\X_\pi$
with projection $\pi$ we can get an invariant function by $f_\pi = f\circ \pi$,
and conversely \cref{thm:maximal-invariant} says that for all choices of $\X_\pi$ 
and all invariant $f$ there's a function $h$ such that $f = h_\pi$.
However, although the map $f\mapsto f_\pi$ is a projection into invariant functions,
it's not equivalent to $\O$.
Decomposing $f  = \O f + f^\sperp$ gives $f_\pi = \O f + f^\sperp_\pi$
so $f_\pi = \O f $ would require $f^\sperp$ to vanish on $\X_\pi$. This is not the
case for many choices of $\X_\pi$, for instance see the example in \cref{fig:orbit}.

Instead, orbit averaging with $\O$ can be interpreted as averaging over all 
images of the cross-section under the action of $\G$, i.e., $\set{g\X_\pi: g\in\G}$.
Define $g_\pi: \X \to \G$ such that for all $x\in\X$ 
$g_\pi(x)$ is any solution to $g_\pi(x)\pi(x) = x$,
then, using the invariance of $\lambda$,
\begin{align*}
    \O f(x)
    &= \int_\G f(gx) \dd\lambda(g)\\
    &= \int_\G f(gg_\pi \pi(x)) \dd\lambda(g)\\
    &= \int_\G f(g \pi(x)) \dd\lambda(g)\\
    &= \int_\G f((g\circ\pi)(x)) \dd\lambda(g)\\
    &= \int_\G f_{g\circ\pi}(x) \dd\lambda(g)
\end{align*}
where $g\circ\pi$ projects onto $g\X_\pi$.

A related view from \parencite[Section 5.1]{eaton1989group} is as follows.
Let $\X$ be a locally compact, second countable and Hausdorff topological space.
Let $\K$ be the set of all continuous $f:\X\to\R$ with compact support.\footnote{%
By compact support, we mean the set $\set{x\in\X:f(x)\ne0}$ is compact.}
Then for all $f\in \K$ there's a unique 
function on the quotient space $\tilde{f}:\X/\G \to \R$ (also continuous with compact support)
such that for all $x\in\X$
\[
    \O f(x) = \tilde{f}(p(x))
\]
where $p: \X \to\X/\G$ is the map that sends each point to its orbit $p(x) = \set{gx:g\in\G}$.
The relation can be summarised in the following commutative diagram.
\[
  \begin{tikzcd}
    \X \arrow{r}{p} \arrow[swap]{dr}{\O f} 
      & \X/\G \arrow{d}{\tilde{f}} \\
     & \R
  \end{tikzcd}
\]
This reinforces the interpretation of $\O$ as averaging over all possible cross-sections.
Indeed, this viewpoint is somewhat more elegant, since it deals directly with $\X/\G$
and doesn't require the choice of a cross-section.

\section{Applications to Neural Networks}\label{sec:neural-networks}
Let $F: \R^d \to \R^k$ be a feedforward neural network with $L$ layers,
layer widths $\kappa_i$ $i=1, \dots, L$ and weights $W^i \in \R^{\kappa_{i+1} \times \kappa_{i}}$
for $i=1, \dots, L$ where $\kappa_1 = d$ and $\kappa_{L+1} = k$. We will assume $F$ has the form
\begin{equation}\label{eq:nn}
    F(x) = W^L \sigma(W^{L-1}\sigma(\cdots \sigma(W^{1} x)\cdots))
\end{equation}
where $\sigma:\R\to\R$ is a non-linearity applied element-wise. 
We refer to this architecture as the multi-layer perceptron (MLP).

\subsection{Invariant and Equivariant Networks}\label{sec:invariant-networks}
The standard method for engineering MLPs to be invariant/equivariant
to the action (specifically, a finite-dimensional representation)
of a group on the inputs is weight tying. 
This method has been around for a while \parencite{wood1996representation}
but has come to recent attention via \textcite{ravanbakhsh2017equivariance}.
We outline this approach, give an estimate of the VC dimension of the resulting networks
and apply the ideas of
this work to propose a new method for learned invariance/equivariance.

The basic idea in 
\parencite{wood1996representation,ravanbakhsh2017equivariance} can be summarised as follows.
Let $\G$ be a compact group admitting the necessary representations.
For each $i=2, \dots, L+1$, the user chooses a matrix representation $\psi_i:\G \to \GL_{\kappa_i}$
of $\G$ that acts on the inputs for each layer $i=2, \dots, L$
and on the outputs of the network when $i=L+1$.%
\footnote{$\psi_1$ is the representation on the inputs, which we consider as an
aspect of the task and not a design choice.}
To avoid degeneracies we assume that the representations are finite,
in the sense that for each $i$ the set $\set{\psi_i(g): g\in \G}$ 
forms a finite matrix group under multiplication.
For $i=2, \dots, L$, these representations must be chosen
such that they commute with the activation function 
\begin{equation}\label{eq:activation}
  \sigma(\psi_i(g) \cdot) = \psi_i(g) \sigma(\cdot)  
\end{equation}
for all $g \in \G$.%
\footnote{This condition is somewhat restrictive,
but note that a permutation representation will commute with
any element-wise non-linearity. For a description of admissible combinations of
representations and activation functions see \parencite[Theorem 2.4]{wood1996representation}.}
One then chooses weights for the network such that at each layer and for all $g \in \G$
\begin{equation}\label{eq:intertwine}
    W^i\psi_i(g) = \psi_{i+1}(g)W^i
\end{equation}
(where there is no implicit sum of $i$).
By induction on the layers, satisfying \cref{eq:activation,eq:intertwine}
ensures that the network is equivariant.
The network is invariant if $\psi_{L+1}$ is the trivial representation.

The condition in \cref{eq:intertwine} can be phrased as saying that that $W^i$ belongs
to the space of \emph{intertwiners} of the representations $\psi_i$ and $\psi_{i+1}$.
By denoting the space of all possible weight matrices in layer $i$ as 
$U = \R^{ \kappa_{i+1} \times \kappa_i }$,
the space of intertwiners is immediately recognisable as $\twine^i(U)$ 
where $\twine^i$ is the linear map with components
\begin{equation}\label{eq:nn-twine-op}
    \twine^i_{abce} = \int_\G  \psi_{i+1}(g^{-1})_{ac}\psi_{i}(g)_{eb}\dd \lambda(g)
\end{equation}
which acts by $\twine^i(W)_{ab} = \twine^i_{abce}W_{ce}$.
This first appeared in \cref{eq:intertwine-linear}, although defined slightly
differently, where $\twine$ was introduced as the
restriction of the operator $\qq$ to linear maps.

We emphasise that the definition of $\twine^i$
does not require the representations to be orthogonal.
Strictly, according to our definition of $\qq$, this means
that $\twine^i$ defined in \cref{eq:nn-twine-op} is not the restriction of $\qq$ to linear maps 
$\R^{\kappa_i} \to \R^{\kappa_{i+1}}$. However, it is still 
a linear projection onto equivariant elements and we can still write
$W = \overline{W} + W^\sperp$ where $\overline{W} = \twine(W)$ and
$\twine(W^\sperp) = 0$, we just lose the orthogonality in \cref{lemma:l2-decomposition}.
It remains well defined because the representations are finite.

\subsubsection{VC Dimension}
The following result 
is a corollary of \parencite[Theorem 6]{bartlett2019nearly}
which 
gives an estimate of the VC dimension of invariant
MLPs with ReLU activation of the kind discussed above.
Similar results are possible for other activation functions.
It applies to the case where the network is layer-wise equivariant and
$\psi_{L+1}$ is the trivial representation.

\begin{proposition}\label{proposition:invariant-vc}
    Consider a $\G$-invariant MLP architecture with ReLU activations and
    weights constrained to intertwine the representations as described above.
    Let $\F$ be the set of binary classifiers of the form
    $\sign(F)$ for all functions $F$ computed by this architecture.
    Then 
    \[
        \vc{\F} 
        \le 
        L 
        +  \frac{1}{2} \alpha L(L+1)\max_{1 \le i \le L} \binner{\chi_{i}}{\chi_{i+1}}
    \]
    where $\alpha 
    = \log_2\left(
        4 \ee \log_2\left(\sum_{i=1}^L 2\ee i \kappa_i\right) \sum_{i=1}^L i\kappa_i
    \right)$, $\chi_i(g) = \tr(\psi_i(g))$ are
    the characters of the representations
    and
    $
        \binner{\chi_i}{\chi_{i+1}} 
        = \int_\G \chi_i(g) \chi_{i+1}(g) \dd\lambda(g)
        $
    is their inner product.
\end{proposition}

\begin{proof}
    For a ReLU network with $t_i$ independent parameters at each layer we have
    \[
        \vc{\F} \le L + \alpha \sum_{i=1}^L (L-i+1) t_i,
    \]
    which is by direct application of \parencite[Theorem 6]{bartlett2019nearly}.
    The condition \cref{eq:intertwine} is the statement that the weight matrix 
    $W^i$ belongs to the space of equivariant maps $\R^{\kappa_{i}}\to \R^{\kappa_{i+1}}$.
    The number of independent parameters at each layer is at most the dimension of this space,
    which is $\binner{\chi_i}{\chi_{i+1}}$.
    The conclusion follows easily.
    % comment that I think I don't need...
    %The proof of \parencite[Theorem 6]{bartlett2019nearly} depends on
    %the representation of the network in terms of piecewise polynomials of bounded degree.
    %Weights are tied only within layers 
    %so weights in different layers can vary independently,
    %and the activation is ReLU, so there is no increase in the degree of said polynomials
    %from weight-tying and the proof given in \parencite{bartlett2019nearly} applies in our case.
\end{proof}

\begin{example}[Permutation invariant networks]
    Permutation invariant networks are 
    studied in many other works, see 
    \parencite{wood1996representation,zaheer2017deep,bloem2020probabilistic} 
    and references therein.
    In particular, multiple authors have given the form of a permutation equivariant weight
    matrix as 
    \[
        W = \alpha I + \beta \bm{1}\bm{1}^\top
    \]
    for scalars $\alpha, \beta \in \R$ and with $\bm{1} = (1, \dots, 1)^\top$.
    Consider an $L$-layer ReLU network with, for simplicity, widths $\kappa_i = d$ for
    all $i$.
    Let $\F$ be the class of all functions realisable by this network,
    then 
    \[
        \vc{\sign(\F)} = O(L^2 \log(Ld \log(Ld))).
    \]
\end{example}

\subsection{Regularisation for Equivariance}
Typically, the practitioner will parameterise the
weight matrices so that they satisfy \cref{eq:intertwine}.
We use the ideas of this work to suggest an alternative method that allows for
learned symmetry using regularisation.
We leave an empirical exploration of this approach to future work.

We suggest the obvious thing: 
regularise ${W^i}^\sperp$ for each layer $i=1, \dots, L$,
where ${W^i}^\sperp \coloneqq W^i - \twine^i(W^i)$.
For instance, one could add a weight decay term to the training
objective 
of the form $\sum_{i=1}^L \fnorm{{W^i}^\sperp}^2$.
Note that the operator $\twine^i$ can be computed before training.

If ${W^i}^\sperp = 0$ for $i=1, \dots, L$ and the activation function 
satisfies \cref{eq:activation},
then the resulting network will be exactly invariant or equivariant
(depending on the choice of last layer representation).
This method could also allow for approximate symmetry.
Indeed, the following result suggests $\fnorm{{W^i}^\sperp}^2$
as a measure of the layer-wise equivariance of the network.

\begin{proposition}\label{prop:regularisation}
    Let $\X=\R^d$ and $\Y=\R^k$, each with the Euclidean inner product.
    Let $f_W: \R^d \to \R^k$ with $f_W(x) = \sigma(Wx)$ be a single neural network layer 
    with $C$-Lipschitz, element-wise activation function $\sigma$.
    Let $\G$ be a compact group with orthogonal representations $\phi$ and $\psi$ on $\X$ and $\Y$
    respectively, and assume that $\psi$ commutes with $\sigma$ as in \cref{eq:activation}.
    Assume $f_W \in \Lmuy$
    where $\mu$ is a $\G$-invariant probability measure and write
    $\Sigma = \int_\X x x^\top \dd\mu(x)$ for its covariance matrix
    which we assume to be finite.
    Then the distance from $f_W$ to its closest equivariant function is bounded by
    \[
        \norm{(\id - \qq)f_W}_\mu^2 
        \le 2C^2\fnorm{W^\sperp\Sigma^{\frac12}}^2
        \le 2C^2 \fnorm{\Sigma^{\frac12}}^2
        \fnorm{W^\sperp}^2
    \]
\end{proposition}
\begin{proof}
    We do the left hand side inequality first.
    In \cref{eq:jensen} below we apply
    Jensen's inequality 
    \cref{lemma:jensen}
    which requires that the integrand is integrable,
    we verify this later. We calculate
    \begin{align*}
        \norm{\qq f_W(x) - f_W(x) }_2^2
        &=  \lnorm{\int_\G  \left(\psi(g^{-1}) f_W(\phi(g) x) - f_W(x)\right)\dd \lambda(g) }_2^2 \\ 
        &\le  \int_\G  \norm{\psi(g^{-1}) f_W(\phi(g) x) - f_W(x)}_2^2 \dd \lambda(g)
        \tag{$\dagger$}\label{eq:jensen}\\ 
        &=  \int_\G  \norm{\psi(g^{-1}) \sigma(W\phi(g) x) - \sigma(Wx)}_2^2 \dd \lambda(g)\\ 
        &=  \int_\G  
        \norm{\sigma(\overline{W}x + \psi(g^{-1})W^\sperp  \phi(g) x) - \sigma(Wx)}_2^2 \dd \lambda(g)\\ 
        &\le  C^2 \int_\G  \norm{\overline{W}x + \psi(g^{-1})W^\sperp  \phi(g) x - Wx}_2^2 \dd \lambda(g)\\ 
        &=  C^2\int_\G  \norm{(\psi(g^{-1})W^\sperp  \phi(g) - W^\sperp) x}_2^2 \dd \lambda(g)\\ 
        &= C^2 x^\top (W^\sperp)^\top W^\sperp x 
        + C^2\int_\G x^\top \phi(g^{-1})(W^\sperp)^\top W^\sperp \phi(g) x \dd\lambda(g)
    \end{align*}
    where the cross terms vanish in the final line because we must have 
    \[
        \int_\G \psi(g^{-1})W^\sperp  \phi(g)\dd\lambda(g) = 0
    \]
    (see \cref{sec:equivariant-regression}).
    Now let $X \sim \mu$.
    By inserting traces, applying Fubini's theorem and using the $\G$-invariance of 
    $\mu$ we get
    \begin{align*}
        &\norm{\qq f_W - f_W}_\mu^2 = \E[\norm{\qq f_W(X) - f_W(X) }_2^2]\\
        &\le  C^2 \tr\left((W^\sperp)^\top W^\sperp\E[X X^\top]\right) 
        + C^2 \int_\G \tr\left((W^\sperp)^\top W^\sperp\E[\phi(g)X X^\top\phi(g^{-1})]\right) 
        \dd\lambda(g)\\
        %&\le  C^2 \tr\left((W^\sperp)^\top W^\sperp\E[X X^\top]\right) 
        %+ C^2 \int_\G \tr\left((W^\sperp)^\top W^\sperp\E[X X^\top]\right) \dd\lambda(g)\\
        & = 2C^2 \tr\left((W^\sperp)^\top W^\sperp\Sigma\right) \\
        & = 2C^2 \fnorm{W^\sperp\Sigma^{\frac12}}^2 
    \end{align*}
    To complete the proof we address our application of Jensen's inequality in \cref{eq:jensen},
    which requires that for all $x\in\X$
    \begin{equation}\label{eq:jensen-condition}
        \psi(g^{-1}) f_W(\phi(g) x) - f_W(x) \in L_1(\G, \Y, \lambda).
    \end{equation}
    We can ignore the second term in the above because $\lambda$ is finite.
    By \cref{thm:bg-LpLqinclusion},
    $L_2(\G, \lambda) \subset L_1(\G, \lambda)$.
    Applying this to each coordinate,
    $L_2(\G, \Y, \lambda) \subset L_1(\G, \Y, \lambda)$.
    The function $m: (g,x) \mapsto \psi(g^{-1}) f_W(\phi(g) x)$
    is $(\lambda\otimes\mu)$-measurable by the beginning of the proof
    of \cref{prop:q-well-defined}, so
    $g \mapsto \psi(g^{-1}) f_W(\phi(g) x)$
    is $\lambda$-measurable for all $x\in\X$ by 
    using 
    \cref{lemma:measurable-collection}
    to apply to 
    \cref{corr:general-sections}
    each component.
    It is sufficient, therefore, to verify that 
    \[
        \int_\G \norm{\psi(g^{-1}) f_W(\phi(g) x)}_2^2 \dd \lambda(g) < \infty.
    \]
    Using the developments after \cref{eq:jensen} and using
    $\norm{\phi(g)}_2 = 1$ by orthogonality, it only remains to calculate
    \begin{align*}
        \int_\G x^\top \phi(g^{-1})(W^\sperp)^\top W^\sperp \phi(g) x \dd\lambda(g)
        &= \int_\G \norm{W^\sperp \phi(g) x}_2^2 \dd\lambda(g)\\
        &\le \int_\G \norm{W^\sperp}^2_2\norm{\phi(g)}^2_2\norm{x}_2^2 \dd\lambda(g)\\
        &= \norm{W^\sperp}^2_2\norm{x}_2^2 \\
        &< \infty.
    \end{align*}
    The right hand side inequality comes from, for all square matrices $A, B$,
    \[
        \fnorm{AB}^2 
        = \tr(B^\top A^\top A B) 
        = \tr(A^\top A BB^\top) 
        \le \tr(A^\top A)\tr( BB^\top) 
        =\fnorm{A}^2\fnorm{B}^2
    \]
    by Cauchy-Schwarz. 
\end{proof}

\Cref{prop:regularisation} shows that the distance between the outputs of
a single layer neural network and its closest equivariant function is bounded by 
the norm
of the $\G$-anti-symmetric component of the weights $W^\sperp$.
This quantity can be interpreted as a measure of the equivariance of the layer
and regularising $\fnorm{W^\sperp}$
will encourage the network to become (approximately)
equivariant.

\section{Connections to Other Works}\label{sec:other-work-comparison}
We discuss some connections to other works.
Apart from \parencite{mei2021learning}, these were found while researching
the literature review.
We became aware of \parencite{mei2021learning} following work 
on \parencite{elesedy2021kernel}.
We organise the comparisons by the relevant parts of this work and
adapt the results from other authors to the notation of this work.

\subsection{$\qq$ on Linear Functions}
\textcite{wood1996representation}
show that a linear map $f$ is 
equivariant if and only if
$\qq f = f$ \parencite[Lemma 2.5]{wood1996representation}.
They also define the \emph{characteristic matrix}
\parencite[Definition 3.2]{wood1996representation},
which we denote by $\proj$ in \cref{sec:invariant-regression},
show that it is the orthogonal projection onto the invariant 
elements \parencite[Theorem 3.4]{wood1996representation} and also show
that its
trace is the dimension of the invariant 
subspace \parencite[Corollary 3.5]{wood1996representation}.
Similarly, \textcite{pal2017max} show 
that when $\G$ acts by a unitary (i.e., orthogonal) representation
on $\R^d$ then $\qq$ is a self-adjoint projection on $\R^d$
\parencite[Lemma 2.3]{pal2017max}.

\subsection{Generalisation in Kernel Ridge Regression}
\textcite{mei2021learning} analyse the
generalisation of invariant random feature and kernel regression in a high
dimensional (large $d$) setting.
They study the risk in the case where the kernel is invariant by design,
whereas we use averaging to isolate the benefit of invariance.
In this way their results are more specific to what would be done in practice.
However, as is stated in \cref{thm:kernel-invariance}, if training with
an invariant kernel gives lower risk than averaging, 
then \cref{thm:kernel-invariance} applies to the former too.

The results in \parencite{mei2021learning} concern a specific
scaling between the number of training examples $n$ and the input dimension $d$,
and are restricted to specific input distributions and inner product kernels.
On the other hand, \cref{thm:kernel-invariance}
places mild assumptions on the kernel and holds for any invariant
input distribution.

It's worth mentioning that \cref{thm:kernel-invariance} and \parencite{mei2021learning}
are consistent: the representations and kernels in \parencite{mei2021learning} 
satisfy \cref{eq:kernel-switch} so from the discussion in \cref{sec:kernel-bias-term}
the lower estimate on the risk of kernel ridge regression
in \cref{thm:kernel-invariance} vanishes
in probability as $d\to \infty$, which matches the comments 
after \parencite[Proposition 1]{mei2021learning} because in their setting
$n\ge d^c$ for some constant $c>0$.

\subsection{Invariant Kernels and the Decomposition of the RKHS}
A \emph{unitary kernel} is one for which 
$k(gx, gy) = k(x, y)$
for all $x, y\in\X$ and all $g\in\G$.
This is similar to \cref{eq:kernel-switch}
and has appeared many times in the literature, for 
instance \parencite[Definition 2.2]{pal2017max},
\parencite[Definition 3.1]{pal2016discriminative} and \parencite{reisert2007learning}.
It turns out that, under this assumption,
Risi Kondor proved a weaker version of \cref{thm:rkhs-decomposition}
in his PhD Thesis. 
\begin{theorem}[{\parencite[Theorem 4.4.3]{kondor2008group}}]
    Let $\G$ be a finite group acting on $\X$ and let $\H$ be a reproducing
    kernel Hilbert space with unitary kernel $k: \X\times\X\to\R$.
    Assume further that 
    $f\circ g \in \H$
    for all $f\in\H$ and all $g\in\G$.
    Then the invariant functions in $\H$
    form a subspace which is a  reproducing kernel Hilbert space with
    kernel
    \[
        \overline{k}(x, y) = \int_\G k(x, gy) \dd\lambda(g).
    \]
\end{theorem}

Separately, \textcite{reisert2007learning} argue that for unitary kernels,
$\qq$ is an orthogonal projection onto the equivariant elements in
the RKHS $\H$.
However, it is not shown that $\qq \H \subset \H$.
Based on this, they present an equivariant version
of the representer theorem, which, specialising to invariance
and the setting of \cref{chap:kernels}, says that any invariant
solution to \cref{eq:krr} is in $\overline{\H}$ 
(defined in \cref{thm:rkhs-decomposition}).

\section{Ideas for Future Work}\label{sec:future-work}

\subsubsection*{Generalisation Gap for Non-Invariant $\mu$}
\newcommand{\Lnuy}{{L_2(\X, \Y, \nu)}}

An interesting possibility for future work is to try to remove the
assumption that $\mu$ is invariant from \cref{lemma:generalisation}.
Let $\nu$ be a probability measure on $(\X, \S_\X)$, let $X\sim\nu$ and
$Y = \fopt(X) + \xi$ with $\fopt$ equivariant, $\E[\xi] = 0$ and $\E[\xi^2]<\infty$.
Let $f\in \Lnuy$ and assume, for the sake of argument, that $\qq f \in \Lnuy$.
Then we can write $f = \bar{f} + f^\sperp$ where $\bar{f} = \qq f$ and $f^\sperp = (\id - \qq)f$,
but the terms may not be orthogonal. The generalisation gap between $f$ and
$\bar{f}$ becomes
\[
    R[f] - R[\bar{f}] 
    = \norm{f - \fopt}_\nu^2 - \norm{\bar{f}}_\nu^2
    = \norm{f^\sperp}_\nu^2 + 2\inner{\bar{f} - \fopt}{f^\sperp}_\nu.
\]
If $\nu$ is invariant, then we know from \cref{lemma:l2-decomposition}
that $\inner{\bar{f} - \fopt}{f^\sperp}_\nu = 0$.
Otherwise, this term would need to be estimated.
One approach might be to consider something like
\[
    \inner{\bar{f} - \fopt}{f^\sperp}_\nu
    \le 
    \norm{\bar{f} - \fopt}_\nu\norm{f^\sperp}_\nu
    \sup_{f, h\in\Lnuy} 
    \frac{\inner{\qq f}{(\id - \qq) h}_\nu}{\norm{\qq f}_\nu \norm{(\id - \qq)h}_\nu},
\]
The supremum term
is in the interval $[-1, 1]$ and vanishes if $\nu$ is invariant. 
It may be related to a distance between $\nu$
and the invariant measure $\bar{\nu}$ defined by $\bar{\nu}(A) =\int_\G \nu(gA) \dd\lambda(g)$
for $A\in\S_\X$.

\let\Lnuy\undefined

\subsubsection*{Application to G-CNNs}
Recent work has generalised the
standard convolutional layer to obtain new group equivariant neural network layers,
for instance \parencite{cohen2016group,cohen2018spherical,cohen2019general}.
As mentioned in \cref{sec:lit-review-nn},
\textcite{kondor2018generalization} 
show that any equivariant neural network
layer can be written in terms of a generalisation of the standard convolution.
Let $f: \X \to\Y$ and let $\X_\pi$ be a measurable cross-section of 
$\X$ with respect to $\G$. 
Then, fixing some $x_\pi\in\X_\pi$, $f$
can be \emph{lifted} to a function on $\G$, $f_{x_\pi}: \G \to \Y$
with values $f_{x_\pi}(g) = f(gx_\pi)$.
The analysis in \parencite{kondor2018generalization} is specialised to
\emph{homogeneous spaces}, those such that for all
$x, y\in\X$ $\exists g\in\G$ such that $gx=y$.
We can apply the result to each orbit independently by lifting.
Let $\Y=\R^k$, fix $x_\pi\in\X_\pi$ and 
suppose that $f$ is the output of an equivariant intermediate layer of a
G-CNN that we lift to
$f_{x_\pi}$, so both $f$ and $f_{x_\pi}$ are equivariant. 
\textcite{kondor2018generalization} show that
the (lifted) linear map in any equivariant convolutional layer must 
take the form of a \emph{group convolution} of the previous layer
with respect to some \emph{filter} $\vartheta: \G \to \R^{m\times k}$,
where $m$ is the input dimension of the following layer,
\[
    (\vartheta \ast f_{x_\pi}) (u) = \int_\G \vartheta(v)f_{x_\pi}(uv^{-1})\dd\lambda(v).
\]
Applying the equivariance of $f_{x_\pi}$, the invariance $\lambda$ and the definition
of $f_{x_\pi}$ gives
\[
    (\vartheta \ast f_{x_\pi}) (u) 
    = \int_\G \vartheta(vu)v^{-1}\dd\lambda(v)f(x_\pi)
    = \bar{\vartheta}(u) f(x_\pi)
\]
where
\[
    \bar{\vartheta}(u) = \int_\G \vartheta(vu)v^{-1}\dd\lambda(v)
\]
which looks, at least formally, rather like $\qq$. In fact,
the map $\vartheta \to \bar{\vartheta}$ is even a projection.
Could this allow for an analysis of G-CNN type architectures
using the techniques developed in \cref{chap:general-theory-i}?

\subsubsection*{Other Loss Functions}
At the core of most of our results on generalisation is \cref{lemma:generalisation},
which quantifies the generalisation gap in regression problems in terms of the 
$\G$-anti-symmetric component
of the predictor.
An interesting line of future work could be to extend this to similar results for
other loss functions. 
As a starter, one might consider the following style of argument for classification;
although \cref{corr:classification} is somewhat weaker than \cref{lemma:generalisation}
in that it does not provide a strict benefit for invariance. 
\begin{theorem}[{\parencite[Theorem 5]{krishnamurthy2017lecture12}}]\label{thm:01loss}
    Let $f: \X \to [0,1]$ and let $(X, Y)$ be a random element of 
    $\X \times \set{0, 1}$. Let $L[f] = \P(\sign(f(X) - 1/2) \ne Y)$ be
    the 0/1 risk and let $R[f] = \E[(f(X) - Y)^2]$ be the squared-error risk.
    Define $\fopt = \argmin_{f: \X \to [0, 1]}\;R[f]$, then
    \[
        L[f] - L[\fopt] \le 2 \sqrt{R[f] - R[\fopt]}.
    \]
\end{theorem}
\begin{corollary}\label{corr:classification}
    Let $X\sim\mu$. It's easy to show that 
    $
        R[f] - R[\fopt] = \norm{f - \fopt}_\mu^2
        $,
    so \cref{thm:01loss} gives $L[f] - L[\fopt] \le 2\norm{f - \fopt}_\mu$.
    If $\fopt$ is invariant, 
    using \cref{lemma:l2-decomposition} and writing
    $f = \bar{f} + f^\sperp$ 
    gives
    \[
        L[\bar{f} + f^\sperp] - L[\fopt] \le 2\norm{\bar{f} - \fopt}_\mu + \norm{f^\sperp}_\mu.
    \]
    Hence, our framework suggests an improvement in generalisation in classification
    if the invariance is correctly specified (i.e., $f^\sperp = 0$). 
\end{corollary}

\subsubsection*{Training First vs.~Averaging First}
Most of this thesis is spent studying the generalisation gap between 
an arbitrary predictor and its equivariant projection.
This corresponds to comparing a trained model with its averaged version. 
It might seem more sensible from a practical perspective, for instance in the case of regression,
to first average the features and then learn the model on the invariant/equivariant
features.
We have stressed throughout that our generalisation results
still apply if this method of averaging then training
produces a predictor with lower risk than training and then averaging.
A possible, albeit highly involved avenue for future work is
to understand the difference in generalisation of these two methods.
In the case of kernel ridge regression, some insight can be gained
by comparing our results to those in \parencite{mei2021learning}.

\subsubsection*{Invariance in other Kernel Methods}
\Cref{lemma:generalisation} holds for all predictors in $\Lmu$
so in principle could be applied to calculate the generalisation gap
for any learning algorithm with outputs in $\Lmu$.
However, in the case of an RKHS with kernel satisfying \cref{eq:kernel-switch}
there is also the RKHS inner product decomposition in \cref{thm:rkhs-decomposition}.
Could this be used to study invariance in other kernels methods?

\subsubsection*{Equivariance in Kernel Regression}
A natural extension of \cref{thm:kernel-invariance} is to equivariance.
It's possible that the equivariant matrix-valued kernels
studied by \textcite{reisert2007learning} are an equivariant generalisation of
$\overline{k}$ from \cref{chap:kernels}, or possibly of kernels
satisfying \cref{eq:kernel-switch}.
The review by \textcite{alvarez2012kernels} may also be relevant.

\appendix
\chapter{Useful Results}\label{chap:useful-results}
We provide some useful results that are relied upon elsewhere in the work.
Any proofs given are for fun/completeness and we claim no originality.

\section{Measure Theory}
\begin{lemma}[{tuples of functions \parencite[Lemma 1.8]{kallenberg2006foundations}}]
    \label{lemma:measurable-collection}
    Let $T$ be a finite set and let $(\Omega, \A)$ and $(S_t, \S_t)$ for $t\in T$
    be measurable spaces. 
    Define $S = \bigtimes_{t\in T} S_t$ and $\S = \bigotimes_{t\in T} \S_t$.
    Consider any functions 
    $f_t : \Omega \to S_t$ and define $f=(f_t: t\in T): \Omega \to S$.
    Then $f$ is $\A/\S$-measurable if and only if 
    $f_t$ is $\A/\S_t$-measurable
    for all $t\in T$.
\end{lemma}

\begin{lemma}[sections {\parencite[Lemma 1.26]{kallenberg2006foundations}}]%
    \label{lemma:section}
    Let $(S, \S, \mu)$ be a $\sigma$-finite measure space and let
    $(T, \mathcal{T}, \nu)$ be a measure space.
    Let $f: S\times T \to \R_+$ be measurable.
    Then
    \begin{enumerate}[(i)]
        \item $f_t(s) = f(s, t)$ is $\mu$-measurable for each $t\in T$;
            \label{point:1-lemma-section}
        \item $
            \bar{f}(t) = \int_S f(s, t) \dd \mu(s) 
            $ is $\nu$-measurable.
    \end{enumerate}
\end{lemma}

The following corollaries follow by applying \cref{lemma:section}
to the decomposition $f = f_+ - f_-$ where
$f_+ = \max(f, 0)$ and $f_- = \max(-f, 0)$. 
Each are integrable when $f$ is integrable and 
the integral of $f$ is defined to be the difference between the integrals
of $f_+$ and $f_-$ \parencite[p.~11]{kallenberg2006foundations}.

\begin{corollary}[sections of real functions]\label{corr:general-sections}
    Let $(S, \S, \mu)$ and $(T, \mathcal{T}, \nu)$ be probability spaces.
    Let $f: S\times T \to \R$ be measurable, then
    $f_t(s) = f(s, t)$ is $\mu$-measurable for each $t\in T$.
\end{corollary}

\begin{corollary}[sections of integrable functions]\label{lemma:integrate-section}
    Let $(S, \S, \mu)$ and $(T, \mathcal{T}, \nu)$ be 
    probability spaces
    and let $f: S\times T \to \R$ be $(\mu \otimes \nu)$-integrable,
    then 
    \[
        \bar{f}(t) = \int_S f(s, t) \dd \mu(s) 
    \]
    is $\nu$-measurable.
\end{corollary}

\section{Linear Transformations}

\begin{lemma}\label{lemma:trace-psd}
    Let $A, B \in \R^{n\times n}$ with $A$ symmetric 
    and $B$ positive semi-definite, then
    \[
       \lmax(A)\tr(B) \ge  \tr(AB) \ge \lmin(A)\tr(B)
    \]
    where $\lmin$ and $\lmax$ denote the minimum and maximum eigenvalue
    respectively.
\end{lemma}
\begin{proof}
    Write $A = \sum_{i=1}^n \gamma_i(A)v_i v_i^\top$
    where $\gamma_1(A), \dots, \gamma_n(A)$ are the eigenvalues
    of $A$ with multiplicity and $v_1, \dots, v_n \in \R^n$ are the respective
    eigenvectors that form an orthonormal basis.
    Then
    \begin{align*}
        \tr(AB) 
        &= \tr\left(\sum_{i=1}^n \gamma_i(A) v_i v_i^\top B\right)\\
        &= \sum_{i=1}^n \gamma_i(A) v_i^\top Bv_i\\
        &\ge \lmin(A) \sum_{i=1}^n v_i^\top Bv_i\\
        &= \lmin(A) \tr(B)
    \end{align*}
    and the left hand side inequality follows by the same approach.
\end{proof}

\begin{lemma}\label{lemma:op-norm}
    Let $A \in \R^{n\times n}$, then
    \[
        \norm{A}_2 \le n \max_{ij}\abs{A_{ij}}.
    \]
\end{lemma}
\begin{proof}
    Let $a_i \in \R^n$ be the $i$\textsuperscript{th} column of $A$, then
    \begin{align*}
        \sup_{\norm{x}_2 = 1} \norm{A x}_2  
        &=\sup_{\norm{x}_2 = 1} \sqrt{\sum_i (a_i^\top x)^2} \\
        &\le \sup_{\norm{x}_2 = 1} \sqrt{\sum_i \norm{a_i}_2^2 \norm{x}_2^2}  \\
        &\le \sqrt{\sum_i \norm{a_i}_2^2 }  \\
        &\le \sqrt{n^2 \max_{ij} A_{ij}^2}. 
    \end{align*}
\end{proof}

\begin{lemma}\label{lemma:pseudo-inverse}
    Let $D \in \R^{d \times d}$ be orthogonal and 
    let $B \in \R^{d \times d}$ be any symmetric matrix,
    then
    \[
        (DBD^\top)^+ = DB^+ D^\top.
    \]
\end{lemma}
\begin{proof}
    Set $X = DB^+ D^\top$ and $A = DBD^\top$.
    It suffices to check that $A$ and $X$ satisfy the Penrose equations,
    the solution of which is unique \parencite{penrose1955generalized}, namely:
    \begin{enumerate*}[i.]
        \item $AXA$ = $A$,
        \item $XAX$ = $X$,
        \item $(AX)^\top = AX$ and
        \item $(XA)^\top = XA$.
    \end{enumerate*}
    It is straightforward to check that this is the case.
\end{proof}

\begin{theorem}[{Adjoints \parencite[Theorem 4.10]{rudin1991functional}}]\label{thm:adjoint}
    Let $H_1$ and $H_2$ be (possibly infinite dimensional) Hilbert spaces.
    For every bounded linear operator $T:H_1\to H_2$ there exists a
    unique bounded linear operator $T^*:H_2\to H_1$ such that
    for all $x\in H_1$ and all $y\in H_2$
    \[
        \inner{Tx}{y}_{H_2} = \inner{x}{T^*y}_{H_1}
    \]
    and $\opnorm{T^*} = \opnorm{T}$.
\end{theorem}

\begin{theorem}[{\parencite[Theorem 4.13]{rudin1991functional}}]%
\label{theorem:bounded-below-surjective}
    Let $H$ be a Hilbert space and let $T: H\to H$ be a bounded linear operator.
    Then $T$ is surjective if and only if $\exists \delta >0$ such that
    $\forall x\in H$ $\norm{T^*x}_H\ge\delta \norm{x}_H$.
\end{theorem}

\begin{lemma}\label{lemma:ridge-operator-invertible}
    Let $H_1$ and $H_2$ be Hilbert spaces, let
    $S: H_1 \to H_2$ be a bounded linear operator with
    adjoint $S^*$ and set $\Sigma = S^*S$.
    Then for any $\rho > 0$, $\Sigma + \rho \id: H_1 \to H_1$ is invertible.
\end{lemma}

\begin{proof}
    For any $v \in H$ we have 
    \[
        \inner{\Sigma v}{v}_{H_1}
        =\inner{S^*S v}{v}_{H_1}
        =\norm{Sv}_{H_2}^2
        \ge 0
    \]
    so
    \[
        \norm{(\Sigma + \rho \id)v}_{H_1}^2
        = 
        \norm{\Sigma v}_{H_1}
        + 2\rho\inner{\Sigma v}{v}_{H_1}
        + \rho^2\norm{v}_{H_1} 
        = \norm{\Sigma v}_{H_1}^2 
        + 2\rho\norm{S v}^2_{H_2}
        + \rho^2\norm{v}_{H_1} 
    \]
    which verifies that the kernel of $\Sigma + \rho \id$
    is trivial so it is injective.
    The calculation also shows that $\Sigma + \rho \id$ is bounded
    below, i.e., $\norm{(\Sigma +\rho \id)v}_{H_1} \ge \rho\norm{v}_{H_1}$.
    Clearly $\Sigma + \rho \id$ is both bounded and self-adjoint,
    so it is surjective by \cref{theorem:bounded-below-surjective}.
\end{proof}

\begin{lemma}\label{lemma:tikhonov-solution}
    Let $H_1$ and $H_2$ be Hilbert spaces, let
    $S: H_1 \to H_2$ be a bounded linear operator with
    adjoint $S^*$ and set $\Sigma = S^*S$ and $T = SS^*$.
    Let $\rho > 0$, then the problem
    \[
        \argmin_{f \in H_1} \; \norm{S f - f'}_{H_2}^2 + \rho\norm{f}_{H_1}^2
    \]
    is solved uniquely by
    \[
        f = (\Sigma + \rho \id)^{-1} S^* f'
    \]
    and $S f = (T + \rho \id)^{-1} T f'$.
\end{lemma}

\begin{proof}
    \newcommand{\Hone}{{H_1}}
    \newcommand{\Htwo}{{H_2}}
    We want to minimise $L_\rho[f] =\norm{S f - f'}_\mu^2 + \rho\norm{f}_\Hone^2$
    over $f \in \Hone$. For all non-zero $h \in \Hone$
    \begin{align*}
        L_\rho[f + h]
        &= \norm{S f + h - f'}_\Htwo^2 + \rho\norm{f + h}_\Hone^2\\
        %&= \norm{S f - f'}_\Htwo^2 + \rho \norm{f}_\Hone^2
        %+ 2\inner{S f -f'}{S h}_\Htwo + 2\rho\inner{f}{h}_\Hone
        %+ \norm{S h}_\Htwo^2 + \rho\norm{h}_\Hone^2\\
        &= L_\rho[f]
        + 2\inner{S f -f'}{S h}_\Htwo + 2\rho\inner{f}{h}_\Hone
        + \norm{S h}_\Htwo^2 + \rho\norm{h}_\Hone^2\\
        &> L_\rho[f]
        + 2\inner{S f -f'}{S h}_\Htwo + 2\rho\inner{f}{h}_\Hone.
    \end{align*}
    Any $f$ that solves $ \inner{S f -f'}{S h}_\Htwo + \rho\inner{f}{h}_\Hone = 0$
    simultaneously for all $h \in \Hone$ is then the unique global minimum of $L_\rho$.
    We calculate
    \begin{align*}
        \inner{S f -f'}{S h}_\Htwo + \rho\inner{f}{h}_\Hone
        &= \inner{S^*(S f -f')}{h}_\Hone + \rho\inner{f}{h}_\Hone\\
        &= \inner{(\Sigma + \rho \id)f - S^* f'}{h}_\Hone \\
        &= \inner{(\Sigma + \rho \id)f - S^* f'}{h}_\Hone.
    \end{align*}
    This completes the first part of the proof, 
    with \cref{lemma:ridge-operator-invertible} verifying that
    $\Sigma + \rho \id$ is invertible.
    Finally, let $h\in\Htwo$ and set
    $f = (\Sigma + \rho \id)^{-1}S^* h$ then
    \[
        S(S^*S + \rho \id) f = S S^* h
    \]
    which implies $Sf = (T + \rho \id)^{-1} T h$.
    Again, \cref{lemma:ridge-operator-invertible} validates
    the inverse.
\end{proof}

\section{Probability and Statistics}

\begin{lemma}[{Jensen's inequality \parencite[Lemma 3.5]{kallenberg2006foundations}}]%
    \label{lemma:jensen}
    Let $\xi$ be an integrable element of $\R^d$ and let $f: \R^d\to\R$
    be convex, then
    \[
        \E[f(\xi)] \le f(\E[\xi]).
    \]
\end{lemma}

\subsection{Inverse Wishart Matrices}\label[appendix]{sec:wishart}

\begin{lemma}[\parencite{gupta1968some}]\label{lemma:expected-inv-wishart}
    Let $\xx \in \R^{n \times d}$ have \iid~$\normal(0, 1)$ elements with $n > d + 1$. Then
    \[
        \E[(\xx^\top \xx)^+] = \frac{1}{n - d - 1}I.
    \]
\end{lemma}

\begin{remark}
    It is well known that the expectation in \Cref{lemma:expected-inv-wishart}
    diverges for $d \le n \le d+1$.
    To see this, first notice that since the normal distribution is $\orth_d$ invariant
    $R\E[(\xx^\top \xx)^+]R^\top = \E[(\xx^\top \xx)^+]$ for any $R \in \orth_d$ 
    by \cref{lemma:pseudo-inverse}.
    Hence $\E[(\xx^\top \xx)^+]$ is a scalar multiple of the identity: it is symmetric
    so diagonalisable, hence diagonal in every basis by the invariance, then 
    permutation matrices can be used to show the diagonals are all equal.
    It remains to consider the eigenvalues. 
    The eigenvalues $\lambda_1, \dots, \lambda_d$ of $\xx^\top \xx$ have 
    joint density (w.r.t.~Lebesgue) that is proportional to 
    \[
        \exp\left(-\frac12 \sum_{i=1}^d \lambda_i\right) 
        \prod_{i=1}^d \lambda_i^{(n - d - 1)/2} \prod_{i < j}^d \abs{\lambda_i - \lambda_j}  
    \]
    when $n \ge d$ and $0$ otherwise \parencite[Corollary 3.2.19]{muirhead2009aspects}.
    We need to calculate the mean of $1 / \lambda$ with respect to this density, 
    which diverges unless $n \ge d+2$.
    Taking the mean of $\lambda_k^{-1}$, there is a term from the expansion of
    the Vandermonde product that
    does not contain $\lambda_k$,
    so the integrand in the expectation goes like $\sqrt{\lambda_k^{n - d -3}}$ as $\lambda_k \to 0$.
\end{remark}

\begin{lemma}[{\parencite[Theorem 2.1]{cook2011mean}}]\label{lemma:expected-inv-wishart-singular}
    Let $\xx \in \R^{n \times d}$ have \iid~$\normal(0, 1)$ elements with $n < d - 1$. Then
    \[
        \E[(\xx^\top \xx)^+] = \frac{n}{d(d - n - 1)}I.
    \]
\end{lemma}
\begin{remark}
   The statement of \Cref{lemma:expected-inv-wishart-singular} in
   \parencite[Theorem 2.1]{cook2011mean}
   gives the condition $n < d - 3$, but this is not necessary for the first moment,
   which can be seen from the proof.
   In addition, the proof uses a transformation followed by an application of 
   \Cref{lemma:expected-inv-wishart}
   with the roles of $n$ and $d$ switched.
    Using this transformation, it follows (e.g., from the earlier remark)
    that the expectation diverges when $d \ge n \ge d-1$.
\end{remark}

\subsection{Isotropic Random Projections}
\begin{lemma}[Will Sawin]\label{lemma:proj-variance}
    Let $E \sim \unif \Gr_n(\R^d)$ where $0 < n < d$ 
    and let $P_E$ be the orthogonal projection onto
    $E$, then in components
    \[
        \E[P_E \otimes P_E]_{abce} =
        \frac{n(d-n)}{d(d-1)(d+2)}(\delta_{ab}\delta_{ce} +
         \delta_{ac}\delta_{be} +  \delta_{ae}\delta_{bc})
        + \frac{n(n-1)}{d(d-1)} \delta_{ab}\delta_{ce}.
    \]
\end{lemma}
\begin{proof}
    We use the Einstein convention of implicitly summing over repeated indices.
    The distribution of $E$ is orthogonally invariant, so $\E[P_E \otimes P_E]$ is isotropic.
    Thus, $\E[P_E \otimes P_E]$ must have components (e.g., by \parencite{hodge1961})
    \[
        \Gamma_{abce}\coloneqq 
        \E[P_E \otimes P_E]_{abce}
        = \alpha \delta_{ab}\delta_{ce} + \beta \delta_{ac}\delta_{be} + \gamma \delta_{ae}\delta_{bc}.
    \]
    Contracting indices gives
    \begin{alignat*}{2}
        &n^2 &&= \E[\tr(P_E)^2] = \Gamma_{aabb} = d^2 \alpha + d\beta + d\gamma \\ 
        &n &&= \E[\tr(P_E^\top P_E)] = \Gamma_{abab} = d\alpha + d^2\beta + d\gamma \\
        &n &&= \E[\tr(P_E^2)] = \Gamma_{abba} = d\alpha + d\beta + d^2\gamma
    \end{alignat*}
    from which one finds
    \begin{align*}
        \beta &= \frac{n(d-n)}{d(d-1)(d+2)}\\
        \alpha &= \beta + \frac{n(n-1)}{d(d-1)} \\ 
        \gamma &= \beta.
    \end{align*}
\end{proof}

\section{Reproducing Kernel Hilbert Spaces}
RKHS stands for reproducing kernel Hilbert space.
See \cref{sec:rkhs-basics} for a brief introduction.

\begin{lemma}[{RHKS of measurable kernels \parencite[Lemma 4.24]{steinwart2008support}}]%
    \label{lemma:rkhs-measurable}
    Let $\H$ be an RKHS of functions $f: \X\to\R$ with kernel $k:\X\times\X\to\R$.
    Then all $f\in\H$ are measurable if and only if $k(\cdot, x): \X\to\R$ is measurable
    for all $x\in\X$.
\end{lemma}

\begin{lemma}[{RHKS of continuous functions \parencite[Lemma 4.28]{steinwart2008support}}]%
    \label{lemma:rkhs-continuous}
    Let $\H$ be an RKHS of functions $f: \X\to\R$ with kernel $k:\X\times\X\to\R$.
    Then all $f\in\H$ are bounded and continuous if and only if 
    $k(\cdot, x): \X\to\R$ is bounded and continuous for all $x\in\X$.
\end{lemma}

\begin{lemma}[{separable RKHS \parencite[Lemma 4.33]{steinwart2008support}}]%
    \label{lemma:rkhs-separable}
    Let $(\X, \tau)$ be a separable topological space and let
    $k: \X \times \X \to \R$ be a continuous kernel, then
    the RKHS of $k$ is separable.
\end{lemma}

\begin{theorem}[{integral operators of kernels %
\parencite[Theorem 4.26]{steinwart2008support}}]\label{thm:rkhs-integral-op-1}
    Let $(\X, \S_\X, \mu)$ be a $\sigma$-finite measure space and let
    $\H$ be a separable RKHS of functions $f: \X \to\R$ with measurable 
    kernel $k:\X\times\X\to\R$.
    Let $p\in [1, \infty)$ and define the function $m_k: \X\to\R$ by
    $m_k(x) = \sqrt{k(x, x)}$. 
    If 
    \[
        \norm{m_k}_{L_p(\mu)} \coloneqq 
        \left(\int_\X m_k(x)^p \dd\mu(x)\right)^{\frac1p} < \infty
    \]
    then all $f\in\H$ satisfy $\norm{f}_{L_p(\mu)} < \infty$
    and the inclusion map $\iota : \H \to L_p(\mu)$ is continuous
    with operator norm $\opnorm{\iota}\le \norm{m_k}_{L_p(\mu)}$.
    Moreover, the adjoint of the inclusion operator is $S_k: L_q(\mu) \to \H$
    where
    \[
        S_k f(x) = \int_\X k(x, x')f(x') \dd\mu(x')
    \]
    and $1/p + 1/q = 1$. In addition:
    \begin{enumerate}[i)]
        \item $\iota \H$ is dense in $L_p(\mu)$ if and only if $S_k$
            is injective
        \item $S_k L_q(\mu)$ is dense in $\H$ if and only if $\iota$ is injective.
    \end{enumerate}
\end{theorem}

\chapter{Excluded Works}\label{chap:other-phd-work}

\newcommand{\paper}[5]{%
        \section*{#1} % Name
        \begin{center}
            #2\\ % Authors
            \medskip
            \ifempty{#3}{}{\emph{#3}\\} % Venue
            \medskip
            \ifempty{#4}{}{\href{#4}{#4}\\} % Link
        \end{center}
        \bigskip
        \subsubsection*{Abstract}
        #5 % Abstract
        \clearpage
    }

\newcommand{\paperref}[1]{\emph{\nameref{#1}}}

\paper{%
        Lottery Tickets in Linear Models: An Analysis of Iterative Magnitude Pruning
    }{%
        {\bfseries Bryn Elesedy}, Varun Kanade, Yee Whye Teh
    }{%
        Sparsity in Neural Networks Workshop, 2021
    }{%
        https://arxiv.org/abs/2007.08243%
    }{%
        We analyse the pruning procedure behind the lottery ticket
        hypothesis \parencite{frankle2018lottery},
        \emph{iterative magnitude pruning} (IMP), when applied to linear models trained by
        gradient flow.  We begin by presenting sufficient conditions on the
        statistical structure of the features under which IMP prunes those
        features that have smallest projection onto the data.  Following this,
        we explore IMP as a method for sparse estimation.
}

\paper{%
        Effectiveness and Resource Requirements of Test, Trace and Isolate Strategies
    }{%
        Bobby He*, Sheheryar Zaidi*, {\bfseries Bryn Elesedy}*, Michael Hutchinson*,
        Andrei Paleyes, Guy Harling, Anne Johnson and Yee Whye Teh, on behalf of Royal
        Society DELVE group (*equal contribution)
    }{%
        Royal Society Open Science, 2021
    }{%
        https://royalsocietypublishing.org/doi/10.1098/rsos.201491%
    }{%
        We use an individual-level transmission and contact simulation model to explore
        the effectiveness and resource requirements of various test-trace-isolate (TTI)
        strategies for reducing the spread of SARS-CoV-2 in the UK, in the context of
        different scenarios with varying levels of stringency of non-pharmaceutical
        interventions. Based on modelling results, we show that self-isolation of
        symptomatic individuals and quarantine of their household contacts has a
        substantial impact on the number of new infections generated by each primary
        case. We further show that adding contact tracing of non-household contacts of
        confirmed cases to this broader package of interventions reduces the number of
        new infections otherwise generated by 5–15\%. We also explore impact of key
        factors, such as tracing application adoption and testing delay, on overall
        effectiveness of TTI.
}

\paper{%
        Efficient Bayesian Inference of Instantaneous Reproduction Numbers at
        Fine Spatial Scales, with an Application to Mapping and Nowcasting the
        COVID-19 Epidemic in British Local Authorities.  
    }{%
        Yee Whye Teh, 
        Avishkar Bhoopchand*, Peter Diggle*, {\bfseries Bryn Elesedy}*,
        Bobby He*, Michael Hutchinson*, Ulrich Paquet*, Jonathan Read*, Nenad Tomasev*,
        Sheheryar Zaidi* (*alphabetical ordering).
    }{%
        Royal Society Special Topic Meeting on R, Local R and Transmission of COVID-19.
    }{%
        https://rss.org.uk/RSS/media/File-library/News/2021/WhyeBhoopchand.pdf%
    }{%
        The spatio-temporal pattern of Covid-19 infections, as for most infections
        disease epidemics, is highly heterogeneous as a consequence of local variations
        in risk factors and exposures. Consequently, the widely quoted national-level
        estimates of reproduction numbers are of limited value in guiding local
        interventions and monitoring their effectiveness. It is crucial for national
        and local policy makers as well as health protection teams that accurate,
        well-calibrated and timely predictions of Covid-19 incidences and transmission
        rates are available at fine spatial scales. 
        Obtaining such estimates is challenging, not least due to the prevalence of
        asymptomatic Covid-19 transmissions, as well as difficulties of obtaining high
        resolution and frequency data. In addition, low case counts at a local level
        further confounds the inference for Covid-19 transmission rates, adding
        unwelcome uncertainty. 
        In this paper we develop a hierarchical Bayesian method for inference of
        incidence and transmission rates at fine spatial scales.  Our model
        incorporates both temporal and spatial dependencies of local transmission rates
        in order to share statistical strength and reduce uncertainty. It also
        incorporates information about population flows to model potential
        transmissions across local areas.  A simple approach to posterior simulation
        quickly becomes computationally infeasible, which is problematic if the system
        is required to provide timely predictions.  We describe how to make  posterior
        simulation for the model efficient, so that we are able to provide daily
        updates on epidemic developments.
        Real-time estimates from our model can be viewed on our website:
        \href{https://localcovid.info/}{localcovid.info}.
    }

\paper{%
        U-Clip: On-Average Unbiased Stochastic Gradient Clipping
    }{%
        {\bfseries Bryn Elesedy}, Marcus Hutter
    }{%
        preprint
    }{%
        https://arxiv.org/abs/2302.02971%
    }{%
        U-Clip is a simple amendment to gradient clipping that can be applied
        to any iterative gradient optimization algorithm. Like regular
        clipping, U-Clip involves using gradients that are clipped to a
        prescribed size (e.g., with component wise or norm based clipping) but
        instead of discarding the clipped portion of the gradient, U-Clip
        maintains a buffer of these values that is added to the gradients on
        the next iteration (before clipping). We show that the cumulative bias
        of the U-Clip updates is bounded by a constant. This implies that the
        clipped updates are unbiased on average. Convergence follows via a
        lemma that guarantees convergence with updates $u_i$ as long as
        $\sum_{i=1}^t (u_i - g_i) = o(t)$ 
        where $g_i$ are the gradients. Extensive experimental
        exploration is performed on CIFAR10 with further validation given on
        ImageNet.  

        \subsubsection*{Note}
        Work performed while on an internship at DeepMind.
        Unfortunately, after finishing, we found out that the main results
        are basically a special case of \parencite{karimireddy2019error}.
}

%%next line adds the Bibliography to the contents page
%%uncomment next line to change bibliography name to references
\renewcommand{\bibname}{References}
\addcontentsline{toc}{chapter}{References}
%\bibliography{refs}        %use a bibtex bibliography file refs.bib
%\bibliographystyle{plain}  %use the plain bibliography style

\printbibliography

\end{document}

%\begin{document}

%\begin{abstract}
%\end{abstract}